\documentclass[11pt]{article}
\usepackage[a4paper,margin=1in,headheight=14pt]{geometry}
\usepackage[T1]{fontenc}
\usepackage{amsmath,amsthm,mathtools}
\usepackage{XCharter}
\usepackage[xcharter,vvarbb]{newtxmath}
\IfFileExists{inter.sty}{\usepackage[scale=0.9]{inter}}{}

\usepackage[final]{microtype}
\usepackage{graphicx,booktabs,array,multirow,tabularx}
\usepackage[dvipsnames,svgnames,table]{xcolor}
\usepackage{enumitem,xspace,nicefrac,bm}
\usepackage[font=footnotesize,labelfont={bf,sf},labelsep=space,skip=6pt]{caption}
\usepackage{titlesec}
\usepackage{fancyhdr}
\usepackage[numbers,super,sort&compress]{natbib}
\usepackage{placeins}
\usepackage{aliascnt}

\definecolor{ink}{HTML}{1F2937}
\definecolor{muted}{HTML}{6B7280}
\definecolor{linkblue}{HTML}{1D4ED8}
\definecolor{citeteal}{HTML}{0F766E}

\usepackage[colorlinks=true,linkcolor=linkblue,citecolor=citeteal,urlcolor=linkblue,
            pdfauthor={Diaoxuan He, Linqian Wu, Xicheng Wang, Chang Kong and Haoming Peng},pdfusetitle,bookmarksnumbered]{hyperref}
\usepackage[capitalize,noabbrev,nameinlink]{cleveref}

\titleformat{\section}{\sffamily\bfseries\large\color{ink}}{\thesection}{0.7em}{}
\titleformat{\subsection}{\sffamily\bfseries\normalsize\color{ink}}{\thesubsection}{0.6em}{}
\titleformat{\subsubsection}{\sffamily\bfseries\small\color{ink}}{\thesubsubsection}{0.6em}{}
\titleformat{\paragraph}[runin]{\bfseries\color{ink}}{}{}{}
\titlespacing*{\section}{0pt}{2.2ex plus 0.6ex minus 0.3ex}{1.0ex plus 0.2ex}
\titlespacing*{\subsection}{0pt}{1.7ex plus 0.5ex minus 0.3ex}{0.6ex plus 0.2ex}
\titlespacing*{\subsubsection}{0pt}{1.4ex plus 0.5ex minus 0.3ex}{0.5ex plus 0.2ex}
\titlespacing*{\paragraph}{0pt}{0.9ex plus 0.3ex minus 0.2ex}{0.6em}

\fancypagestyle{plain}{\fancyhf{}\fancyfoot[C]{\small\thepage}}

\newtheoremstyle{sfthm}{0.6\baselineskip}{0.6\baselineskip}{\itshape}{}{\sffamily\bfseries}{.}{0.5em}{}
\theoremstyle{sfthm}
\newtheorem{theorem}{Theorem}
\newaliascnt{proposition}{theorem}
\newtheorem{proposition}[proposition]{Proposition}
\aliascntresetthe{proposition}
\newaliascnt{lemma}{theorem}
\newtheorem{lemma}[lemma]{Lemma}
\aliascntresetthe{lemma}
\newaliascnt{corollary}{theorem}
\newtheorem{corollary}[corollary]{Corollary}
\aliascntresetthe{corollary}
\theoremstyle{definition}
\newtheorem{definition}{Definition}
\crefname{theorem}{Supplementary Theorem}{Supplementary Theorems}
\crefname{proposition}{Supplementary Proposition}{Supplementary Propositions}
\crefname{lemma}{Supplementary Lemma}{Supplementary Lemmas}
\crefname{corollary}{Supplementary Corollary}{Supplementary Corollaries}
\crefname{definition}{Supplementary Definition}{Supplementary Definitions}
\crefname{figure}{Fig.}{Figs.}
\crefname{table}{Table}{Tables}
\crefname{equation}{equation}{equations}
\crefname{edfigure}{Extended Data Fig.}{Extended Data Figs.}
\crefname{edtable}{Extended Data Table}{Extended Data Tables}
\crefname{sfigure}{Supplementary Fig.}{Supplementary Figs.}
\crefname{stable}{Supplementary Table}{Supplementary Tables}
\crefname{snote}{Supplementary Note}{Supplementary Notes}
\crefname{sequation}{Supplementary equation}{Supplementary equations}
\crefname{msection}{}{}
\creflabelformat{msection}{#2`#1'#3}

\newcommand{\vphi}{\varphi}
\newcommand{\vphiraw}{\varphi_{\mathrm{raw}}}
\DeclareMathOperator*{\argmin}{arg\,min}

\newcommand{\KL}{\mathrm{KL}}
\newcommand{\kl}{\mathrm{kl}}
\newcommand{\TV}{\mathrm{TV}}
\newcommand{\Hnull}{\mathcal H_c}
\newcommand{\cl}[1]{\overline{#1}}
\newcommand{\Mt}{M_\tau}
\newcommand{\E}{\mathbb E}

\setlist{itemsep=2pt,topsep=3pt}
\renewcommand{\figurename}{Fig.}

\title{The price of knowing how a network computes}
\author{Diaoxuan He \and Linqian Wu \and Xicheng Wang \and Chang Kong \and Haoming Peng}

\begin{document}
\thispagestyle{plain}
\begin{center}
\vspace*{2pt}
{\sffamily\bfseries\LARGE The price of knowing how a network computes\par}
\vspace{18pt}
{\large Diaoxuan (Nora) He\textsuperscript{1,$\ast$}\hspace{0.25em}Linqian Wu\textsuperscript{2,3,$\ast$,$\dagger$}\hspace{0.25em}Xicheng Wang\textsuperscript{4}\hspace{0.25em}Chang Kong\textsuperscript{5}\hspace{0.25em}Haoming Peng\textsuperscript{6}\par}
\vspace{6.4326pt}
{\normalsize\renewcommand{\arraystretch}{1}%
\begin{tabular}{@{}c@{}}
\textsuperscript{1}\,University of Debrecen, Debrecen, Hungary\\
\textsuperscript{2}\,University of Waterloo, Waterloo, Ontario, Canada\\
\textsuperscript{3}\,Perimeter Institute for Theoretical Physics, Waterloo, Ontario, Canada\\
\textsuperscript{4}\,Massachusetts Institute of Technology, Cambridge, Massachusetts, USA\\
\textsuperscript{5}\,National University of Singapore, Singapore\\
\textsuperscript{6}\,The Chinese University of Hong Kong, Hong Kong
\end{tabular}\par}
\vspace{8pt}
{\small \textsuperscript{$\ast$}\,These authors contributed equally\qquad \textsuperscript{$\dagger$}\,e-mail: {\hypersetup{urlcolor=black}\href{mailto:lwu1@perimeterinstitute.ca}{\texttt{lwu1@perimeterinstitute.ca}}}\par}
\end{center}
\vspace{10pt}

% Main display items.  Panel letters: \p{a}
\newcommand{\p}[1]{\textbf{#1}}

\newcommand{\FigOverview}{%
\begin{figure}[tp]
\centering
\includegraphics[width=\linewidth]{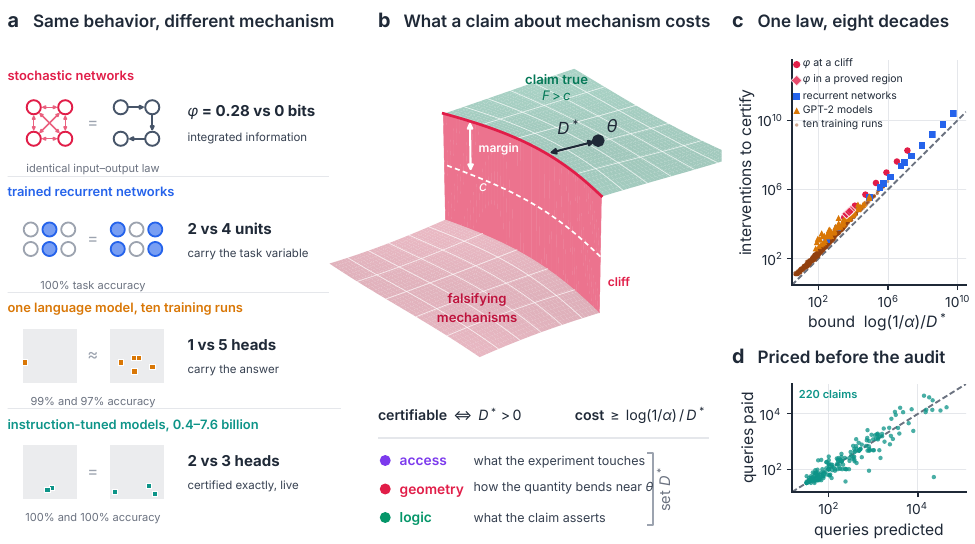}
\caption{\textbf{What a finite experiment can certify about how a network computes.}
\p{a}, Systems that behave alike compute differently, at every scale studied here. A recurrent network and a feed-forward chain with identical input--output laws differ in integrated information $\vphi$; recurrent networks with $100\%$ task accuracy differ in the number of hidden units that carry the task variable; two of ten training runs of one language model, which differ only in random seed, carry the answer to the same task in one attention head and in five (smallest sufficient sets found at threshold $0.9$; accuracy $99\%$ and $97\%$); and two instruction-tuned models that both recall facts with $100\%$ accuracy have circuits certified to be exactly two and exactly three heads (1.5 and 3 billion parameters; the audits span $0.4$ to $7.6$ billion). Grids: layers by heads; dark: a sufficient set.
\p{b}, A claim $F>c$ about a mechanism $\theta$ holds on one side of an edge in the space of mechanisms. What an experiment can establish is set by the information distance $D^\ast$ from $\theta$ to the nearest mechanism that falsifies the claim, not by the margin $F-c$: the claim can be certified exactly when $D^\ast>0$, and then costs at least $\log(1/\alpha)/D^\ast$ interventions at error level $\alpha$. Access, geometry and logic set $D^\ast$.
\p{c}, Median number of interventions needed to certify a claim against this lower bound, for $\vphi$ beside a cliff (circles) and at four mechanisms in and around a proved region (diamonds), for circuit claims in recurrent networks (squares), in six GPT-2 models (triangles) and along ten training runs of a language model (dots; seven claims with an unfinished audit are omitted). Every audit of every plotted claim ended in a certificate.
\p{d}, Live audits of seven instruction-tuned language models: interventions paid for each of 220 certified claims against the cost predicted when the claim was registered.}
\label{fig:overview}
\end{figure}}

\newcommand{\FigAccess}{%
\begin{figure}[tp]
\centering
\includegraphics[width=0.9\linewidth]{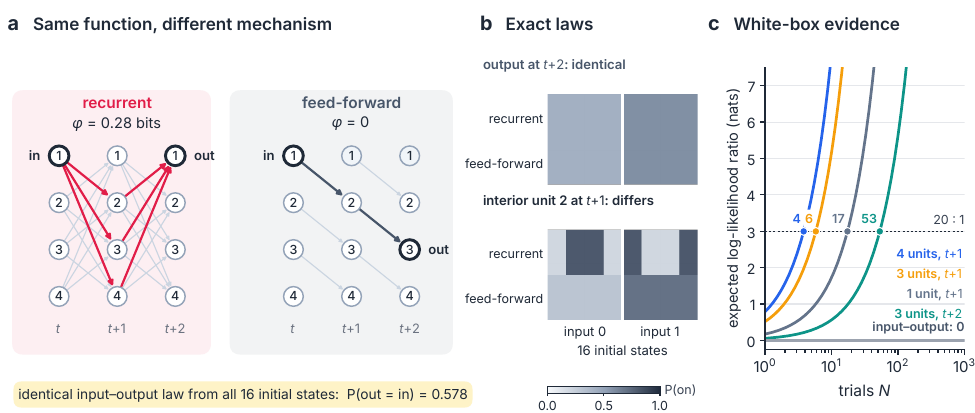}\\[4pt]
\includegraphics[width=0.9\linewidth]{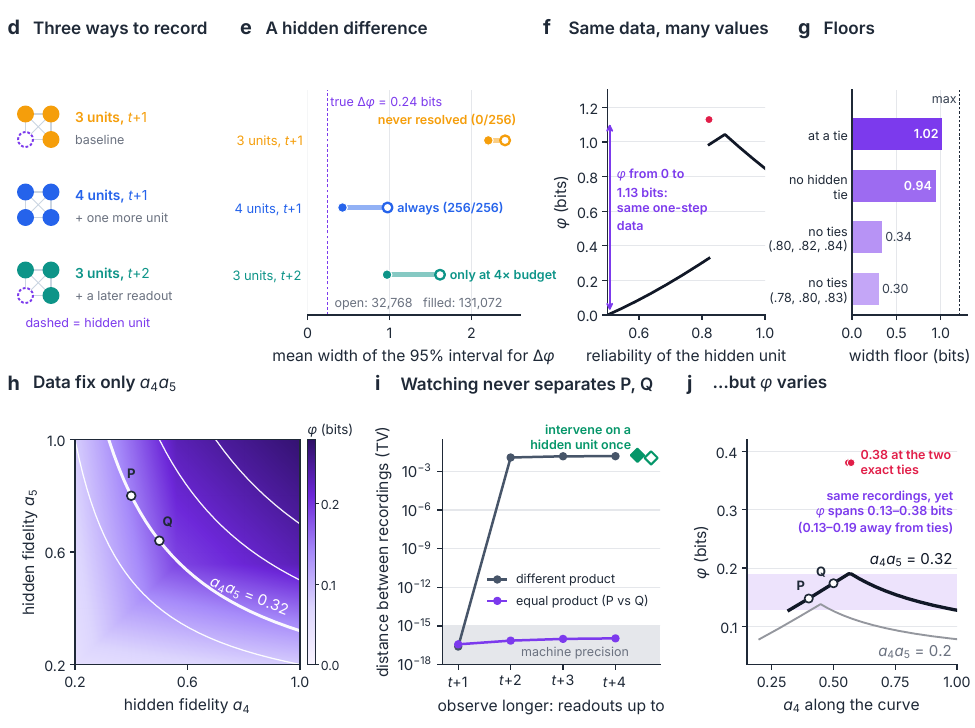}
\caption{\textbf{Access: behavior and partial recordings can be exactly blind.}
\p{a}, A recurrent parity network ($\vphi=0.28$ bits) and a feed-forward chain ($\vphi=0$), unrolled over two updates. Highlighted arrows trace how the input (unit~1 at time $t$) reaches the output: through every other unit in the recurrent network, along a single path in the chain.
\p{b}, Exact probability that a unit is on, from each of the 16 initial states, grouped by the input bit. The output at $t{+}2$ coincides in the two systems (top); interior unit~2 at $t{+}1$ differs (bottom).
\p{c}, Expected log-likelihood ratio in favor of the recurrent network, when it is the true system, as trials accumulate, for four internal recordings (sensor error $0.10$). Markers: number of trials at which the expectation reaches $\ln20$. Input--output data give zero evidence at any number of trials.
\p{d}, Three designs record the four-unit parity network after the same intervention; the dashed unit is never recorded.
\p{e}, Mean width of $95\%$ intervals for the difference $\Delta\vphi$ between two conditions that differ only in the hidden unit (256 paired datasets per design and budget), and how often its sign was resolved. Open and filled markers: budgets of 32{,}768 and 131{,}072 per condition; dashed line: the true $\Delta\vphi=0.24$ bits.
\p{f}, With recorded reliabilities fixed at $421/512$, every hidden reliability on the horizontal axis gives the same one-step recording of units 1--3, so every value of $\vphi$ on the curve and at the red dot (an exact four-way tie), from $0$ to $1.13$ bits, is compatible with the data.
\p{g}, Lower bounds on the expected width of any $95\%$ interval, at any sample size, for four sets of recorded reliabilities. Dashed line: the largest possible $\vphi$ of a four-unit system, $1.21$ bits.
\p{h}, $\vphi$ of a five-unit network over the fidelities $a_4,a_5$ of its two hidden units. Observation reveals only the product $a_4a_5$ (white curves; thick: $a_4a_5=0.32$, through configurations P and Q).
\p{i}, Total-variation distance between the recorded trajectories of P and Q (equal product) and of a pair with a different product, as the recording is extended; the residual $10^{-16}$ is floating-point round-off. Setting one hidden unit once (diamonds) separates P from Q.
\p{j}, $\vphi$ along curves of constant product. Shaded band: the spread of $\vphi$ along $a_4a_5=0.32$ away from exact ties; red dots: the two exact ties on that curve.}
\label{fig:access}
\end{figure}}

\newcommand{\FigGeometry}{%
\begin{figure}[tp]
\centering
\includegraphics[width=0.9\linewidth,trim=0 2.36in 0 0,clip]{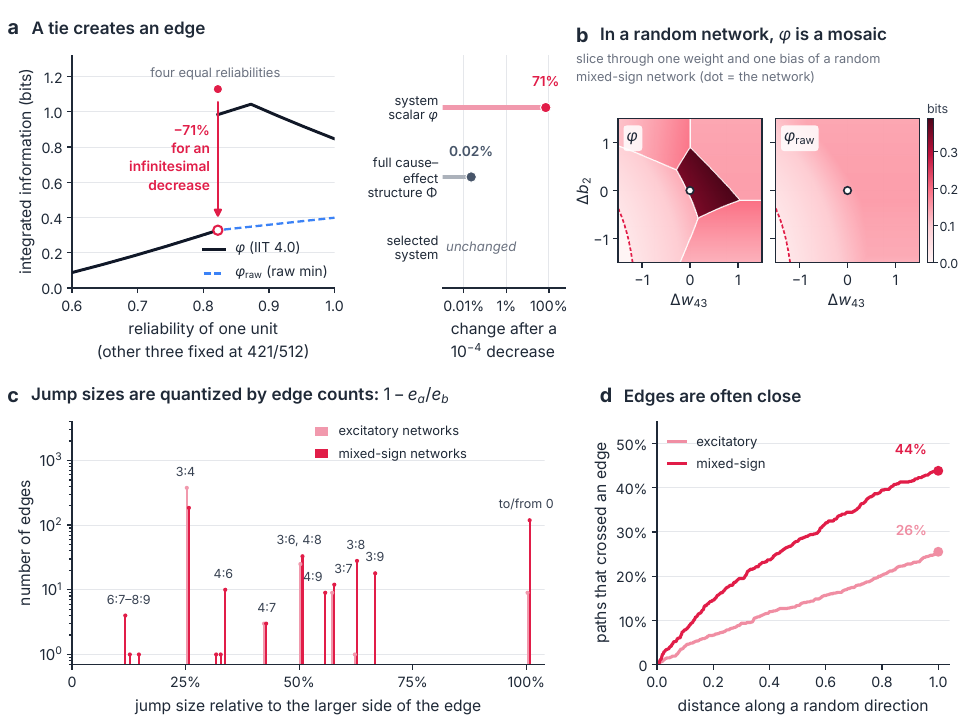}\\[4pt]
\includegraphics[width=0.9\linewidth]{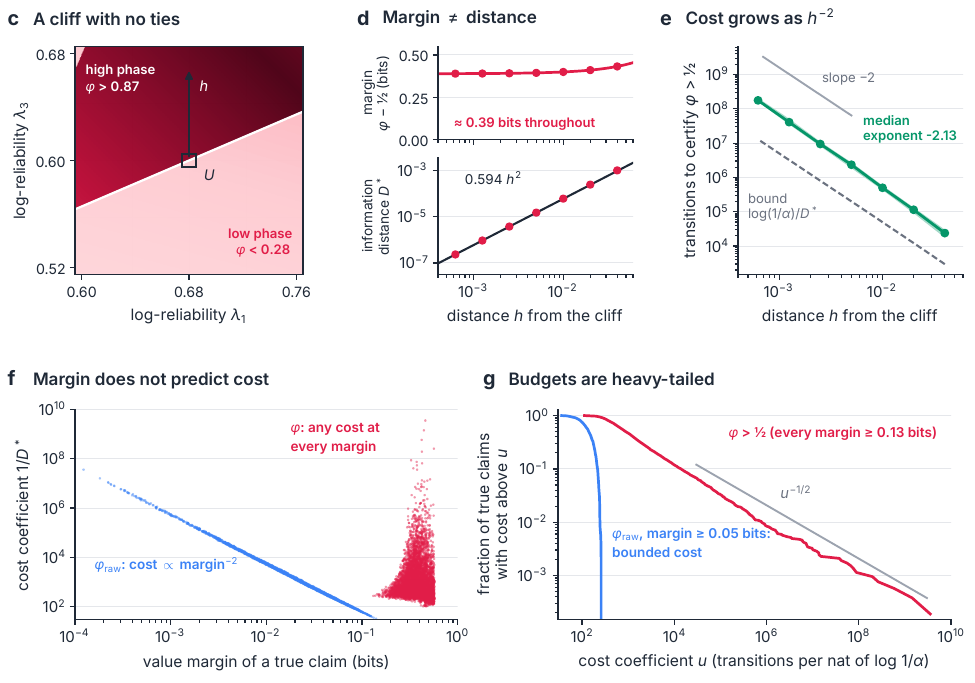}
\caption{\textbf{Geometry: cliffs are robust, and margin is not evidence.}
\p{a}, $\vphi$ (solid) and the comparison quantity $\vphiraw$ (dashed) of a parity network as one unit's reliability varies with the other three at $421/512$. At the four-way tie $\vphi=1.13$ bits; an arbitrarily small decrease gives $0.33$ bits. Right: over the same $10^{-4}$ perturbation the full cause--effect structure of IIT changes by $0.02\%$ and the selected system not at all.
\p{b}, $\vphi$ and $\vphiraw$ over a slice through one weight and one bias of a random mixed-sign network (dot: the network). White lines: discontinuities where the selected cut changes its number of severed connections; dashed line: a switch of the specified cause and effect states.
\p{c}, $\vphi$ over a slice of the parity family through $\lambda^0$: a low phase and a high phase separated by a hyperplane, with no ties and no symmetry. Square: the region $U$ of the theorem, to scale; arrow: the path into the high phase.
\p{d}, Along the path, the margin $\vphi-1/2$ stays near $0.39$ bits (top) while the information distance to the falsifying mechanisms falls as $0.594\,h^2$ (bottom; dots: numerical projection onto the exact falsifying set; line: theory).
\p{e}, Transitions needed by the direct certificate at $\alpha=0.05$ (median and interquartile range of 300 runs per distance; every run certified), with the lower bound $\log(1/\alpha)/D^\ast$ (dashed).
\p{f}, Over 6{,}000 mechanisms, the cost coefficient $1/D^\ast$ of each true claim against its margin, for $\vphi>1/2$ (red) and for $\vphiraw$ at its population median (blue). For $\vphiraw$ a margin buys a bounded budget; for $\vphi$ it does not.
\p{g}, Fraction of true claims whose cost coefficient exceeds $u$. For $\vphi$ the tail follows $u^{-1/2}$ although every such claim holds with a margin of more than $0.12$ bits; for $\vphiraw$, claims with a margin of at least $0.05$ bits have bounded cost (blue).}
\label{fig:geometry}
\end{figure}}

\newcommand{\FigCircuits}{%
\begin{figure}[tp]
\centering
\includegraphics[width=0.9\linewidth,trim=0 2.36in 0 0,clip]{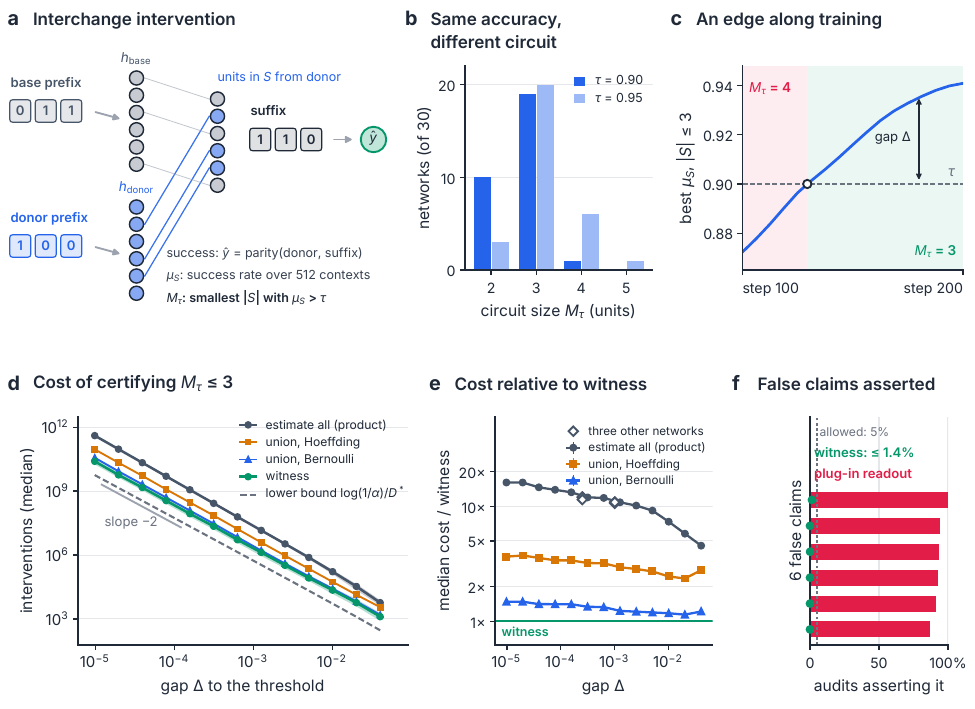}\\[4pt]
\includegraphics[width=0.9\linewidth]{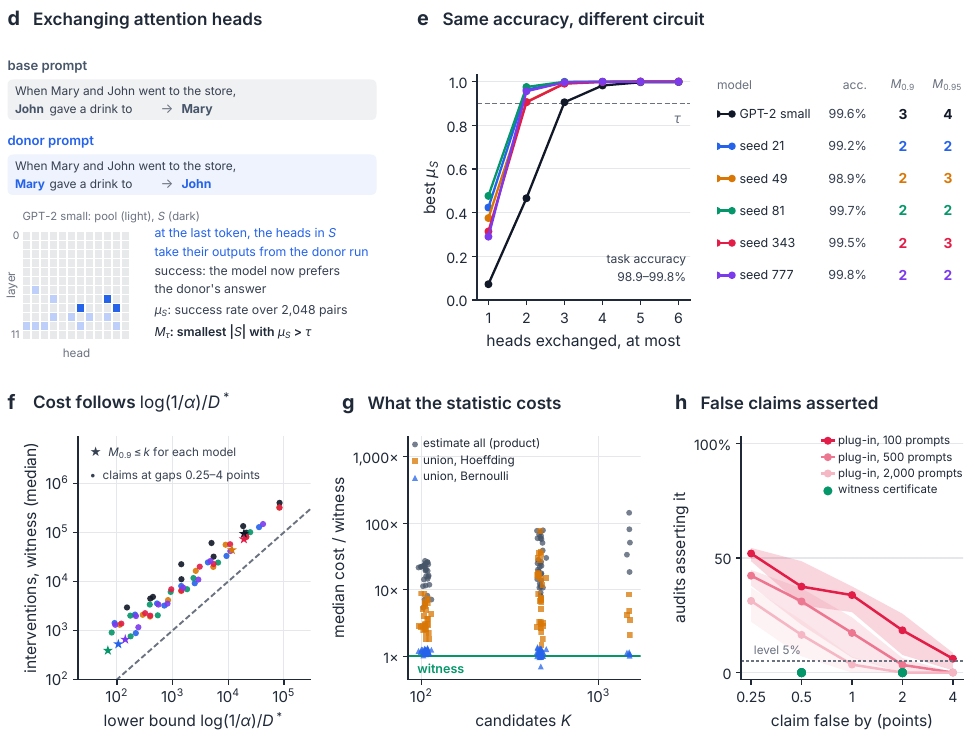}
\caption{\textbf{Circuits: same accuracy, different circuit, one law.}
\p{a}, Interchange intervention in a recurrent network: the activations of a set $S$ of hidden units are replaced by those that a donor prefix induces, and the intervention succeeds if the output is the parity dictated by the donor. $\mu_S$ is the success probability over all 512 contexts and $\Mt$ the size of the smallest $S$ with $\mu_S>\tau$.
\p{b}, $\Mt$ of 30 independently trained networks, all with $100\%$ task accuracy, at $\tau=0.90$ and $\tau=0.95$.
\p{c}, An edge along training: between two checkpoints of one network the best set of at most three units crosses $\tau=0.9$, so $\Mt$ drops from 4 to 3; task accuracy is $100\%$ on both sides of the crossing (63 of 64 sequences at the first checkpoint).
\p{d}, The same intervention on attention heads. Base and donor prompts differ in the repeated name and dictate different answers; at the last token the heads of $S$ take their outputs from the donor run, and the intervention succeeds if the model then prefers the donor's answer. Grid: the 14 candidate heads of GPT-2 small (light) and its smallest sufficient set (dark).
\p{e}, Largest $\mu_S$ over sets of at most $k$ heads, for GPT-2 small and five models trained with different random seeds (exact for 2{,}048 prompt pairs). Right: task accuracy and $\Mt$ at $\tau=0.9$ and $\tau=0.95$.
\p{f}, Median number of interventions of the witness certificate ($\alpha=0.05$) against the lower bound, for the claim $\Mt\le k$ at $\tau=0.9$ in each model (stars; 1{,}000 audits each) and with the threshold placed $0.25$ to $4$ percentage points below the best set of each size (dots; 200 audits each). Every audit ended in a certificate.
\p{g}, Median cost of three other valid certificates relative to the witness certificate, on identical queries, against the number of candidates.
\p{h}, False claims, with the threshold above the best set by the amount on the horizontal axis. Lines: fraction of 2{,}000 plug-in readouts that assert the claim when every candidate is evaluated on the same 100, 500 or 2{,}000 prompts (mean over the six models; bands: range). Dots: fraction of 300 audits in which the witness certificate asserted it.}
\label{fig:circuits}
\end{figure}}

\newcommand{\FigRuns}{%
\begin{figure}[tp]
\centering
\includegraphics[width=\linewidth]{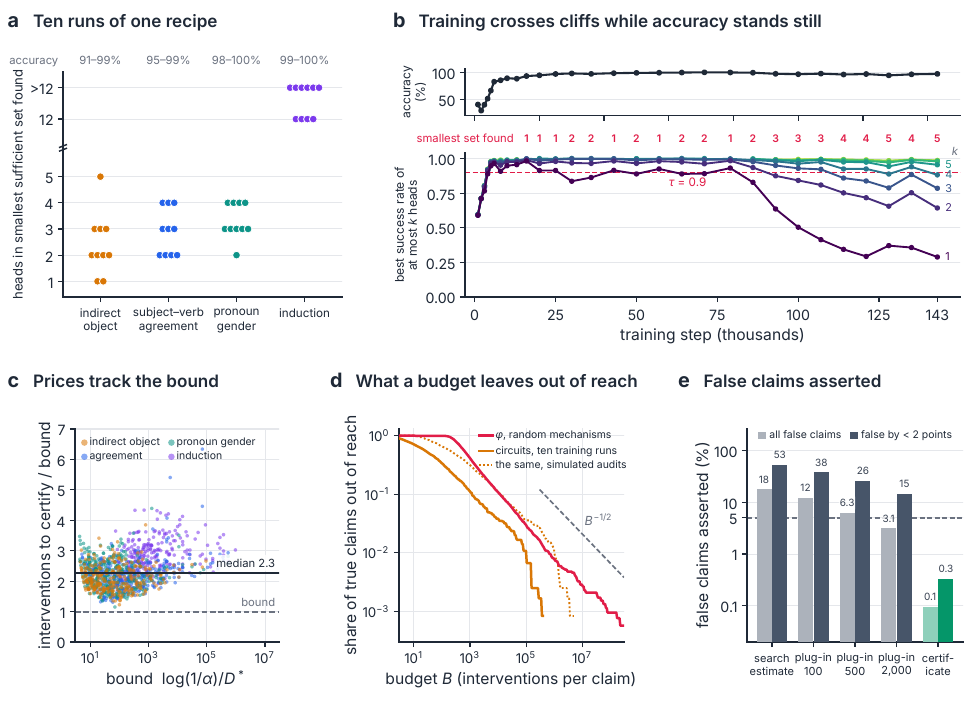}
\caption{\textbf{Ten training runs of a language model, one law.}
\p{a}, Pythia-160m at the end of ten training runs that differ only in random seed: size of the smallest set of attention heads found with success rate above $\tau=0.9$, for four tasks (one dot per run; top: range of task accuracy over the runs). Success rates are exact for a population of 1{,}000 prompt pairs; single heads are evaluated exhaustively, larger sets are proposed by a search.
\p{b}, One run, from step 1{,}000 to the end of training, on indirect-object identification: task accuracy (top) and the largest success rate found over sets of at most $k$ heads (bottom; $k=1$ to 8, labelled up to 5). Numbers: size of the smallest sufficient set found at $\tau=0.9$, at each checkpoint from step 16{,}000.
\p{c}, Median number of interventions of the witness certificate ($\alpha=0.05$; 100 simulated audits per claim), relative to the lower bound, for the 1{,}192 of the 1{,}199 true claims `at most $k$ heads suffice' (all checkpoints and tasks with accuracy of at least $90\%$, six thresholds) that were certified in every audit.
\p{d}, Share of true claims that no certificate of level $\alpha$ can assert with probability one half within a budget of $B$ interventions (\cref{prop:budget}), for $\vphi>1/2$ in 5{,}241 random mechanisms and for the claims of \p{c}. Dotted: share of the claims of \p{c} whose simulated audits needed more than $B$ at the median. Dashed: slope $-1/2$. The curves for circuits end at the smallest gap that 1{,}000 prompt pairs can show.
\p{e}, Claims that one head fewer suffices, false for every proposed set (855 claims; dark: the 179 that are false by at most two points): share asserted by the estimate on the 256 prompt pairs used by the search, and share of repetitions in which they are asserted by a plug-in readout that evaluates every candidate on the same 100, 500 or 2{,}000 sampled prompts and by the certificate (171{,}000 audits of 20{,}000 queries). Dashed line: the $5\%$ allowed to a certificate.}
\label{fig:runs}
\end{figure}}

\newcommand{\FigLive}{%
\begin{figure}[tp]
\centering
\includegraphics[width=\linewidth]{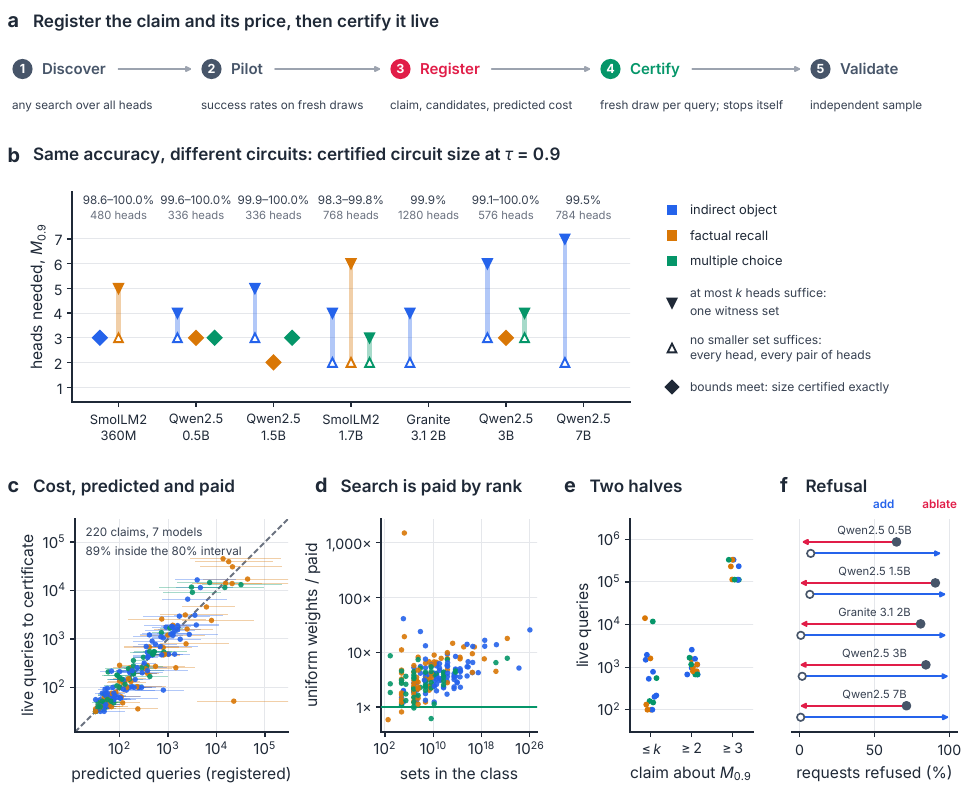}
\caption{\textbf{Prospective audits of language models.}
\p{a}, The protocol. A search proposes sets of attention heads; a pilot estimates their success rates; the claim, its candidate sets and the predicted cost of its certificate are registered; the certificate is then run on live queries, each an independently drawn prompt pair and a real intervention, and stops by itself; the certified set is finally evaluated on an independent validation sample.
\p{b}, Certified circuit size at $\tau=0.9$ for seven instruction-tuned models (the largest run with four-bit weights) and up to three tasks. Filled triangles: certified upper bounds (some set of that many heads succeeds with probability above $0.9$). Open triangles: certified lower bounds over all heads of the model (no single head, or no single head and no pair of heads, reaches $0.9$). Diamonds: the two bounds meet. Top: task accuracy and number of attention heads.
\p{c}, Live queries needed by each of 220 certificates against the cost predicted at registration (median of the prediction; horizontal lines: its 10th to 90th percentiles).
\p{d}, Cost that a certificate with uniform weights over all sets of at most $k$ heads would have needed, relative to the cost paid (line: equal cost), against the number of such sets.
\p{e}, Live queries for the two halves of a statement about circuit size at $\tau=0.9$: the upper bound $\Mt\le k$ (one witness), and the lower bounds $\Mt\ge2$ (every head) and $\Mt\ge3$ (every pair of heads).
\p{f}, Refusal in the five models that refuse most harmful requests. Filled circles: fraction of harmful requests refused; red arrows: the same with the direction certified for ablation removed from the residual stream. Open circles and blue arrows: harmless requests, before and after adding the direction certified for addition.}
\label{fig:live}
\end{figure}}

\noindent\textbf{Whether an artificial neural network is safe, correctly explained or conscious depends on how it computes, not only on what it outputs\cite{geiger2025causal,hubinger2024sleeper,butlin2023consciousness}. Such claims are tested by intervening on internal components\cite{geiger2021causal,wang2023interpretability}, but no theory says which a finite experiment can establish, or at what cost. Here we show that one quantity decides both: the information distance from the mechanism to the nearest mechanism that would falsify the claim. A claim can be certified exactly when this distance is positive, at a price inversely proportional to it, which adaptive experiments approach. Behavior can carry no evidence at all, and quantities that select one thing and report another, such as integrated information\cite{albantakis2023iit4} or a circuit\cite{conmy2023towards}, have cliffs beside which the price diverges whatever the margin. Median certificate costs are within a factor of 20 of this price across networks, including ten training runs of a language model\cite{vanderwal2025polypythias} whose comparably accurate networks implement one task with one, two or at least three attention heads; plug-in estimates from 100 prompts assert near-threshold false claims $38\%$ of the time. In language models of up to 7.6 billion parameters, 220 of 223 audited claims about circuits, and a refusal mechanism\cite{arditi2024refusal}, were certified live, nine in ten within the cost range predicted beforehand. What an experiment can reveal about mechanism can thus be predicted and budgeted before it is run.}

\vspace{6pt}
\noindent Two networks can compute exactly the same function and still differ in how they compute it (\cref{fig:overview}a). For a growing set of questions about artificial intelligence, that difference is the whole point. Mechanistic interpretability asks which algorithm a network implements\cite{geiger2025causal,conmy2023towards,wang2023interpretability}; audits ask whether a hidden mechanism lies behind benign behavior\cite{hubinger2024sleeper,marks2025auditing}; assessments of machine consciousness ask whether a system has the internal organization that scientific theories associate with experience\cite{butlin2023consciousness,chalmers2023could,long2024taking}. Each is a claim about a system's causal mechanism rather than about its input--output behavior. Behavior is weak evidence for such claims, and for consciousness it is compromised outright, because a system trained on human text reproduces the outward signs whether or not anything lies behind them\cite{birch2024edge,perez2023towards}. Serious assessments therefore look inside\cite{butlin2026identifying,findlay2024dissociating}.

\FigOverview

Looking inside is an experiment, and experiments are finite. Machine learning has a mature theory of how well a model predicts. It has no comparable account of when an experiment \emph{determines} how a model computes: whether a given claim can be established at all, how many interventions that takes and with what guarantee. Warnings abound. Explanations can be non-identifiable\cite{meloux2025everything}, causal abstractions too permissive\cite{sutter2025nonlinear}, circuit scores fragile and noisy\cite{miller2024transformer,meloux2025variance} and audits blind to planted mechanisms\cite{casper2024blackbox,goldwasser2022planting}. Hypothesis tests and confidence sequences now put error bars on the score of a given circuit\cite{shi2024hypothesis,asiaee2026certified}. What is missing is a law: which claims an experiment can settle, and what settling them must cost.

Here we provide one. Call the rule by which a system's state produces its next state its mechanism, $\theta$. A quantity such as integrated information or the size of a circuit is a number $F$ computed from $\theta$; a claim says that $F$ exceeds a threshold $c$; an experiment chooses what to set, what to record and where to intervene; and a certificate is a stopping rule that asserts the claim wrongly with probability at most $\alpha$, whatever the mechanism. One quantity then decides what a finite experiment can establish: the information distance $D^\ast$ from the mechanism to the nearest mechanism that falsifies the claim. The claim can be certified exactly when $D^\ast>0$; no certificate that is sure to assert it can use fewer than $\log(1/\alpha)/D^\ast$ trials on average; and, as $\alpha\to0$, an experimenter who knows nothing of the mechanism in advance attains that price (\cref{fig:overview}b). Three things set $D^\ast$: what the experiment touches (\emph{access}), how the measured quantity bends (\emph{geometry}) and what the claim asserts (\emph{logic}).

We establish this law where every answer can be checked against exact ground truth, and then use it where none is available. The first system is the integrated information $\vphi$ that integrated information theory (IIT) identifies with consciousness\cite{tononi2016integrated,albantakis2023iit4}, in stochastic networks of four and five binary units. For a theory of measurement $\vphi$ is the ideal stress test: behavior does not determine it, and it is defined by selecting one quantity and reporting another. Whether IIT is science at all is contested\cite{doerig2019unfolding,iitconcerned2025unscientific}; our question about $\vphi$ is metrological rather than metaphysical---could any experiment tell us its value? The second is the object that mechanistic interpretability studies, the smallest set of components that causally carries a task variable\cite{geiger2021causal}, computed exactly in trained recurrent networks, in six language models of the GPT-2 family\cite{radford2019language,mistral2021} and along ten training runs of a further language model\cite{biderman2023pythia,vanderwal2025polypythias}. The same law governs all of them (\cref{fig:overview}c): behavior can be silent about mechanism, cliffs make the price of evidence depend on distance to the cliff and not on the margin of the claim, prices across networks are heavy-tailed with exponents near the predicted one half, and certificates matched to the claim's logical form cost 4 to 143 times less than estimates of the whole mechanism. We then run the theory forwards. In seven instruction-tuned language models of $0.4$ to $7.6$ billion parameters, claims about circuits and about a safety mechanism were registered together with their predicted cost and then certified on live interventions, nine times in ten within the predicted range (\cref{fig:overview}d).

\section*{Results}

\subsection*{One currency for evidence}
A design offers a finite set of actions. Each action prepares a state by intervention, lets the system update and records chosen variables; under mechanism $\theta$, action $a$ returns an outcome with law $P^a_\theta$, and the next action may depend on everything seen so far. For the claim $F>c$, let $\Hnull$ be the set of mechanisms that falsify it. For a design that spends a fraction $w_a$ of its trials on action $a$,
\begin{equation}\label{eq:Dstar}
D^\ast(\theta)=\max_w\ \inf_{\theta'\in\cl{\Hnull}}\ \sum_a w_a\,\KL\bigl(P^a_\theta\,\|\,P^a_{\theta'}\bigr)
\end{equation}
is the information per trial that separates $\theta$ from the closest falsifying mechanism under the best allocation of trials (\cref{fig:overview}b). We prove five statements (\cref{thm:price0,thm:adaptive,prop:budget}). If $D^\ast(\theta)=0$, every certificate of level $\alpha$ asserts the claim with probability at most $\alpha$, at every sample size. If $D^\ast(\theta)>0$, every certificate that is sure to assert the claim needs at least $\log(1/\alpha)/D^\ast(\theta)$ trials in expectation. The bound is attained, as $\alpha\to0$, by testing the claim itself: assert it the first time
\begin{equation}\label{eq:direct}
E_t=\frac{R_t}{\sup_{\theta'\in\Hnull}L_t(\theta')}\ \ge\ \frac1\alpha ,
\end{equation}
where $L_t(\theta')$ is the likelihood of the first $t$ trials and $R_t$ the probability that a fixed mixture over mechanisms assigns to them: the claim is asserted once the data are $1/\alpha$ times better explained by an average over all mechanisms than by the best falsifying one. The best allocation need not be known: a design that tracks the mechanism estimated so far attains the same price. And a budget of $B$ trials succeeds with probability at most $\{B\,D^\ast(\theta)+\log2\}/\log(1/\alpha)$, whatever the certificate. The guarantees hold for every mechanism and at every time, with no prior, no fitted model and no sample size fixed in advance.

Every obstacle described below is a statement about $D^\ast$: access decides whether it is zero, geometry how it depends on where the mechanism sits, and logic whether a practical statistic attains it.

\subsection*{Access: when behavior is silent}
\FigAccess
The sharpest version of the access problem is a pair of systems that compute the same function. A recurrent network of four noisy parity units and a feed-forward chain have identical input--output laws at their response delay, yet $\vphi=0.28$ bits for the first and $\vphi=0$ for the second (\cref{fig:access}a,b; \cref{prop:samefunction}). For any experiment that sets the input and reads the output at this delay, $D^\ast=0$: no number of such behavioral trials can certify that the recurrent network is integrated. The construction makes the unfolding argument of consciousness science\cite{doerig2019unfolding} exact, and it is general: every recurrent network has a feed-forward twin with $\vphi=0$ and the same input--output law at any chosen horizon (\cref{prop:unfolding}). Behavior can be exactly silent about mechanism without any training to deceive. One internal recording breaks the tie: a single interior unit gives odds of 20:1 in about 17 trials, and all four units in about four (\cref{fig:access}c).

White-box access is rarely complete. When one unit goes unrecorded (\cref{fig:access}d,e), its parameters drop out of every one-step recording, so mechanisms whose $\vphi$ differs by up to $1.13$ bits produce identical data. In the configuration of \cref{fig:access}f, any valid $95\%$ interval must then have an expected width of at least $1.02$ bits, $84\%$ of the largest value a four-unit system can take, at every sample size, and of at least $0.30$ bits in every configuration examined (\cref{fig:access}g; \cref{thm:floor}). With several hidden units the blindness becomes a symmetry: in parity networks, recordings of any length depend on the hidden units only through the product of their fidelities (\cref{thm:product}), which leaves a third of $\vphi$ undetermined. A single intervention on a hidden unit removes the symmetry (\cref{fig:access}h--j).

\subsection*{Geometry: selection creates cliffs}
\FigGeometry
Complete access leaves a second limit, which lies in the quantity itself. To compute $\vphi$ one \emph{selects} the way of cutting the system that does least damage to its cause--effect power per connection cut, and \emph{reports} the total damage of that cut. When a small change in the mechanism hands the selection to a cut that severs a different number of connections, the reported value jumps, by a ratio of small whole numbers (\cref{prop:edges}). Such cliffs are common. Scanning random directions from every state with $\vphi>0$ of 400 random networks, we found 849 cliffs, $99\%$ of which change $\vphi$ by $25$--$100\%$, at sizes quantized exactly as predicted (\cref{fig:geometry}a,b; \cref{fig:edgestats}). We prove that an explicit open set of mechanisms, free of ties and symmetry, splits into a phase with $\vphi<0.273$ bits and a phase with $\vphi>0.870$ bits, and that the split persists when all 64 transition probabilities vary freely (\cref{thm:twophase}; \cref{fig:geometry}c).

Cliffs separate margin from evidence. On the high side of this cliff the claim $\vphi>1/2$ holds with $0.39$ bits to spare however close the mechanism is to the edge, yet the information distance shrinks as the square of the distance $h$ to the edge, $D^\ast=0.594\,h^2$, so the cost of certification grows as $h^{-2}$ (\cref{thm:price}). Simulated certificates cost 8 to 14 times the bound over a 64-fold range of $h$, while the margin never fell below $0.389$ bits (\cref{fig:geometry}d,e). The law is general: beside any smooth edge of any quantity that selects and reports, an experiment must resolve which of two competing candidates wins the selection, so the evidence per trial shrinks as the square of the gap between their scores, and the margin of the reported value plays no part (\cref{thm:edgeprice}). Budgets are therefore heavy-tailed. Over any population of mechanisms with positive density at a cliff, the probability that a claim costs more than $u$ falls only as $u^{-1/2}$ (\cref{thm:tail,cor:tail}): among the 5{,}241 true claims of 6{,}000 random mechanisms, all of which hold by more than $0.12$ bits, the 99th-percentile cost was 880 times the median (\cref{fig:geometry}f,g). Near a cliff a good fit is not a good measurement either: surrogate networks fitted to $2^{18}$ interventional transitions, with a median predictive divergence of $4\times10^{-5}$ nats, still misreported $\vphi$ by more than $20\%$ in $1\%$ of fits (\cref{fig:fitting}).

\subsection*{Logic: test the claim itself}
What survives at a cliff is the one-sided claim. No estimator of $\vphi$ is uniformly accurate near an edge (\cref{thm:envelope}), but the question `is $\vphi$ at least $c$?' has a uniformly valid answer: at a symmetric tie where plug-in estimates erred by $0.80$ bits, lower confidence bounds from 65{,}536 transitions certified $\vphi>0.31$ bits in every dataset, on both sides of the edge (\cref{tab:onesided}). The currency also says how to answer it. Certifying that a confidence set for the whole mechanism lies inside a region where the claim is proved (\cref{thm:region}) is sound but pays to localize every parameter: it needed 7.5 million transitions for a $95\%$ chance of establishing $\vphi>1.004$ bits, and $2.6\times10^{5}$ with the wiring assumed known. Testing the claim itself with \cref{eq:direct}, against a falsifying set that we describe exactly for parity mechanisms (\cref{thm:null}), needed $5.3\times10^{4}$ under the same assumption, and certified mechanisms that the first method never does (\cref{fig:logic}).

\subsection*{Same accuracy, different circuit}
\FigCircuits
Nothing in the theory is specific to integrated information. In mechanistic interpretability the analogous claim is that a small set of components carries a task variable. An interchange intervention\cite{geiger2021causal} replaces the activations of a set $S$ of components by those that a different, donor input induces; its success rate $\mu_S$ is high when $S$ carries the variable, and the circuit size $\Mt$ is the size of the smallest $S$ with $\mu_S>\tau$ (\cref{fig:circuits}a,d). $\Mt$ selects the sets that clear a threshold and reports the size of the smallest, so it has edges (\cref{fig:circuits}c). We computed every $\mu_S$ exactly in two settings: for all sets of hidden units of 30 gated recurrent networks trained on the running parity of binary sequences, and, on 2{,}048 prompt pairs, for all 16{,}384 subsets of the 14 most influential attention heads of GPT-2 small\cite{radford2019language} and of five models of the same architecture that share one training recipe and differ only in random seed\cite{mistral2021}, on indirect-object identification\cite{wang2023interpretability}.

Behavior does not fix the circuit. All 30 recurrent networks reach $100\%$ accuracy, and at $\tau=0.9$ their circuit size is 2 in ten of them, 3 in nineteen and 4 in one (\cref{fig:circuits}b). The six language models are $98.9$ to $99.8\%$ accurate; GPT-2 small needs three heads and each seed model two, and over all 144 heads no single head and no pair suffices in GPT-2 small while exactly one pair does in each seed model (\cref{fig:circuits}e; \cref{fig:lmdetail}; \cref{tab:lm}).

The claim $\Mt\le k$ is existential: among the sets of at most $k$ components, at least one clears $\tau$. Its information distance is that of its best witness, and shrinks as the square of the gap between the best set and the threshold (\cref{prop:witnessrate}). Audits at 13 gaps in a recurrent network and of 66 claims in the language models followed this inverse-square law over nearly eight decades of cost, within a factor of $3.5$ to $19$ of the bound (\cref{fig:circuits}f; \cref{fig:gruaudit}a). They also showed how much the statistic matters. Applied candidate by candidate, \cref{eq:direct} multiplies the evidence of all candidates and so pays for every mean that the claim does not concern. An existential claim needs one witness. A \emph{witness certificate} averages evidence over candidates instead of multiplying it, and pays only the logarithm of their number (\cref{thm:witness}). On identical queries the product needed $4.5$ to $16$ times as many interventions in the recurrent networks and 5 to 143 times as many in the language models, whereas a union bound over Bernoulli confidence sequences, which shares the existential form, stayed within a factor of $0.7$ to $1.5$ (\cref{fig:circuits}g; \cref{tab:baselines,tab:lm}).

Current practice estimates a circuit's accuracy on a batch of prompts and reads the claim off the estimate. On 100 prompts this plug-in readout asserted claims that are false by half a percentage point $38\%$ of the time, and on 2{,}000 prompts $16\%$ of the time (\cref{fig:circuits}h). No certificate exceeded its error level: none asserted such a claim in 3{,}600 audits in the language models, and with every candidate of a recurrent network exactly on the threshold the witness certificate asserted the false claim in $1.4\%$ of 50{,}000 audits, against the $5\%$ allowed (\cref{fig:gruaudit}c).

\subsection*{Ten training runs, one law}
\FigRuns
If cliffs and their price belong to the measurement of mechanism rather than to one network, they should reappear whenever a network is trained again. Pythia-160m, a language model with 12 layers of 12 attention heads\cite{biderman2023pythia}, was released with checkpoints from throughout its training and with nine further runs that differ only in random seed\cite{vanderwal2025polypythias}. For 56 checkpoints of the ten runs and four tasks we computed, on fixed populations of 1{,}000 prompt pairs, the success rate of every single head and of the sets of heads proposed by a search.

The runs end comparably able and differently built (\cref{fig:runs}a). On indirect-object identification all ten are $91$ to $99\%$ accurate, yet the answer is carried by exactly one head in two of them, by exactly two in four and by exactly three in three (no single head and none of the 10{,}296 pairs of heads suffices at $\tau=0.9$), and in another no pair suffices and the smallest sufficient set found has five heads. The best single head carries the answer on $99.4\%$ of prompts in one run and on $23.6\%$ in another, over all 144 heads. For subject--verb agreement and pronoun gender the search finds sufficient sets of two to four heads, and for induction none smaller than twelve.

Training itself crosses the edges: the heads that implement a task are known to turn over during training\cite{tigges2024llm}, and every turnover that changes the size of the smallest set clearing the threshold crosses an edge of $\Mt$. Between successive analyzed checkpoints whose accuracy differed by less than two points, the size of the smallest sufficient set found, at $\tau=0.8$, $0.9$ or $0.95$, changed in 106 of 161 cases (in 33 of them exactly: a single head suffices at one checkpoint and none at the other), and at least once in every run (\cref{fig:runsdetail}). Accuracy barely registers these crossings. In the run that we followed most closely, accuracy on indirect-object identification stayed between $94$ and $100\%$ from step 20{,}000 to the end of training, while the share of prompts on which the best single head carries the answer fell from a peak of $93\%$ to $29\%$, the smallest sufficient set found changed size twelve times between successive checkpoints (seven of them between one head and more, which is exact), from one head to five, and the lower bound on the price of certifying it varied 38-fold (\cref{fig:runs}b).

The prices track the bound. Over the 1{,}199 true claims `at most $k$ heads suffice' of this population, simulated audits cost a median of $2.3$ times the bound (interquartile range $1.9$ to $2.6$) across five decades (\cref{fig:runs}c). Their distribution is heavy-tailed, as the theory predicts: the share of claims whose bound exceeds $u$ falls as $u^{-0.47}$ ($u^{-1/2}$ in theory; $u^{-0.57}$ for $\vphi$ in random mechanisms), down to the smallest gap that a population of 1{,}000 prompt pairs can show, and the costliest $1\%$ of claims carry $54\%$ of the summed bounds. A budget therefore resolves a population slowly: the share of true claims that a budget of $B$ interventions each must leave out of reach falls no faster than $B^{-1/2}$ (\cref{cor:resolution}; \cref{fig:runs}d).

The population also contains 855 claims that one head fewer suffices, false for every set that the search proposed. The certificate asserted them in 163 of 171{,}000 audits ($0.10\%$; at most $2.0\%$ for any one claim, against the $5\%$ allowed). A plug-in readout on 100 sampled prompts asserted them $12\%$ of the time, and $38\%$ of the time when the claim was false by at most two points; on 2{,}000, $3.1\%$ and $15\%$. The search's own estimate, the number that a discovery method would report, asserted 151 of the 855 (\cref{fig:runs}e).

\subsection*{Prospective audits of language models}
\FigLive
An audit of a deployed model has no table of exact answers. We therefore ran the theory forwards, on seven instruction-tuned models of $0.4$ to $7.6$ billion parameters from three families\cite{yang2024qwen25,allal2025smollm2,granite2024granite} and on three tasks: indirect-object identification, recall of a country's capital, and multiple choice in the format of a benchmark\cite{mueller2025mib}, 16 combinations of model and task in all (\cref{tab:live}). For each, a search over every attention head proposed small sets of heads, and a pilot estimated their success rates. Before the first live query we registered the claims `at most $k$ heads suffice at threshold $\tau$' for seven thresholds, the candidate sets of each claim and the predicted cost of its certificate (\cref{fig:live}a). Every live query was then an independently drawn prompt pair and a real intervention.

Of 223 claims audited, 220 were certified, at 32 to $4.5\times10^{4}$ live interventions (\cref{fig:live}b), and every certified set cleared its threshold on an independent validation sample. Taken together, and however the audits depend on one another, the 220 certificates keep the expected share of false ones among them below $5.1\%$, with all 224 registered claims counted\cite{wang2022false}. The cost fell inside the registered $80\%$ interval for $89\%$ of them (\cref{fig:live}c; \cref{fig:livedetail}): the price of a claim about mechanism can be predicted from a pilot before the audit is run. The search and the pilot that proposed the candidates ran on cached activations, $8\times10^{6}$ interventions in all; the certificates, which carry the guarantee, took $3.4\times10^{5}$ live queries. Searching a vast class added almost nothing to that price. The sets of at most $k$ heads number up to $10^{26}$, but evidence about a candidate starts when the candidate is proposed, so a certificate that weights candidates in the order of their proposal is valid whatever the search did, and pays for the rank of the witness in the search rather than for the size of the class (\cref{thm:search}). Uniform weights over the class would have cost a median of $3.4$ times as many queries, and more than ten times as many for 16 claims (\cref{fig:live}d).

The two halves of a statement about circuit size have different prices. `At most $k$ heads suffice' needs one witness. `No smaller set suffices' is universal: its certificate must clear every smaller set, which took about $10^{3}$ interventions for all single heads of a model and $1$ to $3\times10^{5}$ for all of its 56{,}280 to 165{,}600 pairs (\cref{prop:universal}; \cref{fig:live}e). All 118 such claims were certified, and where the two bounds meet, in six of the 16 combinations, the circuit size is certified exactly over every head of the model (each bound at its own level $\alpha$). Equal accuracy again hides different circuits: factual recall, which three models of one family solve with at least $99.9\%$ accuracy, takes exactly two heads in the 1.5-billion-parameter model and exactly three in its 0.5- and 3-billion-parameter siblings (\cref{fig:live}b).

The same audit applies to a safety mechanism. Instruction-tuned models refuse harmful requests (here, a refusal phrase among the first 12 generated tokens), and refusal reportedly hinges on one direction of the residual stream\cite{arditi2024refusal}. `Removing some direction stops the model from refusing more than $90\%$ of harmful requests' is an existential claim over candidate directions. In the five models that refuse most harmful requests it was certified with 120 to 200 live queries; with the certified direction removed the models refused at most $1.4\%$ of 572 harmful requests, down from $65$ to $91\%$. The converse claim was certified as well: adding one such direction made them refuse $94.5$ to $99.8\%$ of harmless requests (\cref{fig:live}f; \cref{tab:refusal}). How often a model refuses says nothing about what its refusals depend on; two hundred live queries certify that removing one direction abolishes them.

\section*{Discussion}
Black boxes, hidden units, cliffs and inefficient statistics look like four problems. They are four statements about one quantity, the information distance from a mechanism to the nearest mechanism that falsifies the claim. Each of its determinants can be computed, or estimated from a pilot, before an experiment is run, which turns the measurability of mechanism into a question of design (\cref{tab:summary}).

The ingredients have precedents. That interventions identify what observation cannot is the foundation of causal inference\cite{pearl2009causality,peters2017elements}; that discontinuous functionals admit only one-sided inference is classical in statistics\cite{donoho1988one,hirano2012impossibility}; that networks trained alike can implement different algorithms, with components that turn over during training, has been observed\cite{zhong2023clock,tigges2024llm}; certifying a property throughout a region is the logic of neural-network verification\cite{katz2017reluplex,cohen2019certified}, now applied to circuits\cite{hadad2026formal}; and anytime-valid tests and the price of identifying the best of finitely many options come from sequential analysis and bandit theory\cite{howard2021time,ramdas2023game,garivier2016optimal,kaufmann2021mixture}, and have been applied to behavioral audits\cite{zhou2026adaptive} and to the fidelity of a given circuit\cite{asiaee2026certified}. Those applications bound one number for one supplied object. What this account adds is the claim about mechanism as the object of measurement: a criterion for which claims can be certified at all; the law that prices them, with cliffs at which the price is unrelated to the margin and a heavy tail whose exponent, one half, is set by the geometry of an edge; certificates for claims over classes of circuits too large to list; and the demonstration that the price can be predicted before the experiment.

Four lessons transfer directly to interpretability and auditing. First, a circuit is an object that is selected and then reported: its size, faithfulness or location is stated after it has been chosen by a thresholded score\cite{conmy2023towards}. It therefore has edges, and ordinary training carries networks across them: runs of one recipe end on different sides, and a single run crosses back and forth while its accuracy stands still. Second, the cost of a circuit claim is set by its gap to the threshold and grows as the inverse square; margins in the reported quantity---one unit of circuit size, half a bit of integration---do not bound it. Budgets for a population of models are therefore dominated by its hardest members, and the share of claims that a budget leaves open shrinks only as its inverse square root. Third, the logical form of a claim is a resource. `Some circuit of at most $k$ components suffices' is existential and is certified by one witness, under any search procedure, at a price that does not depend on how many circuits there are; `no smaller circuit suffices' is universal, and its price is the sum of the prices of all smaller circuits. Fourth, the price can be posted in advance. A pilot fixes the information distance well enough to predict what a certificate will cost, so a claim about mechanism can be registered with its budget before it is tested, as clinical trials are. Open weights make all of this cheaper, not unnecessary: prompts are too many to enumerate, so even a fully observed model is measured through finite samples of interventions.

For consciousness science, the same-function pair shows what makes claims about implementation empirical. Behavior at the response delay cannot separate the two systems, but one internal recording can: every equivalence we constructed is tied to a delay or a recording, and another recording or intervention breaks it. The unfolding argument therefore shows that behavior alone cannot decide between implementations, not that implementations cannot be told apart. For IIT, which adversarial collaborations now confront with neural data\cite{cogitate2025adversarial}, the results turn a philosophical dispute into requirements on experiments: access to the declared units at the declared grain, interventions wherever units are hidden, one-sided claims in place of point values near an edge, and budgets planned from the distance to the nearest cliff. The same requirements apply to architectural indicators such as recurrence or a global workspace\cite{butlin2023consciousness}, which behavior need not reveal, and to perturbation experiments on neural circuits, where a trial is an animal rather than a forward pass and the price of a claim decides whether it is worth asking.

The certificates ask of a candidate only a stream of outcomes, so nothing in them is specific to attention heads: learned subspaces, sparse features and the output of automated circuit discovery qualify as candidates.

Whether a claim about mechanism can be established is not settled by how much data an experiment collects. It is settled by what the experiment touches, by the geometry of the measured quantity and by the logic of the claim, and one number prices all three. Behavior underdetermines mechanism, but targeted access and certificates matched to the claim establish what finite data can---in machines and, by the same arithmetic, in brains.

\section*{Methods}

\subsection*{Systems and integrated information}
We study networks of $n$ binary units that update synchronously and, given the previous global state, independently, so that a mechanism is a transition law $K_\theta(y\mid x)=\prod_iK_{\theta,i}(y_i\mid x)$. System integrated information $\vphi$ follows IIT~4.0\cite{albantakis2023iit4} as implemented in the reference software PyPhi\cite{mayner2018pyphi}. At a current state, the cause and effect states that the current state constrains most strongly are specified; for every directed cut of the system (150 for four units, 1{,}061 for five) the loss $L_C$ of cause--effect power and the number $e_C$ of severed connections are computed; the cut with the smallest $L_C/e_C$ is selected, with ties broken toward the larger raw loss; and the reported value is the raw loss of the selected cut, capped by the intrinsic information of the specified states (\cref{app:phi}). The comparison quantity $\vphiraw$ selects and reports the same raw loss; it is not part of IIT and isolates the effect of the selection rule. In a parity network each unit fires when an odd number of its inputs is active and computes this parity correctly with reliability $p_i$; we use the fidelity $a_i=2p_i-1$ and the log-reliability $\lambda_i=\log_2(2p_i)$. Our vectorized implementation of the general recipe reproduces closed forms for parity networks to $10^{-15}$ and, to nine decimals, seven reference values from an independent exact-rational implementation (\cref{app:validation}). All logarithms in $\vphi$ are to base two (bits); information distances are in nats.

\subsection*{Designs, claims and certificates}
A design is a finite set of actions; an action prepares a state by intervention, lets the system update and records chosen units at chosen delays, possibly through a sensor that flips all recorded bits of a trial together with a probability estimated from calibration trials. A certificate of level $\alpha$ for the claim $F>c$ is a rule that may stop after any trial and assert the claim, and that asserts it with probability at most $\alpha$ under every mechanism with $F\le c$, for any adaptive choice of actions and any duration (\cref{def:cert}). \Cref{thm:price0} is proved by a change-of-measure bound from best-arm identification\cite{garivier2016optimal,kaufmann2016complexity}, universal inference for composite nulls\cite{wasserman2020universal} and Ville's inequality for nonnegative martingales\cite{howard2021time,ramdas2023game}; the design of \cref{thm:adaptive} follows the tracking principle of that literature\cite{garivier2016optimal,degenne2019pure}. The same change of measure bounds what a budget buys: a certificate asserts a true claim within $B$ trials with probability $\pi$ only if $\kl(\pi,\alpha)\le B\,D^\ast$ (\cref{prop:budget}), so over a population with positive density at an edge the share of claims left unresolved falls no faster than $B^{-1/2}$ (\cref{cor:resolution}). When actions differ in cost, all statements hold with the information distance per unit of cost (\cref{eq:costs}). \Cref{tab:novelty} separates the classical ingredients from the new results. Unless stated otherwise $\alpha=0.05$.

\subsection*{Access}
The same-function pair consists of the recurrent parity network in which each unit reads the other three, with reliabilities $(0.78,0.80,0.83,0.85)$, and a feed-forward chain whose first unit is a fair coin and whose later units copy their predecessor with reliability $0.697$; the input is unit~1 at time $t$ and the output is unit~1 of the network (unit~3 of the chain) at $t+2$. Evidence per trial is the expected log-likelihood ratio between the two systems under balanced interventions and a sensor error of $0.10$. Recording designs, budgets and interval constructions for the hidden-unit analyses are given in \cref{app:designs}; intervals are simultaneous $95\%$ confidence sets built from exact Clopper--Pearson intervals\cite{clopper1934use} and transported through the closed form for $\vphi$. Identification floors are $(1-2\alpha)$ times the range of $\vphi$ over mechanisms that produce identical data under every action of the design (\cref{thm:floor}). For the product symmetry, recorded trajectory laws of five-unit parity networks were computed exactly by forward recursion over all recorded histories. Analyses of calibration, inference procedures and model checking are reported in \cref{fig:pipeline} and \cref{app:joint,app:calibration,app:logistic}.

\subsection*{Geometry}
\emph{Edge prevalence.} We drew 200 excitatory and 200 mixed-sign stochastic logistic networks of four units, and 60 of each with five units. From every state with $\vphi>0$ we scanned random directions through parameter space over a distance of one on either side and confirmed every discontinuity by 45 bisections (\cref{app:new}). \emph{Two-phase region.} The network has input masks $(6,12,15,7)$: unit~1 reads units 2 and 3, unit~2 reads units 3 and 4, unit~3 reads all four units and unit~4 reads units 1--3. The region is the cube of log-reliabilities within $1/200$ of $\lambda^0=(0.68,0.83,0.60,0.72)$, split by $g(\lambda)=7\lambda_3-3\lambda_1-3\lambda_4$ (\cref{thm:twophase}). On 40{,}000 mechanisms drawn from the cube the exact engine gave values below $0.2716$ bits where $g<0$ and above $0.8722$ bits where $g>0$; on 60{,}000 mechanisms with 64 free probabilities within $0.004$ of $\lambda^0$ every value was below $0.270$ or above $0.874$ bits. \emph{Price at the cliff.} Information distances to the closed null were computed by numerical projection onto the seven pieces of \cref{eq:null}, and costs by running the direct certificate on simulated transitions at seven distances $h$ from $0.04$ to $0.000625$ (300 runs each; \cref{app:directexp}). \emph{Cost population.} We drew 6{,}000 mechanisms with reliabilities uniform on $[0.76,0.86]$ and computed $D^\ast$ for every true claim $\vphi>1/2$ and, for comparison, $\vphiraw>0.2714$ bits (its population median). \emph{Surrogates.} For 100 random networks per family we fitted logistic networks of the correct form by maximum likelihood to $2^{10}$--$2^{18}$ interventional transitions, 64 times each, and computed the plug-in $\vphi$ (\cref{app:new}).

\subsection*{Certificates for integrated information}
One-sided bounds at the tie use simultaneous Clopper--Pearson intervals for the four reliabilities and the monotonicity of $\vphiraw\le\vphi$ (\cref{app:new}). The region theorem (\cref{thm:region}) is proved with rational enclosures of all logarithms; certification by containment asserts the claim when a simultaneous confidence set for the transition probabilities lies inside the region, and its power is an exact product of binomial probabilities. The direct certificate evaluates \cref{eq:direct} with a product of uniform mixtures over the four reliabilities and an upper bound on the null supremum obtained by branch and bound over one scalar per branch of \cref{eq:null}; each bound is a dual value evaluated in 60-digit interval arithmetic, so an unresolved search can only abstain (\cref{app:direct}). Runs evaluate the certificate on a geometric grid of sample sizes (ratio $1.08$), which preserves its level.

\subsection*{Trained recurrent networks}
Gated recurrent networks\cite{cho2014learning} with six hidden units read binary sequences of length six and were trained on the running parity of all prefixes (cross-entropy, Adam\cite{kingma2015adam}, 2{,}000 steps), 30 networks in the main study and ten for replication; all reach $100\%$ accuracy. A context is a triple of three-bit strings (base prefix, donor prefix, suffix). The network reads both prefixes; the hidden units in $S$ take the donor's values and the others the base's; the network then reads the suffix, and the intervention succeeds if the output sampled from the softmax equals the parity dictated by the donor prefix and the suffix. $\mu_S$ is the exact success probability over the 512 contexts. A query names a candidate and returns the outcome of one intervention at a random context; audits were simulated exactly from the tabulated $\mu_S$. Edges were located on linear interpolations between consecutive training checkpoints; the audited configurations lie on one such path at gaps $\Delta=0.04\times2^{-j}$, $j=0,\dots,12$ (a single candidate exceeds $\tau$ at the eleven gaps $\Delta\le0.01$; three and two do at $\Delta=0.04$ and $0.02$), and on the first isolated crossing of three replication networks (\cref{app:gru}).

\subsection*{Certificates for existential claims}
For $K$ candidates with success counts $s_S$ in $N_S$ queries, let $U_S$ be a uniform mixture of Bernoulli likelihoods divided by the largest likelihood with mean at most $\tau$, and let $\Lambda_{S,\ell}$ be the likelihood ratio of mean $q_\ell=\tau+(1-\tau)2^{-\ell}$ against $\tau$. The witness certificate asserts the claim when
\begin{equation}\label{eq:witnessmain}
E^{\mathrm W}=\frac1{2K}\sum_{S}\Bigl\{U_S+\sum_{\ell\ge1}\frac{\Lambda_{S,\ell}}{\ell(\ell+1)}\Bigr\}\ \ge\ \frac1\alpha .
\end{equation}
It is valid for any adaptive choice of queries, and a witness at gap $\Delta$ is certified within $O(\Delta^{-2}\{\log(K/\alpha)+\log\log(1/\Delta)\})$ of its own queries (\cref{thm:witness}). We compared it, on identical sequences of queries, with four other valid statistics (\cref{app:witness}): the product $\prod_SU_S$, which is \cref{eq:direct} for the whole table of candidates; a product of one-sided mixtures, the product form of multi-armed testing by betting\cite{sandoval2026multiarmed}; and union bounds over candidates of Bernoulli (Beta-mixture) and sub-Gaussian (Hoeffding) confidence sequences\cite{howard2021time}. The sampling rule sees counts only: one query of every candidate; then any candidate with fewer than $\sqrt t/K$ queries; otherwise the empirically best candidate if its mean exceeds $\tau$, else the candidate with the largest index $\hat\mu_S+\sqrt{2\log(t+1)/N_S}$. Queries are issued in batches of $5\%$ of the queried candidate's count, so stopping times are resolved to about $5\%$. The plug-in readout evaluates every candidate on $n$ queries and asserts the claim if any empirical accuracy exceeds $\tau$. When candidates are proposed by a search and cannot be enumerated, candidate $j$ receives a weight $\omega_j$ when it is proposed, with $\sum_j\omega_j\le1$, and its evidence is computed from the queries made after its proposal; the weighted sum of these evidences is a certificate for any search (\cref{thm:search}), in the spirit of online testing\cite{foster2008alpha,xu2024online}. A universal claim, that every candidate of a list has $\mu_S<\tau$, is asserted when the evidence of every candidate against $\mu_S\ge\tau$ has reached $1/\alpha$, which needs no correction for the number of candidates (\cref{prop:universal}).

\subsection*{Language models with exact tables}
We used GPT-2 small as released by OpenAI\cite{radford2019language} and the five GPT-2-small models of the Mistral project\cite{mistral2021}, which were trained on OpenWebText with one recipe and different random seeds (and which rescale the attention logits of each layer, as our forward pass does). The task is indirect-object identification\cite{wang2023interpretability}: in prompts such as `When Mary and John went to the store, John gave a drink to', the correct continuation is the name that is not repeated. A prompt pair consists of a base prompt and a donor prompt that differ only in the repeated name and therefore dictate different answers (12 templates, 149 names). An interchange intervention on a set $S$ of attention heads replaces the outputs of those heads at the last token position by their outputs on the donor prompt; it succeeds if the model then prefers the donor's answer to the base's. For each model, the 14 heads with the largest individual effect on a discovery set of 1{,}024 pairs form the candidate pool, and the outcome of the intervention was computed for all $2^{14}$ subsets of the pool on each of 2{,}048 independently drawn pairs, which gives every $\mu_S$ exactly for this population. Every single head and every pair of all 144 heads was evaluated in the same way, which makes the circuit sizes at $\tau=0.9$ exact over all heads. A query names a set $S$ and returns the outcome on a random pair of the population; the candidates of the claim $\Mt\le k$ are the subsets of at most $k$ pool heads (106, 470 and 1{,}471 candidates for $k=2,3,4$). Our forward pass agrees with the reference implementation\cite{wolf2020transformers,paszke2019pytorch} to $5\times10^{-13}$ in double precision. For one seed model we repeated the computation, for sets of at most four heads of its final pool, at eight checkpoints of its training. Details are given in \cref{app:lm}.

\subsection*{Ten training runs of a language model}
Pythia-160m\cite{biderman2023pythia} is a transformer language model with 12 layers of 12 attention heads and 162 million parameters, trained for 143{,}000 steps and released with 154 checkpoints; nine further runs repeat the training with other random seeds for the initial weights and the order of the data\cite{vanderwal2025polypythias}. We analyzed 56 checkpoints: the last step of all ten runs, 28 earlier checkpoints of the original run from step 1{,}000 on, and steps 57{,}000 and 100{,}000 of each further run. Each of four tasks has a fixed population of 1{,}000 pairs of a base and a donor prompt with the same number of tokens and the same final token and different one-token answers: indirect-object identification (as above), subject--verb agreement (the verb after a subject and an intervening noun; `is' or `are'), pronoun gender (`she' or `he' after a clause about a named person) and induction (the continuation of a pair of random word tokens that has occurred twice). Accuracy is the fraction of prompts on which the model prefers the prompt's own answer to its partner's. For every checkpoint and task we computed, on every pair, the outcome of exchanging each of the 144 heads alone at the final token; ran a beam search of width 4 for sets of up to 12 heads, over the 24 heads with the largest mean effect, on the first 256 pairs; and evaluated the three best sets of every size on all 1{,}000 pairs. Success rates over the population are therefore exact up to the half-precision arithmetic of the forward pass (the same intervention evaluated twice can differ on one pair in 1{,}000), a query of an audit draws a pair uniformly from the population, and audits are simulated exactly from the tables. The circuit size of a checkpoint is the size of the smallest evaluated set with $\mu_S>\tau$; statements about single heads are exact over all 144. For every checkpoint and task with accuracy of at least $90\%$ and every $\tau\in\{0.5,0.6,0.7,0.8,0.9,0.95\}$ we audited the true claim `at most $k$ heads suffice' for the smallest such $k$, with up to six candidate sets weighted $1/\{j(j+1)\}$ in the order of their success on the 256 search pairs (100 audits), and, where $k\ge2$, the claim that $k-1$ heads suffice, which is false for every proposed set (200 audits of 20{,}000 queries; plug-in readouts repeated 1{,}000 times on prompts drawn with replacement). Tail exponents are least-squares slopes of the logarithm of the empirical survival function of $1/D^\ast$ between its 50th and 98th percentiles. Details are given in \cref{app:runs}.

\subsection*{Prospective audits}
We audited seven instruction-tuned models of the Qwen2.5\cite{yang2024qwen25} ($0.5$, $1.5$, 3 and $7.6$ billion parameters), SmolLM2\cite{allal2025smollm2} ($0.36$ and $1.7$ billion) and Granite~3.1\cite{granite2024granite} ($2.5$ billion) families, run in half precision, the largest with four-bit weights; the circuit tasks use plain-text prompts and the refusal audit each model's chat template. The 16 audited combinations of model and task are the three tasks for the Qwen2.5 models of up to 3 billion parameters and for SmolLM2 (two tasks for its smaller model), and indirect-object identification for Granite and for the largest model; task accuracy, the fraction of prompts on which the answer is preferred to the alternative, is at least $98.2\%$ in all of them. The tasks are indirect-object identification, recall of the capital of a country, and a four-option multiple-choice task in the format of a benchmark\cite{mueller2025mib}; in each, a base prompt and a donor prompt of equal length dictate different one-token answers, and an intervention exchanges attention heads at the final token as above. For each model and task we (1)~measured accuracy on 2{,}000 pairs; (2)~measured, on 256 pairs, the effect of exchanging each head of the model alone, and ran a beam search of width 6 over the 32 heads with the largest effect for sets of up to 16 heads; (3)~evaluated the three best sets of each size on 4{,}000 further pairs; (4)~registered, for $\tau\in\{0.5,0.6,0.7,0.8,0.9,0.95,0.98\}$, the claims $\Mt\le k$ for the smallest $k$ whose best pilot rate exceeds $\tau+0.002$ and for $k+1$, each with up to six candidate sets weighted $1/\{j(j+1)\}$ in the order of their pilot rates and with a predicted cost (median and 10th to 90th percentiles of the certificate's stopping time in 400 simulated audits with success probabilities drawn from the pilot posteriors), in a file written before the first live query; (5)~ran live every certificate whose predicted median cost was within the budget of its model (30{,}000 to 200{,}000 queries), with the statistic $\sum_j\omega_jE_j$ of \cref{thm:search}, one independently drawn prompt pair per query and batches of at least four; (6)~certified, over all heads of the model, that no single head reaches $\tau$ (where $k\ge2$) and, at $\tau=0.9$ in models with at most 200{,}000 pairs of heads, that no pair of heads does (where $k\ge3$; \cref{prop:universal}); and (7)~evaluated every certified set on 20{,}000 independently drawn pairs (10{,}000 for models above a billion parameters; for the largest model, 5{,}000, with a search over 24 heads and a pilot of 2{,}000 pairs). The refusal audit follows the same steps with candidate directions in place of sets of heads: the differences between the mean residual-stream activations on harmful and on harmless instructions\cite{arditi2024refusal} at each layer and each of the last five positions of the prompt. A response is scored as a refusal when its first 12 generated tokens contain a refusal phrase, and only that binary outcome is stored. Discovery and pilot used $6.5\times10^{6}$ and $1.5\times10^{6}$ interventions on cached activations; the 220 live certificates used $3.4\times10^{5}$ queries. The intervention engine reproduces ordinary forward passes to $10^{-13}$ in double precision and agrees with the independent implementation used for the GPT-2 tables to $1.2\times10^{-5}$. Details are given in \cref{app:live,app:refusal}.

\subsection*{Statistics and reproducibility}
Every reported probability of certification, coverage or error is a count over independent simulated datasets or audits whose number is stated with the result; intervals for ratios of medians are $95\%$ paired bootstrap intervals (2{,}000 resamples). Exact quantities ($\vphi$, data laws, information distances, interchange accuracies) are computed without sampling. Audits of the ten training runs were simulated exactly from tables of outcomes on fixed populations of prompt pairs. The live audits of language models were each run once, and their outcomes are reported for every claim registered in the 16 audited combinations of model and task; each certificate has level $\alpha$ for its own claim. Applied to the final evidence of all 224 registered claims (with evidence one for the claim that was not run), the e-BH procedure\cite{wang2022false} retains the 220 certified claims at a false discovery rate of $0.051$ under arbitrary dependence. Random seeds are fixed in the code.

\section*{Data availability}
The language models are publicly available from the Hugging Face hub (\texttt{openai-community/gpt2}, the five \texttt{stanford-crfm/*-gpt2-small-*} models, \texttt{EleutherAI/pythia-160m} with its nine \texttt{-seed} runs and the instruction-tuned models listed in \cref{app:live}). The instruction sets of the refusal audit are those released with ref.~\citenum{arditi2024refusal}. All other data are generated by the code or enter it as tabulated values, which its documentation lists. The exact candidate tables of the audited recurrent networks and language models, and the outputs of all experiments, are provided with the code.

\section*{Code availability}
Code for the exact computation of $\vphi$, for the certificates, for the recurrent-network and language-model studies and for all figures is provided with the manuscript as supplementary software, together with the exact candidate tables, and will be deposited in a public repository upon publication. Its documentation lists the simulation results that enter the scripts as tabulated values.

\bibliographystyle{naturemag}
\bibliography{references}

% ============================================================================
% Extended Data
% ============================================================================
\clearpage
\FloatBarrier
\setcounter{figure}{0}
\setcounter{table}{0}
\renewcommand{\figurename}{Extended Data Fig.}
\renewcommand{\tablename}{Extended Data Table}
\renewcommand{\theHfigure}{ED.\arabic{figure}}
\renewcommand{\theHtable}{ED.\arabic{table}}
\crefalias{figure}{edfigure}
\crefalias{table}{edtable}
\section*{Extended Data}

\begin{table}[h!]\centering\small
\caption{\textbf{Prospective audits of instruction-tuned language models.} For each model and task: parameters, attention heads, accuracy on 2{,}000 prompts (the answer preferred to the alternative); the certified range of the circuit size $\Mt$ (a single number where the certified upper and lower bounds meet); for the claim $\Mt\le k$ at $\tau=0.9$ with the smallest registered $k$, the number of live queries to the certificate, the registered prediction (median) and the success rate of the certified set on the validation sample; the number of live queries for the certificates that no single head and no pair of heads has $\mu_S\ge0.9$, over all heads of the model; and the number of registered claims that were certified. Bounds certified at one threshold are carried to the thresholds they imply. Each certificate has level $\alpha=0.05$.}\label{tab:live}
\setlength{\tabcolsep}{3pt}
\resizebox{\linewidth}{!}{%
\begin{tabular}{@{}llrrrcccrrrrrc@{}}
\toprule
 & & & & & \multicolumn{3}{c}{certified $\Mt$} & \multicolumn{3}{c}{$\Mt\le k$, $\tau=0.9$} & \multicolumn{2}{c}{lower bound, $\tau=0.9$} & \\
\cmidrule(lr){6-8}\cmidrule(lr){9-11}\cmidrule(lr){12-13}
model & task & params ($10^9$) & heads & acc.\ (\%) & $\tau=0.5$ & $0.9$ & $0.95$ & queries & predicted & $\mu_S$ validation & singles & pairs & certified \\
\midrule
SmolLM2 360M & indirect object & 0.36 & 480 & 100.0 & 2 & 3 & 3 & 216 & 160 & 0.981 & 960 & 233{,}210 & 13/14 \\
SmolLM2 360M & factual recall & 0.36 & 480 & 98.6 & 2--3 & 3--5 & 3--7 & 14{,}093 & 15{,}414 & 0.907 & 960 & 230{,}944 & 12/14 \\
Qwen2.5 0.5B & indirect object & 0.49 & 336 & 99.6 & 2 & 3--4 & 3--4 & 201 & 162 & 0.972 & 672 & 112{,}688 & 14/14 \\
Qwen2.5 0.5B & factual recall & 0.49 & 336 & 100.0 & 2 & 3 & 3 & 100 & 240 & 0.973 & 800 & 114{,}148 & 14/14 \\
Qwen2.5 0.5B & multiple choice & 0.49 & 336 & 99.6 & 2 & 3 & 3 & 160 & 100 & 0.990 & 672 & 112{,}560 & 14/14 \\
Qwen2.5 1.5B & indirect object & 1.54 & 336 & 99.9 & 2--4 & 3--5 & 3--6 & 532 & 374 & 0.940 & 672 & 112{,}560 & 14/14 \\
Qwen2.5 1.5B & factual recall & 1.54 & 336 & 100.0 & 1 & 2 & 2 & 132 & 172 & 0.974 & 800 & -- & 14/14 \\
Qwen2.5 1.5B & multiple choice & 1.54 & 336 & 100.0 & 2 & 3 & 3 & 148 & 160 & 0.976 & 672 & 113{,}072 & 14/14 \\
SmolLM2 1.7B & indirect object & 1.71 & 768 & 99.8 & 2--3 & 2--4 & 2--5 & 1{,}489 & 1{,}380 & 0.930 & 1{,}536 & -- & 13/14 \\
SmolLM2 1.7B & factual recall & 1.71 & 768 & 99.8 & 2--4 & 2--6 & 2--7 & 1{,}604 & 2{,}479 & 0.923 & 1{,}536 & -- & 14/14 \\
SmolLM2 1.7B & multiple choice & 1.71 & 768 & 98.3 & 1 & 2--3 & 2--4 & 552 & 418 & 0.938 & 1{,}664 & -- & 14/14 \\
Granite 3.1 2B & indirect object & 2.53 & 1280 & 99.9 & 2--3 & 2--4 & 2--5 & 1{,}944 & 3{,}836 & 0.923 & 2{,}560 & -- & 14/14 \\
Qwen2.5 3B & indirect object & 3.09 & 576 & 99.1 & 2--4 & 3--6 & 3--6 & 100 & 200 & 0.979 & 1{,}152 & 331{,}200 & 14/14 \\
Qwen2.5 3B & factual recall & 3.09 & 576 & 100.0 & 2 & 3 & 3 & 100 & 140 & 0.983 & 1{,}152 & 331{,}456 & 14/14 \\
Qwen2.5 3B & multiple choice & 3.09 & 576 & 100.0 & 2--3 & 3--4 & 3--5 & 11{,}794 & 3{,}053 & 0.925 & 1{,}152 & 331{,}200 & 14/14 \\
Qwen2.5 7B & indirect object & 7.62 & 784 & 99.5 & 2--4 & 2--7 & 2--8 & 777 & 550 & 0.938 & 1{,}568 & -- & 14/14 \\
\bottomrule
\end{tabular}}
\end{table}

\begin{table}[h!]
\centering\small
\caption{\textbf{Access, geometry and logic.} What limits a claim about mechanism, what does not help, and what does.}
\label{tab:summary}
\renewcommand{\arraystretch}{1.25}
\begin{tabularx}{\linewidth}{@{}>{\raggedright\arraybackslash}p{0.13\linewidth}>{\raggedright\arraybackslash}X>{\raggedright\arraybackslash}X>{\raggedright\arraybackslash}X@{}}
\toprule
 & \textsf{What limits the claim} & \textsf{What does not help} & \textsf{What helps}\\
\midrule
\textsf{\textbf{Access:} black boxes} & behaviorally equivalent mechanisms differ in the quantity ($D^\ast=0$) & more input--output trials & recording internal units: odds of 20:1 in 4--53 trials\\
\textsf{\textbf{Access:} hidden units} & marginalization; only the product of hidden fidelities reaches the recordings & more trials; longer recordings of the same units & recording more units; later readouts; one intervention on a hidden unit\\
\textsf{\textbf{Geometry:} cliffs} & quantities that select and report jump; cost is set by distance to the cliff, not by margin & point estimates; plug-in values from well-fitting models; larger margins & maps of the cliffs; structural knowledge; budgeting by information distance\\
\textsf{\textbf{Logic:} certificates} & statistics that estimate more than the claim asserts & confidence sets for the whole mechanism & one-sided claims; testing the claim directly; witness certificates for existential claims\\
\bottomrule
\end{tabularx}
\end{table}

\begin{figure}[h!]
\centering
\includegraphics[width=\linewidth,trim=0 0 0 2.50in,clip]{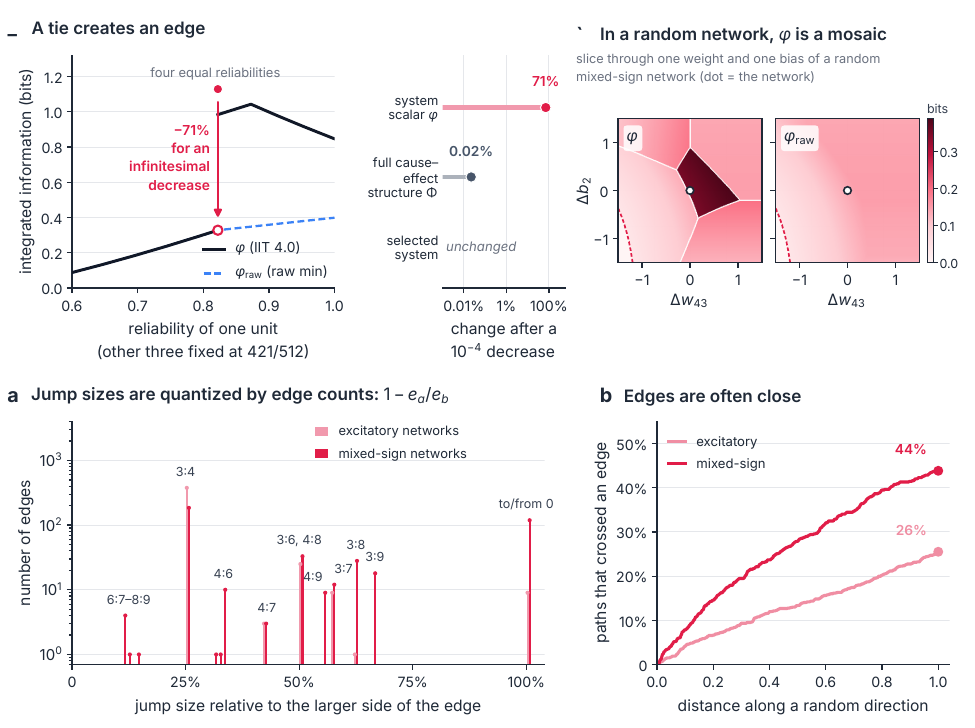}
\caption{\textbf{Cliffs are common in random networks.}
\p{a}, Sizes of 849 discontinuities found on 1{,}976 random parameter paths through the 988 states with $\vphi>0$ of 317 random networks, relative to the larger side, for excitatory (light) and mixed-sign (dark) networks. Partition switches sit exactly at $1-e_a/e_b$ for the numbers of connections $e_a<e_b$ severed by the competing cuts; jumps to or from zero come from switches of the specified states.
\p{b}, Fraction of random paths (two per state with $\vphi>0$) that cross an edge within a given parameter distance of the state.}
\label{fig:edgestats}
\end{figure}

\begin{figure}[h!]
\centering
\includegraphics[width=\linewidth]{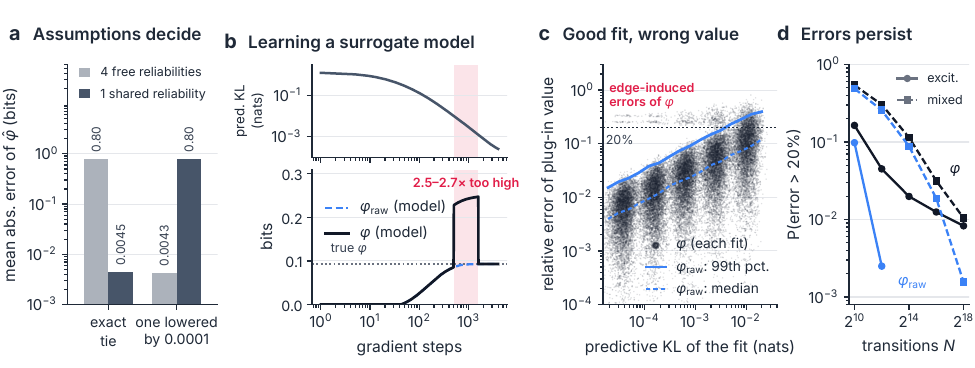}
\caption{\textbf{A good fit is not a good measurement.}
\p{a}, Mean absolute error of $\vphi$ estimated from 65{,}536 complete transitions of the parity network, assuming four free reliabilities or one shared reliability, when all four are equal (left) and when one is lowered by $10^{-4}$ (right); 256 datasets each. The structural assumption, not the data, picks the branch.
\p{b}, Maximum-likelihood training of a surrogate network on $2^{16}$ interventional transitions from a mixed-sign target within $0.03$ of an edge. Top: predictive divergence from the target. Bottom: $\vphi$ (solid) and $\vphiraw$ (dashed) of the surrogate and the true $\vphi$ (dotted); in the shaded window the surrogate's $\vphi$ is $2.5$--$2.7$ times the true value.
\p{c}, Relative error of the plug-in values against the fit's predictive divergence, for 100 random excitatory networks, five sample sizes and 64 fits each. Dots: plug-in $\vphi$ of each fit; lines: median and 99th percentile of the relative error of the plug-in $\vphiraw$.
\p{d}, Probability of a relative error above $20\%$ against the number of transitions, for $\vphi$ (dark) and $\vphiraw$ (blue) in excitatory (solid) and mixed-sign (dashed) networks.}
\label{fig:fitting}
\end{figure}

\begin{figure}[h!]
\centering
\includegraphics[width=0.95\linewidth]{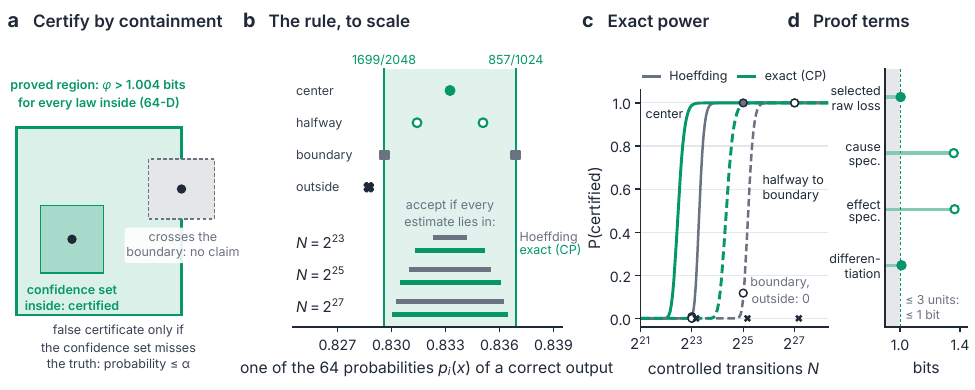}\\[4pt]
\includegraphics[width=0.95\linewidth]{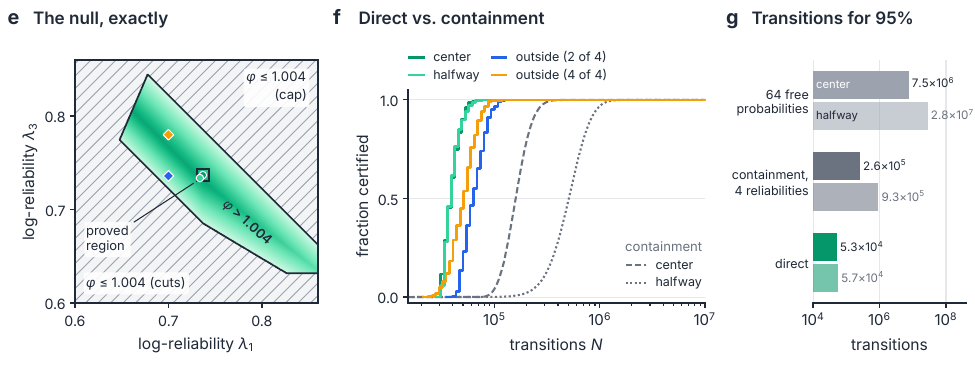}
\caption{\textbf{Certificates of a claim about integrated information.}
\p{a}, A theorem guarantees $\vphi>1.004$ bits for every transition law in a 64-dimensional region. If a simultaneous confidence set lies inside the region the claim is certified; otherwise no claim is made.
\p{b}, The region and the rule to scale for one of the 64 probabilities: four test sources (top) and the ranges that every estimate must fall in, with Hoeffding (grey) and exact Clopper--Pearson (green) intervals.
\p{c}, Exact probability of certification by containment against the number of transitions, for sources at the center of the region (solid) and halfway to its boundary (dashed). Circles: Monte Carlo runs; crosses: boundary and outside sources, never certified.
\p{d}, The four uniform lower bounds in the proof; the smallest sets the certified value.
\p{e}, A slice of the parity family. Shading: $\vphi$. Hatched: the falsifying set $\vphi\le1.004$ bits, assembled from one cap constraint and six branches. Square: the proved region; markers: test sources inside (circles) and outside it (diamonds, projected onto the slice).
\p{f}, Fraction of runs in which the direct certificate has asserted $\vphi>1.004$ bits by a given number of transitions ($\alpha=0.05$; 400 runs per source). Grey: exact power of containment under the same four-parameter model, which is zero for the outside sources at every sample size.
\p{g}, Transitions needed for a $95\%$ probability of certification at the center of the region (dark) and halfway to its boundary (light), for containment with 64 free probabilities, containment with four reliabilities, and the direct certificate.}
\label{fig:logic}
\end{figure}

\begin{figure}[h!]
\centering
\includegraphics[width=\linewidth]{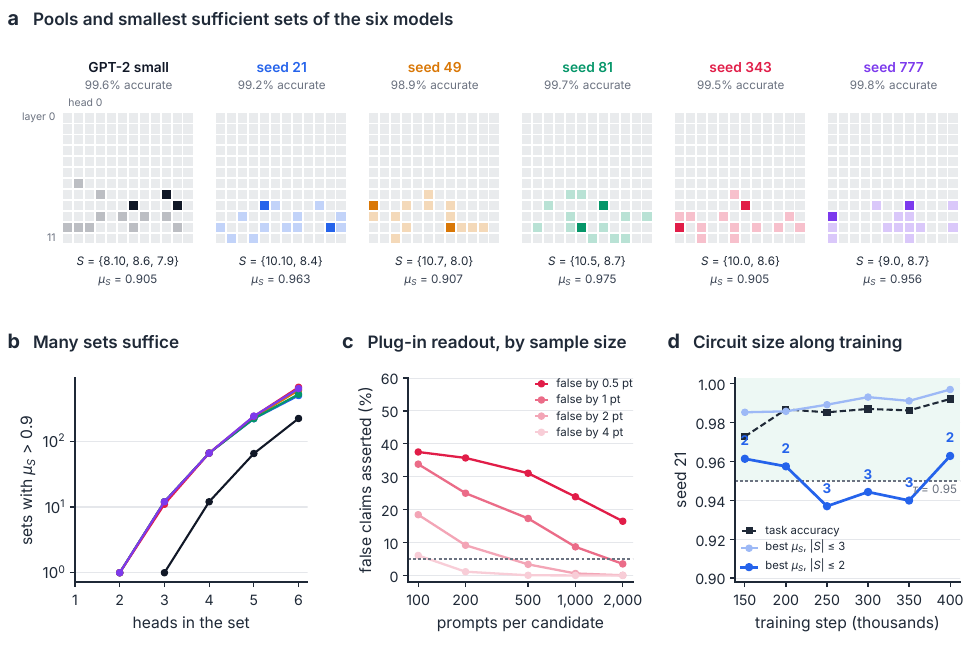}
\caption{\textbf{Circuits of the six language models.}
\p{a}, Candidate pool (light) and smallest set $S$ with $\mu_S>0.9$ (dark) for each model, on the grid of 12 layers (rows; first layer at the top) and 12 heads. Below each grid: the heads of $S$ (layer.head) and its interchange accuracy. In every model the smallest sufficient set is unique among the subsets of the pool, and in the five seed models it is the only sufficient pair among all 10{,}296 pairs of heads.
\p{b}, Number of sets of each size with $\mu_S>0.9$.
\p{c}, Fraction of false claims asserted by the plug-in readout against the number of prompts on which every candidate is evaluated, for thresholds placed $0.5$ to $4$ percentage points above the best set of each model's circuit size (mean over models); dashed line: $5\%$.
\p{d}, One model over the last 250 thousand of its 400 thousand training steps: task accuracy and the largest interchange accuracy over sets of at most two and three heads of its final pool, at six checkpoints. Numbers: circuit size at $\tau=0.95$. Task accuracy stays between $97.3\%$ and $99.2\%$ ($98.5\%$ or more from 200 thousand steps on) while the best pair of heads crosses the threshold twice.}
\label{fig:lmdetail}
\end{figure}

\begin{figure}[h!]
\centering
\includegraphics[width=\linewidth,trim=0 0 0 2.40in,clip]{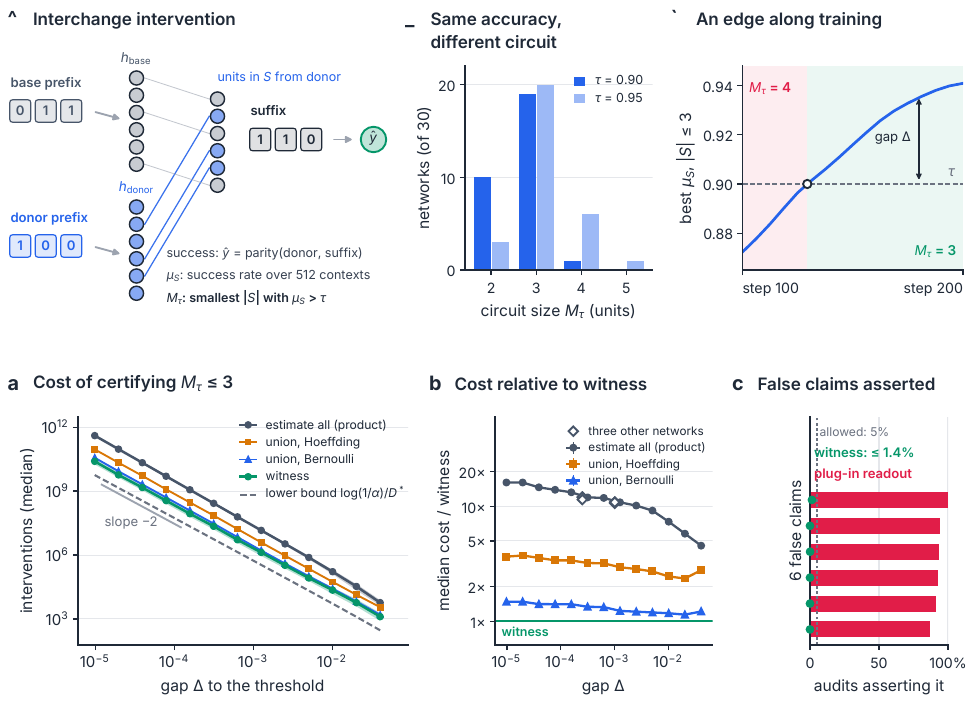}
\caption{\textbf{Certificates of circuit claims in recurrent networks.}
\p{a}, Median number of interventions needed to certify $\Mt\le3$ at $\alpha=0.05$ against the gap $\Delta$ between the best candidate and the threshold (1{,}000 audits per gap; bands: interquartile range, for the product and witness certificates), for four certificates evaluated on identical queries.
\p{b}, Cost relative to the witness certificate (ratio of medians with $95\%$ bootstrap intervals), for the network of \cref{fig:circuits}c and for three independently trained networks (diamonds; product certificate).
\p{c}, Six configurations in which the claim is false. A plug-in readout that queries every candidate 32 times asserts it with probability $87$--$100\%$ (bars); the witness certificate did so in at most $1.4\%$ of audits (dots; 2{,}000 to 50{,}000 audits per configuration), within its level of $5\%$ (dashed).}
\label{fig:gruaudit}
\end{figure}

\begin{figure}[h!]
\centering
\includegraphics[width=\linewidth]{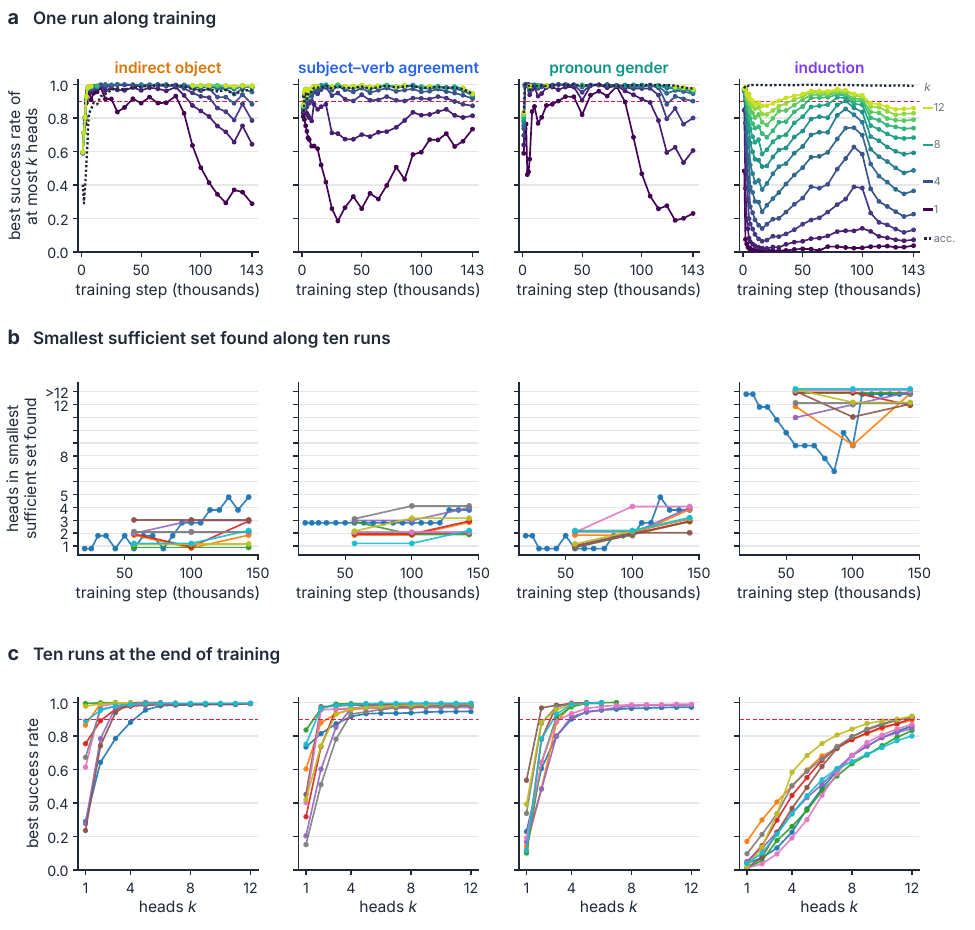}
\caption{\textbf{Circuits along ten training runs.}
\p{a}, The original run of Pythia-160m from step 1{,}000 to step 143{,}000: largest success rate found over sets of at most $k$ heads ($k=1$ to 12, dark to light) and task accuracy (dotted), for each of the four tasks. Dashed line: $\tau=0.9$.
\p{b}, Size of the smallest set of heads found with success rate above $0.9$ along the ten runs, from step 20{,}000 (one color per run; lines are displaced slightly so that equal values remain visible).
\p{c}, Largest success rate against the number of heads at the end of training, for the ten runs. Every rate is exact for the task's population of 1{,}000 prompt pairs.}
\label{fig:runsdetail}
\end{figure}

\begin{figure}[h!]
\centering
\includegraphics[width=\linewidth]{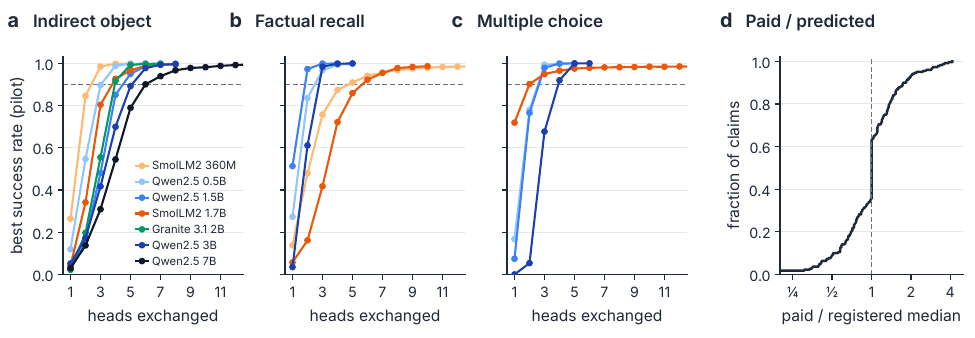}
\caption{\textbf{Prospective audits: sets found by the search, and the accuracy of the registered predictions.}
\p{a}--\p{c}, Largest success rate on the pilot prompt pairs (4{,}000; 2{,}000 for the largest model) among the sets of each size proposed by the search, for each instruction-tuned model, on indirect-object identification (\p{a}), factual recall (\p{b}) and multiple choice (\p{c}). Dashed line: $\tau=0.9$.
\p{d}, Distribution over all certified claims of the number of live queries relative to the median predicted at registration.}
\label{fig:livedetail}
\end{figure}

\begin{figure}[h!]
\centering
\includegraphics[width=\linewidth]{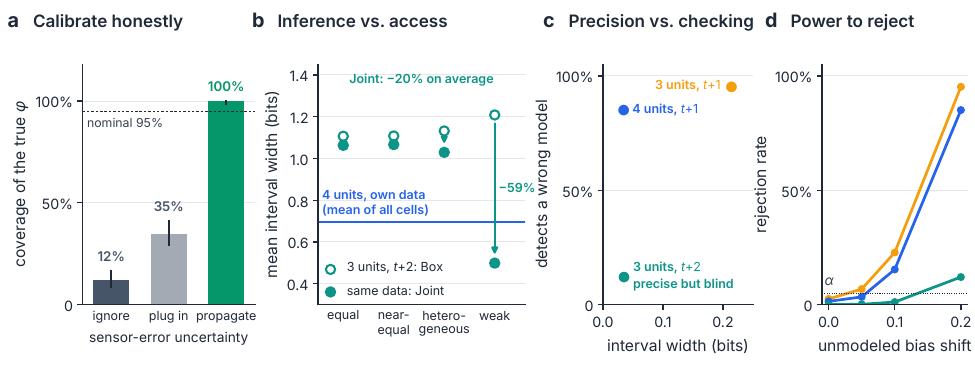}
\caption{\textbf{From attainable to delivered evidence.}
\p{a}, Coverage of the true $\vphi$ (heterogeneous source, sensor error $0.10$, 32{,}768 recording and 128 calibration trials) when sensor error is ignored, plugged in as an estimate, or propagated jointly with recording uncertainty (31, 89 and 256 of 256 datasets; whiskers: exact $95\%$ binomial intervals).
\p{b}, Mean interval width on identical two-step data with two inference procedures for four sources (216 cells, 55{,}296 datasets); blue line: four recorded units on their own data.
\p{c}, Mean interval width under a correct model against the rate of rejecting a wrong model (a recorded unit's bias shifted by $0.20$), for eight random logistic networks.
\p{d}, Rejection rate against the size of the unmodeled shift; dotted line: the bound $\alpha=0.05$ on rejecting a correct model. The design that is most precise under the assumed model is the worst at detecting that the model is wrong.}
\label{fig:pipeline}
\end{figure}
\FloatBarrier

% ============================================================================
% Supplementary Information
% ============================================================================
\clearpage
\FloatBarrier
\setcounter{section}{0}
\setcounter{figure}{0}
\setcounter{table}{0}
\setcounter{equation}{0}
\setcounter{theorem}{0}
\setcounter{definition}{0}
\renewcommand{\figurename}{Supplementary Fig.}
\renewcommand{\tablename}{Supplementary Table}
\renewcommand{\theequation}{S\arabic{equation}}
\renewcommand{\theHsection}{SI.\arabic{section}}
\renewcommand{\theHsubsection}{SI.\arabic{section}.\arabic{subsection}}
\renewcommand{\theHfigure}{SI.\arabic{figure}}
\renewcommand{\theHtable}{SI.\arabic{table}}
\renewcommand{\theHequation}{SI.\arabic{equation}}
\crefalias{section}{snote}
\crefalias{subsection}{snote}
\crefalias{figure}{sfigure}
\crefalias{table}{stable}
\crefname{theorem}{Theorem}{Theorems}
\crefname{proposition}{Proposition}{Propositions}
\crefname{lemma}{Lemma}{Lemmas}
\crefname{corollary}{Corollary}{Corollaries}
\crefname{definition}{Definition}{Definitions}
\titleformat{\section}{\sffamily\bfseries\large\color{ink}}{Supplementary Note \thesection}{0.7em}{}
\thispagestyle{plain}
\begin{center}
\phantomsection\label{pg:si}
{\sffamily\bfseries\Large Supplementary Information\par}
\vspace{8pt}
{\sffamily\bfseries\large The price of knowing how a network computes\par}
\vspace{8pt}
{\normalsize Diaoxuan (Nora) He, Linqian Wu, Xicheng Wang, Chang Kong and Haoming Peng\par}
\end{center}
\vspace{6pt}
\noindent\textsf{\textbf{Contents}}\quad
\textbf{Supplementary Note 1}: systems, quantities, experiments and certificates.
\textbf{Supplementary Note 2}: system integrated information, closed forms and validation.
\textbf{Supplementary Note 3}: theorems and proofs.
\textbf{Supplementary Note 4}: experimental protocols.
\textbf{Supplementary Note 5}: additional results (Supplementary Tables).
\vspace{4pt}

\section{Systems, quantities, experiments and certificates}\label{app:setting}

\paragraph{Systems as causal models.}
We study networks of $n$ binary units that update synchronously and, given the previous global state, independently---the standard stochastic recurrent network, the parallel-update kinetic Ising model, and the usual discrete-time abstraction of spiking circuits all fit this description. A \emph{mechanism} is a transition law
\begin{equation}\label{eq:kernel}
K_\theta(y\mid x)=\prod_{i=1}^{n}K_{\theta,i}(y_i\mid x),\qquad x,y\in\{0,1\}^n ,
\end{equation}
indexed by parameters $\theta$ (for example the weights and biases of logistic units, $K_{\theta,i}(1\mid x)=\sigma(\sum_j w_{ji}x_j+\beta_i)$). The exact part of the paper works with networks of four and five units, small enough that every quantity below can be computed without approximation and every inference checked against ground truth---the strategy that made toy models a workhorse of mechanistic interpretability \citep{elhage2022toy}. The same analysis is then applied to trained recurrent networks with real-valued hidden states and to language models (\cref{app:gru,app:lm}).

\paragraph{First target: integrated information $\vphi$.}
Informally, $\vphi$ asks how much of a system's power to constrain its own past and next states is destroyed by the least damaging way of cutting it into parts. It is zero when some part can be disconnected without loss---as in a feed-forward chain, whose first unit receives no inputs---and positive when every cut destroys something. IIT~4.0 makes this precise \citep{albantakis2023iit4}. At a current state $x$, the recipe implemented in the reference software PyPhi \citep{mayner2018pyphi} has three steps (\cref{app:phi}):
\begin{enumerate}[leftmargin=1.4em,itemsep=1pt,topsep=2pt]
\item \emph{Specify} the past state $z^\ast$ and next state $y^\ast$ that the current state $x$ constrains most strongly---its cause and its effect, scored by intrinsic information.
\item For every directed \emph{cut} $C$ that severs connections between parts of the system (150 cuts for four units), compute the loss $L_C$ of cause--effect power at $(z^\ast,y^\ast)$ and the number $e_C$ of severed connections.
\item \emph{Select} the cut with the smallest loss \emph{per severed connection} and \emph{report} its raw loss, capped by the intrinsic information $I$ of the specified cause and effect:
\begin{equation}\label{eq:phi}
C^\ast=\argmin_C \frac{L_C}{e_C},\qquad \vphi=\min\{I,\;L_{C^\ast}\}.
\end{equation}
\end{enumerate}
Ties in \cref{eq:phi} are broken toward the larger raw loss (the PyPhi default). We also track $\vphiraw=\min\{I,\min_C L_C\}$, which selects and reports the \emph{same} quantity. It is not part of the theory; it isolates the effect of the selection rule. For a four-unit system $\vphi$ can never exceed $B_4=1.21$ bits. Following IIT's own examples, we treat the network's units as the causal substrate.

One family has a closed form and recurs throughout. In a \emph{parity network}, each unit fires when an odd number of its inputs are active and computes this parity correctly with \emph{reliability} $p_i\in[1/2,1]$. Two reparametrizations are convenient: the \emph{fidelity} $a_i=2p_i-1$, which is $0$ for a unit that ignores its inputs and $1$ for a perfect one, and the \emph{log-reliability} $\lambda_i=\log_2(2p_i)\in[0,1]$, the number of bits lost when unit $i$ is replaced by a fair coin. When the noiseless map is invertible, $\vphi$ is piecewise explicit (\cref{thm:null}); for the four-unit network in which each unit reads the other three,
\begin{equation}\label{eq:closed}
\vphi(p)=\min\bigl\{\,r\,q\log_2(2m),\;q\log_2(16q),\;-\log_2 q\,\bigr\},\qquad q=\textstyle\prod_i p_i,\; m=\min_i p_i,
\end{equation}
where $r$ is the number of units attaining the minimum; $\vphiraw$ is the same expression with $r=1$ (\cref{app:closed}). Our exact implementation of the general recipe reproduces every value of \cref{eq:closed} to $10^{-15}$ and, to nine decimals, seven reference values from an independent exact-rational implementation (\cref{app:validation}).

\paragraph{Second target: the size of a circuit.}
In a trained network the analogous question is which internal variables causally carry a computation. For a set $S$ of hidden units or attention heads, the \emph{interchange-intervention accuracy} $\mu_S$ is the probability that the network produces the output dictated by a donor input after the activations of $S$ are replaced by those the donor induces \citep{geiger2021causal}. The \emph{minimal sufficient intervention size} $\Mt=\min\{|S|:\mu_S>\tau\}$ is the size of the smallest set that carries the task variable at reliability $\tau$. Like $\vphi$, it is defined by selecting among candidates and it is not determined by behavior.

\paragraph{Experiments.}
An experimenter cannot read $K_\theta$. A \emph{design} specifies a finite set of \emph{actions}; each action prepares a state by intervention, lets the system update, and records chosen units at chosen delays, possibly through a noisy sensor (we use a sensor that flips all recorded bits of a trial together with probability $\eta<1/2$, estimated from calibration trials). Action $a$ returns an outcome with law $P^a_\theta$; trials can be repeated, and the next action may depend on everything seen so far. We call the mechanism that actually generates the data the \emph{source}.

\paragraph{Intervals.} Any confidence set $\mathcal C$ that contains $\theta$ with probability $1-\alpha$ yields the interval
\begin{equation}\label{eq:transport}
\Bigl[\,\inf_{\theta'\in \mathcal C}F(\theta'),\ \sup_{\theta'\in \mathcal C}F(\theta')\,\Bigr],
\end{equation}
which covers the true $F(\theta)$ with at least the same probability. \Cref{app:samefunction,app:floor,app:product} ask when such intervals can be short---and when, for a given design, no sample size or estimator can make them so.

\paragraph{Claims, certificates and the price of evidence.}
Most questions about mechanisms are one-sided: is the system integrated above some level, does a circuit of at most $k$ units exist. We therefore study \emph{claims} of the form $F(\theta)>c$ and write $\Hnull=\{\theta:F(\theta)\le c\}$ for the mechanisms that falsify the claim.

\begin{definition}[Certificate]\label{def:cert}
A \emph{certificate of level $\alpha$} for the claim $F>c$ is a rule that, after each trial, may stop and assert the claim, and that asserts it with probability at most $\alpha$ under every mechanism in $\Hnull$, whatever the (adaptive) choice of actions and however long the experiment runs.
\end{definition}

The guarantee is uniform over mechanisms and over time: it does not depend on a prior, on a fitted model being right, or on the sample size having been fixed in advance. What such a certificate costs is governed by a single quantity. For a design that uses action $a$ in a fraction $w_a$ of its trials, let
\begin{equation}\label{eq:DstarS}
D_w(\theta)=\inf_{\theta'\in\cl{\Hnull}}\ \sum_a w_a\,\KL\bigl(P^a_\theta\,\|\,P^a_{\theta'}\bigr),\qquad D^\ast(\theta)=\max_w D_w(\theta),
\end{equation}
the \emph{information distance} from $\theta$ to the closure of the falsifying mechanisms, per trial, under the best allocation of trials.

\section{The target: system integrated information}\label{app:phi}

\paragraph{States, cuts and repertoires.}
States are integers $0,\dots,2^n-1$ whose bit $i-1$ is the state of unit $i$. The current state $x$ is declared before any analysis, and past states receive a uniform interventional prior. Conditional on a past state, units update independently, so $K(y\mid x)=\prod_iK_i(y_i\mid x)$. For every partition of the units into $k\ge2$ parts, each part is assigned one of three directions---sever all connections into it, out of it, or both---and the union of these directed cuts defines a cut matrix $C$ with $C_{ji}=1$ when past unit $j$ is replaced by noise as an input to unit $i$. Removing duplicates leaves 22 cuts for three units, 150 for four and 1{,}061 for five. The edge count $e_C$ is the number of entries equal to one, including severed connections that happen to be irrelevant to a unit's function. Under a cut, each unit's firing probability is averaged over uniform assignments to its severed inputs (causal marginalization), and the unit factors are multiplied afterwards to give $K_C$.

\paragraph{Specification, loss and cap.}
Let $E(y)=K(y\mid x)$, $U(y)=2^{-n}\sum_zK(y\mid z)$, $H(z)=K(x\mid z)$ and $B(z)=H(z)/\sum_wH(w)$. The specified effect $y^\ast$ maximizes $E(y)\log_2\{E(y)/U(y)\}$ and the specified cause $z^\ast$ maximizes $B(z)\log_2\{2^nB(z)\}$. The raw effect loss of a cut is $E(y^\ast)\log_2\{K(y^\ast\mid x)/K_C(y^\ast\mid x)\}$ and the raw cause loss is $B(z^\ast)\log_2\{K(x\mid z^\ast)/K_C(x\mid z^\ast)\}$, keeping the uncut cause background. Their nonnegative minimum is $L_C$. The intrinsic cap $I$ is the minimum of the effect specification, the cause specification, $-\log_2E(y^\ast)$ and $-\log_2B(z^\ast)$ \citep{mayner2026intrinsic}.

\paragraph{Selection and reporting.}
\Cref{tab:definitions} lists the rules compared in this paper. The target $\vphi$ follows IIT~4.0 \citep{albantakis2023iit4} and the documented PyPhi default \citep{mayner2018pyphi}: the cut minimizing $L_C/e_C$ is selected; exact ties go to the larger raw loss, then to a fixed cut order; the reported value is $\min\{I,L_{\rm selected}\}$. We analyze the declared system. In the version of ref.~\citenum{mayner2026intrinsic} that we consulted, the displayed partition-selection equation is written with a maximum while the adjacent text describes a minimum; we follow the text and the PyPhi default. A normalized-maximum variant changes the values but not the select-then-report structure that creates edges.

\begin{table}[h]\centering\small
\caption{Selection and reporting rules. All rows share the transition law, the allowed cuts and the cap.}\label{tab:definitions}
\begin{tabular}{@{}p{0.2\linewidth}p{0.36\linewidth}p{0.36\linewidth}@{}}\toprule
Quantity & Partition selection & Role\\\midrule
$\vphi$ (target) & minimum loss per severed connection; ties to the larger raw loss & IIT 4.0 system integrated information\\
$\vphiraw$ & minimum raw loss & comparison, not part of the theory; continuous in the parity family, jumps only where the specified states switch\\
$\vphi_\tau$ & as $\vphi$, with comparison tolerance $10^{-13}$ & values saved by software\\
$\Phi$ & --- & sum over the full cause--effect structure (all distinctions and relations)\\\bottomrule
\end{tabular}
\end{table}

\subsection{Closed form for the first parity network}\label{app:closed}
In the first parity network each unit computes the parity of the other three (input masks $14,13,11,7$) and is correct with probability $p_i\in[1/2,1]$, independent of the past state. The map is linear and invertible, so the output prior is uniform and the correct output and its unique inverse cause both have probability $q=\prod_ip_i$. Under a cut, unit $i$ becomes a fair coin exactly when one of its parity inputs is severed, and the remaining parity rows stay independent. If $A(C)$ is the set of affected units, the cause and effect losses both equal $L_C=q\sum_{i\in A(C)}\log_2(2p_i)$. Each affected unit has at most three severed inputs, so $L_C/e_C\ge q\log_2(2m)/3$ with $m=\min_ip_i$, with equality exactly for cuts that sever all three inputs of affected units that attain $m$. The larger-raw tie rule selects the cut affecting all $r$ weakest units, so the selected raw loss is $rq\log_2(2m)$. The cap is $\min\{q\log_2(16q),-\log_2q\}$, which gives \cref{eq:closed}. At $p_i=421/512$ all four units are weakest and the cap binds ($1.1292943$ bits); a perturbation that makes one unit uniquely weakest gives the limit $q\log_2(2p)=0.3280781$ bits. Selecting by raw loss always chooses a single weakest unit, which gives $\vphiraw$. This is the case $e_S=3|S|$ of \cref{thm:null}.

\subsection{Validation of the implementation}\label{app:validation}
We implemented the general recipe for arbitrary small networks, vectorized over batches of mechanisms. It reproduces \cref{eq:closed} to within $3\times10^{-16}$ on 20{,}000 random parity laws, all seven reference values from an independent exact-rational implementation (\cref{tab:validation}), $\vphi=0$ for the feed-forward chain at every state, the bound of \cref{thm:region} on 8{,}000 random and corner laws of the proved region at all 16 states (smallest value found $1.0197$ bits), the reduced formula \cref{eq:phired} to $6\times10^{-16}$ on 4{,}000 random laws of the second parity network at all 16 states, and both phases of \cref{thm:twophase} on 40{,}000 laws in $U$ and 60{,}000 laws with 64 free probabilities.

\begin{table}[h]\centering\small
\caption{Reference values of $\vphi$ in the first parity network (all states). $\delta$ is the decrease of one reliability from $421/512$.}\label{tab:validation}
\begin{tabular}{@{}lccc@{}}\toprule
Case & $\vphi$ (this work) & $\vphi$ (exact-rational implementation) & $\vphiraw$\\\midrule
$\delta=0$ & 1.129294308 & 1.129294308 & 0.328078083\\
$\delta=2^{-24}$ & 0.328078011 & 0.328078011 & 0.328078011\\
$\delta=2^{-16}$ & 0.328059756 & 0.328059756 & 0.328059756\\
$\delta=15/4096$ & 0.323686219 & 0.323686219 & 0.323686219\\
$\delta=1/32$ & 0.291027385 & 0.291027385 & 0.291027385\\
$(0.78,0.80,0.83,0.85)$ & 0.282429091 & 0.282429091 & 0.282429091\\
$(0.56,0.62,0.68,0.74)$ & 0.028565034 & 0.028565034 & 0.028565034\\\bottomrule
\end{tabular}
\end{table}

\section{Theorems and proofs}\label{app:proofs}

\subsection{What can be certified, and at what price}\label{app:price0}
\begin{theorem}[What can be certified, and at what price]\label{thm:price0}
Let the design have finitely many actions, each with finitely many outcomes, and let every $P^a_\theta$ have full support and depend continuously on $\theta$. For the claim $F>c$ and any level $\alpha\in(0,1)$:
\begin{enumerate}[leftmargin=1.6em,itemsep=1pt,topsep=2pt,label=(\roman*)]
\item \emph{Blind spots.} If $D^\ast(\theta)=0$---in particular if $\theta$ lies in the closure of $\Hnull$, or produces the same data as a mechanism that does---then every certificate of level $\alpha$ asserts the claim with probability at most $\alpha$ under $\theta$, at every sample size.
\item \emph{Price.} If $D^\ast(\theta)>0$, every certificate of level $\alpha$ that asserts the claim with probability one under $\theta$ needs at least $\log(1/\alpha)/D^\ast(\theta)$ trials in expectation.
\item \emph{Attainability.} Let $L_t(\theta')$ be the likelihood of the first $t$ trials and $R_t$ the probability a fixed mixture over mechanisms assigns to them. The \emph{direct certificate}, which asserts the claim the first time
\begin{equation}\label{eq:directS}
E_t=\frac{R_t}{\sup_{\theta'\in\Hnull}L_t(\theta')}\ \ge\ \frac1\alpha ,
\end{equation}
has level $\alpha$ under every design, and any upper bound on the supremum may be used in its place. Let $R_t$ be the product, over actions, of uniform mixtures over outcome laws (or, in the product families of this paper, over cell probabilities), and let actions be drawn independently of the past with probabilities $w$. If $D_w(\theta)>0$, the direct certificate asserts the claim with probability one under $\theta$ and its expected number of trials is $\{1+o(1)\}\log(1/\alpha)/D_w(\theta)$ as $\alpha\to0$. With $w$ attaining $D^\ast(\theta)$ this matches the bound of (ii), so a claim is certifiable at $\theta$ exactly when $D^\ast(\theta)>0$.
\end{enumerate}
\end{theorem}

Throughout, an adaptive design chooses the action $A_t$ from the first $t-1$ actions and outcomes (and independent randomization), and $\tau\le\infty$ is the time at which a certificate asserts the claim.

\paragraph{(i) Blind spots.} Let $D^\ast(\theta)=0$ and take $w$ uniform over the finitely many actions. Since $D_w(\theta)\le D^\ast(\theta)=0$, there are $\theta'_k\in\cl{\Hnull}$ with $\sum_a\KL(P^a_\theta\|P^a_{\theta'_k})\to0$, hence by Pinsker's inequality with $\max_a\TV(P^a_\theta,P^a_{\theta'_k})\to0$; by continuity of $\theta\mapsto P^a_\theta$ each $\theta'_k$ can be replaced by some $\theta_k\in\Hnull$ with $\epsilon_k=\max_a\TV(P^a_\theta,P^a_{\theta_k})\to0$. Under a common design, the laws of the first $n$ trials under $\theta$ and $\theta_k$ differ by at most $n\epsilon_k$ in total variation, so $\Pr_\theta(\tau\le n)\le\Pr_{\theta_k}(\tau\le n)+n\epsilon_k\le\alpha+n\epsilon_k$. Letting $k\to\infty$ and then $n\to\infty$ gives $\Pr_\theta(\tau<\infty)\le\alpha$. A mechanism in $\cl{\Hnull}$, or with the same data law as one, has $D_w=0$ for every $w$.

\paragraph{(ii) Price.} Let $\Pr_\theta(\tau<\infty)=1$ and $\E_\theta\tau<\infty$, and fix $\theta'\in\Hnull$. Let $N_a$ be the number of trials with action $a$ up to $\tau$. Outcomes are finite with full support, so the log-likelihood ratio of the stopped record has bounded increments, and Wald's identity gives $\KL(\mathcal L_\theta\|\mathcal L_{\theta'})=\sum_a\E_\theta[N_a]\KL(P^a_\theta\|P^a_{\theta'})$ for the laws $\mathcal L$ of the record up to $\tau$. By the data-processing inequality applied to the event $\{\tau<\infty\}$, this is at least $\kl(1,\Pr_{\theta'}(\tau<\infty))\ge\log(1/\alpha)$ (Lemma~1 of ref.~\citenum{kaufmann2016complexity}). The inequality extends to $\theta'\in\cl{\Hnull}$ by continuity. With $w_a=\E_\theta[N_a]/\E_\theta\tau$ it reads $\E_\theta\tau\cdot\sum_aw_a\KL(P^a_\theta\|P^a_{\theta'})\ge\log(1/\alpha)$ for every $\theta'\in\cl{\Hnull}$; taking the infimum over $\theta'$ gives $\E_\theta\tau\ge\log(1/\alpha)/D_w(\theta)\ge\log(1/\alpha)/D^\ast(\theta)$.

\paragraph{(iii) Attainability.} \emph{Level.} Fix $\theta'\in\Hnull$. Because $R_t$ is the probability a fixed mixture assigns to the outcomes given the actions, $M_t=R_t/L_t(\theta')$ is a nonnegative martingale under $\theta'$ with $M_0=1$, for any adaptive design. The statistic satisfies $E_t\le M_t$, and remains below $M_t$ if the supremum in \cref{eq:directS} is replaced by any upper bound. Ville's inequality gives $\Pr_{\theta'}(\sup_tE_t\ge1/\alpha)\le\Pr_{\theta'}(\sup_tM_t\ge1/\alpha)\le\alpha$. The statistic itself need not be a martingale \citep{wasserman2020universal}.

\emph{Rate.} Call a \emph{cell} an unknown probability vector that the data address: the outcome law of an action in general, or, in the families of this paper, a Bernoulli probability (four reliabilities; 64 state-dependent reliabilities; one success probability per candidate), several of which a trial may observe independently. We write the argument for Bernoulli cells; for a cell with $k$ outcomes the uniform Dirichlet mixture replaces the uniform mixture below, $\KL$ replaces $\kl$, and the factor $m+1$ becomes $\binom{m+k-1}{k-1}\le(m+1)^{k-1}$. Let $d$ be the number of cells (the total number of free probabilities in general). Take $R_t$ to be the product over cells of the uniform mixture, $\int_0^1u^{s}(1-u)^{m-s}du=\{(m+1)\binom ms\}^{-1}$ for a cell with $s$ successes in $m$ observations. Since $\binom ms(s/m)^s(1-s/m)^{m-s}\le1$, the mixture is within a factor $m+1$ of the maximized likelihood of the cell, so $\log R_t\ge\log\sup_{\theta''}L_t(\theta'')-d\log(t+1)$, the supremum being over all cell probabilities. Moreover $\log\sup_{\theta''}L_t(\theta'')-\log L_t(\theta')=\sum_jm_j\,\kl(\hat\theta_j,\theta'_j)$ with $\hat\theta$ the empirical cell frequencies and $m_j$ the cell counts. Hence
\begin{equation}\label{eq:ratebound}
\log E_t\ \ge\ t\,\hat D_t-d\log(t+1),\qquad \hat D_t=\inf_{\theta'\in\cl{\Hnull}}\sum_j\frac{m_j}{t}\,\kl(\hat\theta_j,\theta'_j),
\end{equation}
where $\hat D_t$ is the information distance of the empirical mechanism under the empirical proportions (and $\log E_t\le t\hat D_t$, because the mixture is below the maximized likelihood).

Let $v_j$ be the probability that a trial addresses cell $j$ when actions are drawn with probabilities $w$, so that $D_w(\theta)=\inf_{\theta'\in\cl{\Hnull}}\sum_jv_j\kl(\theta_j,\theta'_j)$, and write $\psi(\vartheta,u)=\inf_{\theta'\in\cl{\Hnull}}\sum_ju_j\kl(\vartheta_j,\theta'_j)$, so that $\hat D_t=\psi(\hat\theta,m/t)$. The function $\psi$ is lower semicontinuous at $(\theta,v)$. Indeed, let $(\vartheta^k,u^k)\to(\theta,v)$ with near-minimizers $\theta'^k$; the cell probabilities of the falsifying mechanisms lie in the compact cube $[0,1]^d$, so along a subsequence they converge to some $\theta'$ in the closure of that set. Because $\kl$ is jointly lower semicontinuous, with $\kl(\theta_j,\theta'_j)=\infty$ when $\theta_j\in(0,1)$ and $\theta'_j\in\{0,1\}$, $\liminf_k\sum_ju^k_j\kl(\vartheta^k_j,\theta'^k_j)\ge\sum_{j:v_j>0}v_j\kl(\theta_j,\theta'_j)$. If the right-hand side is finite, $\theta'_j\in(0,1)$ for every cell with $v_j>0$, and by continuity of $\kl(\theta_j,\cdot)$ on $(0,1)$ the right-hand side is the limit of $\sum_jv_j\kl(\theta_j,\cdot)$ along falsifying mechanisms that approach $\theta'$, hence at least $D_w(\theta)$.

Fix $\epsilon>0$. By lower semicontinuity there is $\eta>0$ such that $\hat D_t\ge D_w(\theta)-\epsilon$ whenever $|m_j/t-v_j|\le\eta$ for all cells and $|\hat\theta_j-\theta_j|\le\eta$ for the cells with $v_j>0$. Actions are drawn independently of the past, so by Hoeffding's inequality and a union bound the probability that this fails at time $t$ is at most $Ce^{-ct}$ for constants $C,c>0$ that do not depend on $\alpha$; the last time $T_\epsilon$ at which it fails therefore has finite expectation. After $T_\epsilon$, \cref{eq:ratebound} gives $\log E_t\ge t\{D_w(\theta)-\epsilon\}-d\log(t+1)$, so $\tau\le\max\{T_\epsilon,t_\alpha\}+1$ with $t_\alpha$ the largest solution of $t\{D_w(\theta)-\epsilon\}-d\log(t+1)=\log(1/\alpha)$, and $t_\alpha/\log(1/\alpha)\to1/\{D_w(\theta)-\epsilon\}$. Hence $\tau<\infty$ almost surely and $\limsup_{\alpha\to0}\E_\theta\tau/\log(1/\alpha)\le1/D_w(\theta)$. For the matching lower bound, $\E_\theta[N_a]=w_a\E_\theta\tau$ by Wald's identity for the independent action draws, so (ii) reads $\E_\theta\tau\ge\log(1/\alpha)/D_w(\theta)$. Finally, $w\mapsto D_w(\theta)$ is an infimum of linear functions, hence concave and upper semicontinuous on the simplex, and attains its maximum $D^\ast(\theta)$. At a fixed level the term $d\log(t+1)$ is the price of not knowing the mechanism in $d$ directions; \cref{app:witness} shows how much of it can be removed when the claim has more structure.

\paragraph{Envelope.} If $F_\ast(\theta)>c$ there is a neighborhood of $\theta$ on which $F>c$, so $\theta\notin\cl{\Hnull}$; conversely, if $\theta\notin\cl{\Hnull}$ then $F>c$ on a neighborhood and $F_\ast(\theta)\ge c$.

\subsection{The price is attained without knowledge of the mechanism}\label{app:adaptive}
The proportions $w$ that attain $D^\ast(\theta)$ depend on the mechanism under study. The next theorem shows that they can be learned during the experiment at no cost to the leading order: one design, which uses no knowledge of $\theta$, attains the price of every certifiable mechanism. It extends the tracking principle of optimal best-arm identification \citep{garivier2016optimal,degenne2019pure} to claims about mechanism with arbitrary falsifying sets. We use the notation of the proof of \cref{thm:price0}(iii): cells $j=1,\dots,d$, the probability $v_j(w)=\sum_{a\ni j}w_a$ that a trial addresses cell $j$ when actions are drawn with probabilities $w$ (the sum is over the actions that address cell $j$; each action addresses a fixed set of cells), and $\psi(\vartheta,u)=\inf_{\theta'\in\cl{\Hnull}}\sum_ju_j\kl(\vartheta_j,\theta'_j)$ with the convention $0\cdot\infty=0$, so that $D_w(\theta)=\psi(\theta,v(w))$.

\begin{theorem}[Adaptive designs attain the price]\label{thm:adaptive}
Under the hypotheses of \cref{thm:price0}, let every cell be addressed by at least one of the $A$ actions and let $\Hnull$ be nonempty. After $t$ trials let $\tilde\theta_t$ be the empirical cell probabilities smoothed by one pseudo-observation of each outcome, and let $w^t$ be the maximizer of smallest Euclidean norm of $w\mapsto\psi(\tilde\theta_t,v(w))$ over the simplex. The \emph{tracking design} draws the action of trial $t+1$ from $(1-\epsilon_t)w^t+\epsilon_t\,\mathrm{unif}$, with $\epsilon_t=(t+1)^{-1/4}$. With this design the direct certificate of \cref{thm:price0}(iii) has level $\alpha$, and for every mechanism $\theta$ with $D^\ast(\theta)>0$ it asserts the claim with probability one and
\begin{equation}\label{eq:adaptive}
\lim_{\alpha\to0}\frac{\E_\theta\tau}{\log(1/\alpha)}=\frac1{D^\ast(\theta)} .
\end{equation}
\end{theorem}

\paragraph{Proof.} \emph{Level and lower bound.} The level argument of \cref{thm:price0}(iii) and the bound of \cref{thm:price0}(ii) hold for every adaptive design, so the certificate has level $\alpha$ and $\E_\theta\tau\ge\log(1/\alpha)/D^\ast(\theta)$. It remains to show that $\limsup_{\alpha\to0}\E_\theta\tau/\log(1/\alpha)\le1/D^\ast(\theta)$.

\emph{Continuity.} Write $I$ for the set of cell probabilities with every outcome probability positive ($I=(0,1)^d$ for $d$ Bernoulli cells) and $U=[0,1]^d$ for the proportions of the $d$ cells. Every falsifying mechanism has its cell probabilities in $I$, and for $\vartheta\in I$ the objective $\sum_ju_j\kl(\vartheta_j,\theta'_j)$ at a point $\theta'$ of $\cl{\Hnull}$ is the limit of the objective along falsifying mechanisms that approach $\theta'$ (a term with $u_j>0$ and $\theta'_j\in\{0,1\}$ is infinite and is approached from below; a term with $u_j=0$ vanishes throughout). Hence on $I\times U$ the function $\psi$ is the infimum over $\theta'\in\Hnull$ of functions that are continuous in $(\vartheta,u)$, so it is upper semicontinuous there; it is finite because $\Hnull$ is nonempty. The compactness argument given for lower semicontinuity in the proof of \cref{thm:price0}(iii) applies at every point of $I\times U$. So $\psi$ is continuous on $I\times U$. By the maximum theorem, $\vartheta\mapsto D^\ast(\vartheta)=\max_w\psi(\vartheta,v(w))$ is continuous on $I$ and the set $W(\vartheta)$ of maximizers is nonempty, compact and upper hemicontinuous in $\vartheta$; it is convex because $w\mapsto\psi(\vartheta,v(w))$ is an infimum of linear functions. The maximizer of smallest norm in a compact convex set is unique, so $w^t$ is well defined, and it is a function of the finitely many possible records of $t$ trials.

Fix $\theta$ with $D^\ast(\theta)>0$ and $\epsilon>0$. By continuity of $\psi$ and compactness of $W(\theta)$ there is $\rho>0$, smaller than every outcome probability of $\theta$, such that $\psi(\vartheta,u)\ge D^\ast(\theta)-\epsilon$ whenever $\max_j|\vartheta_j-\theta_j|\le\rho$ and $\max_j|u_j-v_j(w)|\le\rho$ for some $w\in W(\theta)$. By upper hemicontinuity there is $\delta\in(0,\rho]$ such that every $w\in W(\vartheta)$ lies within $\rho/4$ of $W(\theta)$ in $\ell_1$ distance whenever $\max_j|\vartheta_j-\theta_j|\le\delta$.

\emph{Concentration.} Let $p_{s+1}=(1-\epsilon_s)w^s+\epsilon_s\,\mathrm{unif}$ be the conditional law of the action of trial $s+1$, let $N_a(t)$ be the number of the first $t$ trials with action $a$, and let $m_j(t)$ be the number of observations of cell $j$. Consider the event $G_t$ on which (a)~$\max_j|\tilde\theta_{s,j}-\theta_j|\le\delta$ for every $s\in[\sqrt t,t]$, and the unsmoothed frequencies satisfy $\max_j|\hat\theta_{t,j}-\theta_j|\le\rho$; and (b)~$\sum_a|N_a(t)-\sum_{s<t}p_{s+1}(a)|\le t\rho/4$. On $G_t$, every $w^s$ with $s\ge\sqrt t$ is within $\rho/4$ of a point $\pi_s\in W(\theta)$, and the average $\bar w$ of these $\pi_s$ lies in $W(\theta)$ by convexity. The first $\sqrt t$ trials and the exploration terms change the average of the $p_{s+1}$ by at most $4t^{-1/2}+2t^{-1}\sum_{s<t}\epsilon_s\le4t^{-1/2}+3t^{-1/4}$ in $\ell_1$, so for $t\ge t_0(\rho)$ we have $\sum_a|N_a(t)/t-\bar w_a|\le3\rho/4$ and therefore $\max_j|m_j(t)/t-v_j(\bar w)|\le\rho$, since $v_j$ is a sum of coordinates of $w$. Hence $\psi(\hat\theta_t,m(t)/t)\ge D^\ast(\theta)-\epsilon$ on $G_t$ for $t\ge t_0$.

The complement of $G_t$ is small. For (b), $N_a(t)-\sum_{s<t}p_{s+1}(a)$ is a martingale with increments in $[-1,1]$, so by the Azuma--Hoeffding inequality (b) fails with probability at most $2A\exp\{-t\rho^2/(32A^2)\}$. For (a), cell $j$ is addressed at trial $s+1$ with conditional probability at least $\epsilon_s/A$, and $\sum_{r<s}\epsilon_r\ge s^{3/4}/2$; the same inequality gives $m_j(s)\ge s^{3/4}/(4A)$ except with probability $\exp\{-s^{1/2}/(32A^2)\}$. The successive observations of a cell are independent draws from its law whatever the design, so the probability that the empirical frequency of cell $j$ deviates from $\theta_j$ by more than $\delta/2$ at some count $n\ge n_0$ is at most $\sum_{n\ge n_0}2e^{-n\delta^2/2}$, and smoothing moves a frequency by at most $1/n$ ($(k-1)/n$ for a cell with $k$ outcomes), which is below $\delta/2$ once $t$ is large. Summing over cells and over $s\ge\sqrt t$ gives $\Pr_\theta(G_t^c)\le Ce^{-ct^{1/4}}$ for large $t$, and hence for all $t$ after enlarging $C$, with constants $C,c>0$ that depend on $\theta$ and $\epsilon$ but not on $\alpha$.

\emph{Rate.} Let $T_\epsilon$ be the last time $t$ at which $G_t$ fails or $t<t_0$. Then $\E_\theta T_\epsilon\le t_0+\sum_t\sum_{s\ge t}\Pr_\theta(G_s^c)<\infty$. The bound \cref{eq:ratebound} holds for every sequence of actions, so after $T_\epsilon$ we have $\log E_t\ge t\{D^\ast(\theta)-\epsilon\}-d\log(t+1)$, and the argument that closes the proof of \cref{thm:price0}(iii) gives $\tau<\infty$ almost surely and $\limsup_{\alpha\to0}\E_\theta\tau/\log(1/\alpha)\le1/\{D^\ast(\theta)-\epsilon\}$. Let $\epsilon\to0$.

\paragraph{Numerical illustration.} For four candidates with success probabilities $0.75$, $0.6$, $0.4$ and $0.1$ and the universal claim $\max_S\mu_S<0.9$, the optimal proportions are $(0.68,0.20,0.08,0.04)$ and $D^\ast=0.0627$ nats per trial, $2.72$ times the information distance of the uniform design. Over 2{,}000 simulated audits per design, the mean number of trials of the tracking design, in units of $\log(1/\alpha)/D^\ast$, was $4.65$, $1.62$, $1.17$ and $1.07$ at $\log(1/\alpha)=3$, $30$, $300$ and $3{,}000$; the design that is told the optimal proportions gave $5.45$, $1.69$, $1.15$ and $1.04$, and the uniform design $11.03$, $3.79$, $2.87$ and $2.74$ (\texttt{exp\_adaptive\_design.py}).

\subsection{Same function, different integration}\label{app:samefunction}
\begin{proposition}[Same function, different $\vphi$]\label{prop:samefunction}
Let $\mathcal R$ be the recurrent parity network in which each unit reads the other three, with reliabilities $(0.78,0.80,0.83,0.85)$, so that $A=\prod_i(2p_i-1)=0.155232$, and let $\mathcal F$ be a feed-forward chain in which unit~1 is a fair coin and each later unit copies its predecessor with reliability $(1+\sqrt A)/2\approx0.697$. Taking the input to be unit~1 at time $t$ and the output to be unit~1 of $\mathcal R$ (unit~3 of $\mathcal F$) at time $t+2$, the two systems have identical input--output laws from every one of the 16 initial states, yet, at every current state, $\vphi(\mathcal R)=0.28$ bits and $\vphi(\mathcal F)=0$.
\end{proposition}

Write each unit as a spin $S_i=(-1)^{x_i}$ and number the units $1,\dots,4$. In the parity network $S_i(t+1)=Z_i(t+1)\prod_{j\ne i}S_j(t)$ with independent noise spins of mean $a_i=2p_i-1$. Expanding twice,
\[
S_1(2)=Z_1(2)\prod_{j\ne1}Z_j(1)\cdot S_1(0)^3\prod_{k\ne1}S_k(0)^2=Z_1(2)\prod_{j\ne1}Z_j(1)\,S_1(0),
\]
because unit~1's initial spin enters all three of its inputs' updates and every other initial spin enters twice. The output is the input times an independent spin of mean $A=\prod_ia_i=0.155232$, whatever the other initial states. In the chain, $S_3(2)=Z_3(2)Z_2(1)S_1(0)$ with link fidelities $\sqrt A=0.394$ (copying reliability $(1+\sqrt A)/2\approx0.697$), so the output is again the input times an independent spin of mean $A$. Hence the two input--output laws coincide from every initial state; the output equals the input with probability $(1+A)/2=0.577616$. In the chain, the cut that severs every connection into unit~1 leaves $K$ unchanged, so its loss is zero and $\vphi=0$ for any capping rule. The equivalence is specific to this delay: comparing unit~1 of the parity network with unit $d+1$ of the chain after $d$ updates gives largest total-variation distances $0.477$ ($d=1$) and $0.074$ ($d=3$).

\begin{proposition}[Unfolding]\label{prop:unfolding}
For every network $K$ of the form \cref{eq:kernel} on $n$ units and every horizon $T$, there is an acyclic feed-forward network on $n(T+1)$ units whose law of (layer-$T$ units at time $T$) given (layer-$0$ units at time $0$) equals the law of $(S(T)\mid S(0))$ under $K$, and whose integrated information is zero at every state.
\end{proposition}
\begin{proof}
Let layer-$0$ units be fair coins with no inputs, whose state at time $0$ is set by the intervention, and let unit $i$ of layer $t\ge1$ update from layer $t-1$ with $K_i$. Because $K$ is a product kernel, layer $t$ at time $t$ has law $K(\cdot\mid\text{layer }t-1\text{ at time }t-1)$, and induction down to layer $0$ at time $0$ reproduces the $T$-step law. A layer-$0$ unit has no inputs, so the cut severing all connections into it leaves the transition law unchanged; its loss is zero and $\vphi=0$.
\end{proof}

\subsection{Marginalization and two-step moments}\label{app:marg}\label{app:moments}
For a product kernel, observing coordinates $O$ gives $Q(y_O\mid x)=\prod_{i\in O}K_i(y_i\mid x)$; summing an unobserved factor over its output gives one, so its parameters vanish from every one-step design. For the parity network in which every unit reads all the others, started from all spins $+1$, expanding a product over a recorded subset $U$ at time two shows that each first-step noise spin $Z_j(1)$ appears $|U|-\mathbf 1\{j\in U\}$ times; even powers equal one, so
\[
\mathbb E\Bigl[\textstyle\prod_{i\in U}S_i(2)\Bigr]=\begin{cases}\prod_{i\in U}a_i^2,&|U|\text{ even},\\ \prod_{i=1}^{n}a_i,&|U|\text{ odd},\end{cases}
\]
for a network of any size. A common sensor spin of mean $1-2\eta$ preserves even moments and multiplies odd moments by $1-2\eta$. Pair moments $u_{ij}=a_i^2a_j^2$ identify each recorded fidelity, $a_i=(u_{ij}u_{ik}/u_{jk})^{1/4}$, and a calibrated odd moment identifies $\prod_ia_i$, from which one hidden fidelity follows by division. Lowering the hidden unit's reliability by $1/32$ from $421/512$ leaves every one-step word probability unchanged; after two steps it moves only the words $000$ and $111$, in which all recorded units agree, by $\mp0.0080$ at sensor error $0.02$ ($\mp0.0067$ at $\eta=0.10$).

\subsection{The identification floor}\label{app:floor}
\begin{proposition}[Identification floor]\label{thm:floor}
If two mechanisms induce the same data distribution under every action of a (possibly adaptive) design and their values of $\vphi$ differ by $D$, then every interval procedure with coverage $1-\alpha$ under both has expected width at least $(1-2\alpha)D$, at every sample size.
\end{proposition}

Let $P^a_\theta$ be the observation law after action $a$. If $P^a_\theta=P^a_{\theta'}$ for every allowed action, then by induction over steps the transcript laws coincide at every finite horizon. Let an interval $I$ cover $\vphi(\theta)$ and $\vphi(\theta')$ each with probability at least $1-\alpha$. Under the common transcript law, the union bound gives $\Pr\{\vphi(\theta),\vphi(\theta')\in I\}\ge1-2\alpha$, on which event $|I|\ge|\vphi(\theta)-\vphi(\theta')|$; hence $\mathbb E|I|\ge(1-2\alpha)D$. For the baseline design, $D$ is the range of $\vphi$ over the hidden reliability $p_4\in[1/2,1]$. With recorded reliabilities $421/512$, the exact four-way tie gives $0.9\times1.1293=1.016$ bits. On a grid of 200{,}001 values that excludes it, the floor is $0.940$ bits (maximum $1.0441$ at $p_4=0.8723$, where the three recorded units remain tied); under the alternative convention that breaks ties toward the smaller raw loss, $\vphi=\vphiraw$, whose maximum $0.399$ bits at $p_4=1$ gives a floor of $0.359$ bits. With no ties, it is $0.299$ bits for recorded reliabilities $(0.78,0.80,0.83)$ and $0.336$ bits for $(0.80,0.82,0.84)$; for $(0.78,0.80,0.83)$ the exact tie at $p_4=0.78$ raises it to $0.467$ bits.

\subsection{The product symmetry}\label{app:product}
\begin{theorem}[Product symmetry]\label{thm:product}
In a parity network in which every unit reads all the others, with state-independent reliabilities and started from a known state, the joint law of the recorded units at \emph{any} set of readout times depends on the fidelities of the hidden units only through their product. A common sensor error on the recorded bits does not change this.
\end{theorem}

Let $R$ be the recorded units and $H$ the $k\ge1$ hidden units, $W(t)=\prod_{h\in H}S_h(t)$ and $P_R(t)=\prod_{j\in R}S_j(t)$. For a recorded unit $i$, $S_i(t+1)=Z_i(t+1)\prod_{j\in R,j\neq i}S_j(t)\cdot W(t)$, so hidden spins enter recorded updates only through $W$. For a hidden unit, $S_h(t+1)=Z_h(t+1)P_R(t)W(t)S_h(t)$; multiplying over $H$,
\[
W(t+1)=\Bigl(\textstyle\prod_{h\in H}Z_h(t+1)\Bigr)P_R(t)^kW(t)^{k+1},
\]
which equals $\prod_hZ_h(t+1)W(t)$ for even $k$ and $\prod_hZ_h(t+1)P_R(t)$ for odd $k$. Thus $(S_R(t),W(t))$ is a Markov chain driven by the recorded noise and by the single spin $\prod_hZ_h(t+1)$, whose law is fixed by its mean $\prod_ha_h$. The initial $W(0)$ is known, so the law of every recorded trajectory depends on the hidden fidelities only through their product. A common sensor flip is independent of the network and of the hidden fidelities and complements all recorded bits of a trial together, so the observed law is a fixed mixture of transformed recorded laws and likewise depends on the hidden fidelities only through their product. Setting a hidden unit by intervention makes $W$ at that time no longer a function of the chain, which is why a single intervention breaks the symmetry; exact enumeration confirms both statements. With recorded reliabilities $(0.78,0.80,0.83)$ and hidden fidelity pairs $(0.40,0.80)$ and $(0.50,0.64)$, recorded trajectories over horizons one to four differ by at most $1.1\times10^{-16}$ in total variation, whereas a pair with a different product differs by up to $0.015$; setting hidden unit 4 or 5 at the first step and reading the second gives $0.018$ and $0.011$. With recorded reliabilities $421/512$ the corresponding values are $1.4\times10^{-16}$, $0.016$, $0.021$ and $0.013$, and an equal-product triple among six units likewise agrees to $10^{-16}$.

\subsection{Edges, envelopes and two-point bounds}\label{app:edges}\label{app:envelope}
\begin{proposition}[Selection switches create quantized jumps]\label{prop:edges}
Suppose that near a hypersurface in parameter space the specified cause and effect states are unique, cut $a$ uniquely minimizes $L_C/e_C$ on one side and cut $b$ on the other, $e_a<e_b$, the two tie on the surface with $L_b>0$, and $I\ge L_b$ there. Then $\vphi$ jumps across the surface from $L_a$ to $L_b=(e_b/e_a)L_a$; relative to the larger side, the jump is $1-e_a/e_b$.
\end{proposition}

\begin{proposition}[One-sided limit]\label{thm:envelope}
Let $\Theta$ be a model class on which the law of the data under the design varies continuously in total variation, and let $\vphi_\ast$ be the lower semicontinuous envelope of $\vphi$ on $\Theta$. At every $\theta\in\Theta$, any lower confidence bound $L$ with coverage $1-\alpha$ uniformly over $\Theta$ satisfies $\Pr_\theta\{L>\vphi_\ast(\theta)\}\le\alpha$. In the parity family with four free reliabilities, $\vphi_\ast=\vphiraw$; if the reliabilities are known to be equal, $\vphi_\ast=\vphi$.
\end{proposition}

\paragraph{Proof of \cref{prop:edges}.} On one side of the switching surface the selected cut is $a$ and $\vphi=\min\{I,L_a\}$; on the other it is $b$ and $\vphi=\min\{I,L_b\}$. Uniqueness of the specified states makes $L_a$, $L_b$ and $I$ continuous at the surface, so the one-sided limits are their values there. At the surface $L_a/e_a=L_b/e_b$, so $L_a=(e_a/e_b)L_b<L_b\le I$, and the one-sided limits are $L_a$ and $L_b$; they differ by $(1-e_a/e_b)L_b$. If the cap binds on both sides there is no jump, and if it binds only on the side of $b$ the jump is smaller than $1-e_a/e_b$; the one exception among the 720 partition-switch edges of our four-unit scans is such a case ($4{:}9$, $32.6\%$ instead of $55.6\%$). The same reasoning shows that every select-then-report step can create jumps, including the choice of specified cause and effect states; those jumps are not quantized, and in our random networks 128 of 129 took $\vphi$ to or from zero.

\paragraph{Proof of \cref{thm:envelope}.} Define $\vphi_\ast(\theta)=\lim_{\epsilon\to0}\inf\{\vphi(\theta'):\theta'\in\Theta,\ \|\theta'-\theta\|<\epsilon\}$; it lies below $\vphi$, is lower semicontinuous on $\Theta$ and is the largest such function. Let $L$ have coverage $1-\alpha$ uniformly over $\Theta$. For $t>\vphi_\ast(\theta)$ choose $\theta_k\in\Theta$, $\theta_k\to\theta$, with $\vphi(\theta_k)<t$; then $\Pr_{\theta_k}(L>t)\le\Pr_{\theta_k}\{L>\vphi(\theta_k)\}\le\alpha$, and continuity of the data law in total variation transfers the bound to $\theta$. Letting $t\downarrow\vphi_\ast(\theta)$ gives $\Pr_\theta\{L>\vphi_\ast(\theta)\}\le\alpha$ (compare ref.~\citenum{donoho1988one}). In the parity family with four free reliabilities, $\vphiraw$ is continuous, lies below $\vphi$, and equals $\vphi$ on the dense set of laws with a unique weakest unit, so $\vphi_\ast=\vphiraw$. If the reliabilities are known to be equal, $\vphi$ is continuous on that one-dimensional class ($r=4$ throughout) and $\vphi_\ast=\vphi$. At the equal-reliability source, an interval that also covers $\vphi$ must contain $[\vphiraw,\vphi]$ with probability at least $1-2\alpha$, giving expected width at least $0.9\times(1.1293-0.3281)=0.721$ bits under every design.

\paragraph{Two-point bound.} For two laws whose targets differ by more than $2\delta$, any estimator with error below $\delta$ under both would solve the testing problem, whose error sum is at least $1-\mathrm{TV}$. With $N$ transitions and per-row divergence at most $d$, the chain rule and Pinsker's inequality give $\max_j\Pr_j\{|\hat\vphi-\vphi_j|\ge\delta\}\ge\{1-\min(1,\sqrt{Nd/2})\}/2$ \citep{yu1997assouad}. For the equal-reliability source and the same source with one reliability lowered by $15/4096$ ($d=4.54\times10^{-5}$ nats), $N=1{,}000$ gives failure probability at least $0.42$. For the two laws separated only by the software tolerance, rational enclosures give $d\approx1.68\times10^{-24}$ nats, so $0.2$-bit accuracy with failure probability $0.05$ requires more than $9.66\times10^{23}$ transitions.

\subsection{Invertible parity mechanisms: reduced formula, exact null and direct certificate}\label{app:direct}
Let unit $i$ output the parity of the units in its input mask $M_i\neq\emptyset$ correctly with probability $p_i\in(1/2,1)$, independently of the past state, and let the noiseless map be a bijection of $\{0,1\}^n$. Write $\lambda_i=\log_2(2p_i)\in(0,1)$, $\lambda(S)=\sum_{i\in S}\lambda_i$ and $q=\prod_ip_i=2^{\lambda([n])-n}$. A cut $C$ disables the set $S_C$ of units that lose at least one parity input and severs $e_C$ connections; we assume that every cut disables at least one unit, as it does when the wiring is strongly connected. Among cuts with the same $S$, only the one with the most severed connections can be selected, which leaves one \emph{form} $(S,e_S)$ per set. Call a form \emph{essential} if its score $\lambda(S)/e_S$ is the strict minimum on some open set of mechanisms, and let $P_S\subset[0,1]^n$ be the closed polyhedral cone on which it is a minimum.

\begin{theorem}[Exact null of the claim $\vphi>c$]\label{thm:null}
For an invertible parity mechanism as above,
\begin{equation}\label{eq:phired}
\vphi(\lambda)=\min\bigl\{\,q\,\lambda(S^\ast),\ n-\lambda([n])\,\bigr\},\qquad S^\ast\in\argmin_S\frac{\lambda(S)}{e_S},
\end{equation}
with ties broken toward the larger $\lambda(S)$. For every $c>0$, the closure in $[0,1]^n$ of the null $\{\lambda\in(0,1)^n:\vphi(\lambda)\le c\}$ is
\begin{equation}\label{eq:null}
\bigl\{\lambda:\ \lambda([n])\ge n-c\bigr\}\ \cup\!\!\bigcup_{S\ \mathrm{essential}}\!\!\bigl\{\lambda\in P_S:\ 2^{\lambda([n])-n}\,\lambda(S)\le c\bigr\}.
\end{equation}
For the network with input masks $(6,12,15,7)$ there are exactly six essential forms: the four single units, each with three severed connections, and the triples $\{1,3,4\}$ and $\{2,3,4\}$, each with ten.
\end{theorem}

\begin{proposition}[One-dimensional search]\label{prop:search}
With $s_i$ correct updates of unit $i$ in $N_i$ recorded transitions, the log-likelihood $\ell(\lambda)=\sum_i\{s_i(\lambda_i-1)\ln2+(N_i-s_i)\ln(1-2^{\lambda_i-1})\}$ is concave. On the branch of an essential form $S$, fixing $t=\lambda(S)$ turns the constraint in \cref{eq:null} into the linear constraint $\lambda([n])\le n+\log_2(c/t)$. The supremum of the likelihood over the closed null is therefore the largest of one concave program (the cap) and, for each essential form, a one-dimensional supremum over $t$ of concave programs over polytopes; weak duality gives a certified upper bound on each.
\end{proposition}

\paragraph{Proof of \cref{eq:phired}.} For every next state $y$, as the past state ranges over all states the correct output ranges over all states, so $U(y)=2^{-n}$ and $\sum_zK(x\mid z)=1$. The correct output has probability $E(y^\ast)=q$ and every other output a strictly smaller probability, and $u\mapsto u\log_2(2^nu)$ is increasing for $u\ge2^{-n}/e$ and negative below $2^{-n}$, so the specified effect is the correct output; likewise the specified cause is the unique preimage of $x$, with $B(z^\ast)=q$. Both specifications equal $q\log_2(2^nq)=q\,\lambda([n])$ and both differentiation terms equal $-\log_2q=n-\lambda([n])$. Under a cut, a unit that loses at least one input in its mask outputs a fair coin, and a unit whose severed inputs lie outside its mask is unchanged. With $S_C$ the set of units of the first kind, $K_C(y^\ast\mid x)=K_C(x\mid z^\ast)=q\prod_{i\in S_C}(2p_i)^{-1}$, so the effect and cause losses both equal $L_C=q\,\lambda(S_C)$. The selected cut minimizes $\lambda(S_C)/e_C$, and $q\lambda(S)\le q\lambda([n])$, which leaves the cap $n-\lambda([n])$. Among cuts with the same $S$, the score is smallest for the largest $e$, strictly so because $\lambda(S)>0$; two forms with equal score vectors have equal $S$ and $e$.

\paragraph{Proof of \cref{thm:null}.} Write $\sigma_S(\lambda)=\lambda(S)/e_S$ and $\rho_S(\lambda)=2^{\lambda([n])-n}\lambda(S)$, and let $\mathcal N$ be the right-hand side of \cref{eq:null} within $[0,1]^n$; it is closed. \emph{The closure is contained in $\mathcal N$.} Let $\vphi(\lambda)\le c$ with $\lambda\in(0,1)^n$. If $n-\lambda([n])\le c$, $\lambda$ is in the cap set. Otherwise the selected loss is at most $c$; by the tie rule it is the largest loss among the forms that minimize $\sigma$ at $\lambda$, so all of them have $\rho_S(\lambda)\le c$. The forms that attain the minimum at $\lambda$ are the only ones that attain it near $\lambda$, and their minimum regions are closed and cover a neighborhood of $\lambda$; one of them therefore has interior, on an open part of which its linear score is strictly below the others. That form is essential, and $\lambda\in P_S$. \emph{$\mathcal N$ is contained in the closure.} A point of the cap set is a limit of points of the open cube that satisfy the cap inequality (move toward $(1-\eta)\mathbf 1$ with $\eta\le c/n$), all of which have $\vphi\le c$. Let $\lambda\in P_S$ with $\rho_S(\lambda)\le c$ for an essential $S$, and let $\lambda^\circ$ be an interior point at which $\sigma_S$ is the strict minimum. If $\rho_S(\lambda)=c$, first replace $\lambda$ by $r\lambda$ with $r<1$ close to one: $P_S$ is a cone, and $\rho_S(r\lambda)<\rho_S(\lambda)$ because $c>0$. The points $(1-\epsilon)\lambda+\epsilon\lambda^\circ$ lie in the open cube, have $S$ as the strict minimizer by linearity, and satisfy $\rho_S<c$ for small $\epsilon$; at such points $\vphi\le\rho_S<c$. \emph{The six forms.} Enumerating the 150 cuts of the network with masks $(6,12,15,7)$ and keeping the largest $e$ for each of the 15 sets $S$ gives $e_S=3|S|$ except for $\{1,3,4\}$ and $\{2,3,4\}$, which have $e_S=10$. A form with $e_S=3|S|$ has score equal to the average of the single-unit scores $\lambda_i/3$ over $S$ and is never strictly below all of them, so the essential forms are among the four single units and the two triples. Each of these is the strict minimizer at an explicit point, hence on an open set: unit $i$ alone at $\lambda_i=0.1$ with the other three at $0.9$; $\{2,3,4\}$ at $\lambda=(0.6,0.5,0.5,0.5)$, where its score is $0.15$ and every other form's is at least $0.16$; and $\{1,3,4\}$ at $\lambda=(0.5,0.6,0.5,0.5)$.

\paragraph{Proof of \cref{prop:search}.} The second derivative of $\ln(1-2^{\lambda_i-1})$ is $-(\ln2)^2p_i/(1-p_i)^2<0$ and the other term is linear, so $\ell$ is concave. On a branch, $2^{\lambda([n])-n}t\le c$ with $t=\lambda(S)$ fixed is the linear constraint $\lambda([n])\le n+\log_2(c/t)$, and the constraints defining $P_S$ are linear.

\paragraph{Implementation.} For an interval $[t_0,t_1]$ of $t$ the constraints $\lambda(S)\in[t_0,t_1]$ and $\lambda([n])\le n+\log_2(c/t_0)$ (omitted when $t_0=0$) relax the branch, so the maximum of $\ell$ over them bounds the branch on that interval from above. For linear constraints $A\lambda\le d$ and any multipliers $\nu\ge0$, weak duality gives
\[
\sup_{A\lambda\le d,\ \lambda\in[0,1]^n}\ell(\lambda)\ \le\ \nu^{\!\top}d+\sum_i\sup_{0\le z\le1}\bigl\{\ell_i(z)-(A^{\!\top}\nu)_iz\bigr\},
\]
and each one-dimensional supremum is bounded by a tangent to a concave function. Multipliers are proposed from the Karush--Kuhn--Tucker conditions of a numerical primal solve; the bound itself is evaluated in 60-digit interval arithmetic with outward rounding and is valid for any nonnegative multipliers. A priority queue bisects the interval with the largest bound; the certificate is issued when the interval lower bound on $\log R_N$ minus the largest remaining upper bound reaches $\log(1/\alpha)$, and an exhausted search budget returns ``not certified''. The mixture $R_N$ is the product of Beta$(1,1)$ mixtures of the four units, enclosed with interval log-gamma functions.

\subsection{The open two-phase region}\label{app:twophase}
\begin{theorem}[An open two-phase region]\label{thm:twophase}
Let $\lambda^0=(0.68,\,0.83,\,0.60,\,0.72)$, let $U=\{\lambda:\|\lambda-\lambda^0\|_\infty<1/200\}$ and let $g(\lambda)=7\lambda_3-3\lambda_1-3\lambda_4$. For every mechanism in $U$ and every current state:
\begin{enumerate}[leftmargin=1.6em,itemsep=1pt,topsep=2pt,label=(\roman*)]
\item if $g(\lambda)<0$, the cut that severs the three connections into unit~3 from the other units is the unique minimizer of $L_C/e_C$ and $\vphi(\lambda)<0.273$ bits;
\item if $g(\lambda)>0$, a ten-connection cut that disables units 1, 3 and 4 is the unique minimizer and $\vphi(\lambda)>0.870$ bits;
\item the four reliabilities are distinct, the specified cause and effect are unique and the cap is inactive.
\end{enumerate}
The same two-phase structure persists in a neighborhood of $\lambda^0$ in the full model in which all 64 state-dependent transition probabilities vary independently: there, for a declared current state, a Lipschitz hypersurface separates a region with $\vphi<1/2-\epsilon$ from a region with $\vphi>1/2+\epsilon$ for some $\epsilon>0$, with a unique minimizing cut on each side.
\end{theorem}

\paragraph{Proof of \cref{thm:twophase}.} By \cref{eq:phired} the selected cut minimizes the linear score $\sigma_C(\lambda)=\lambda(S_C)/e_C$. Let $C_L$ sever the three connections into unit~3 from the other units ($S=\{3\}$, $e=3$) and let $C_H$ be the ten-connection cut with $S=\{1,3,4\}$, so $\sigma_L=\lambda_3/3$ and $\sigma_H=(\lambda_1+\lambda_3+\lambda_4)/10$. At $\lambda^0=(17/25,83/100,3/5,18/25)$ both equal $1/5$, and exact rational enumeration shows that every other one of the 150 cuts has score at least $1/5+1/75$. Each affected unit has at least one severed input, so the coefficient vector of any score has $\ell_1$ norm at most one, and that of $\sigma_L$ has norm $1/3$. For $\|\lambda-\lambda^0\|_\infty<1/200$ the difference between any other score and $\sigma_L$ therefore changes by less than $(1+1/3)/200=1/150$ and stays above $1/150$. Since $\sigma_L-\sigma_H=g/30$, the unique minimizer is $C_L$ where $g<0$ and $C_H$ where $g>0$. On $U$, $\lambda([4])\in(2.81,2.85)$, so $q\in(2^{-1.19},2^{-1.15})$; $\lambda_1+\lambda_3+\lambda_4\in(1.985,2.015)$ and $\lambda_3\in(0.595,0.605)$. Hence the high-phase loss exceeds $2^{-1.19}\times1.985=0.8700$ and is below $2^{-1.15}\times2.015=0.908$, the low-phase loss is below $2^{-1.15}\times0.605=0.2727$, and the cap $\min\{q\lambda([4]),4-\lambda([4])\}$ exceeds $1.15$, so it is inactive. The four coordinate ranges are disjoint, so the reliabilities are distinct, and the specified states are unique by \cref{app:direct}. The values do not depend on the current state.

\paragraph{Persistence with 64 free probabilities.} Let all $r_{x,i}$ (past state $x$, unit $i$) vary independently, $\lambda_{x,i}=\log_2(2r_{x,i})$, and fix a declared current state. At the base point (all rows equal to $\lambda^0$) the specified states are unique, $C_L$ and $C_H$ beat every other cut by a positive margin, the cap is inactive, and the two candidate values $0.267$ and $0.889$ bits straddle $1/2$; all of these are strict inequalities between continuous functions of the 64 parameters and persist on a neighborhood. There, cause and effect losses of a cut need no longer coincide, and the selection is decided by the sign of $G=\min\{L^{E}_{L},L^{B}_{L}\}/3-\min\{L^{E}_{H},L^{B}_{H}\}/10$, a locally Lipschitz function ($E$: effect, $B$: cause). Let $t$ be the coordinate along the direction that raises $\lambda_{x,3}$ equally for all $x$ and $z$ the 63 complementary coordinates. Each of the four analytic differences $L^{\cdot}_{L}/3-L^{\cdot}_{H}/10$ has $t$-derivative $7q_0/30>0$ at the base point, $q_0=2^{-1.17}$, because on the state-independent line both losses of a cut equal $q\lambda(S)$. On a smaller neighborhood all four derivatives lie in an interval $[a,b]$ with $a>0$, so along every line of constant $z$ the function $G$ is absolutely continuous with derivative in $[a,b]$ almost everywhere, hence strictly increasing with a unique zero $t_0(z)$; $t_0$ is Lipschitz because $G$ is Lipschitz in $z$ and increases at rate at least $a$ in $t$. Above the graph $C_H$ is the unique minimizer and $\vphi>1/2+\epsilon$; below it $C_L$ is and $\vphi<1/2-\epsilon$.

\subsection{A universal price of evidence at an edge}\label{app:edgeprice}
The inverse-square law of \cref{thm:price} is not special to integrated information. It holds next to every smooth edge of every claim, with a constant given by Fisher information.

\begin{theorem}[Price of evidence at an edge]\label{thm:edgeprice}
Let mechanisms be indexed by an open set $\Theta\subset\mathbb R^d$, let the design have finitely many actions with finitely many outcomes, and let $\theta^0\in\Theta$. Assume:
\begin{enumerate}[leftmargin=1.9em,itemsep=1pt,topsep=2pt,label=(\alph*)]
\item on a neighborhood $N$ of $\theta^0$ every outcome probability $P^a_\theta(y)$ is positive and twice continuously differentiable in $\theta$, with Fisher information matrices $J_a(\theta)$, and $\sum_aJ_a(\theta^0)$ is positive definite;
\item within $N$, the closure of the falsifying mechanisms is $\{\theta\in N: g(\theta)\le0\}$ for a continuously differentiable $g$ with $g(\theta^0)=0$ and $\nabla g(\theta^0)\ne0$;
\item away from $\theta^0$ the falsifying mechanisms are separated from it: $\inf_{\theta'\in\cl{\Hnull}\setminus N'}\max_a\TV(P^a_{\theta^0},P^a_{\theta'})>0$ for every neighborhood $N'$ of $\theta^0$.
\end{enumerate}
Then, as $\theta\to\theta^0$ within $\{g>0\}$,
\begin{equation}\label{eq:edgeprice}
D^\ast(\theta)=\frac{g(\theta)^2}{2V^\ast}\,\{1+o(1)\},\qquad
\frac1{V^\ast}=\min_{\delta:\,\nabla g(\theta^0)^{\!\top}\delta=1}\ \max_a\ \delta^{\!\top}J_a(\theta^0)\,\delta
=\max_{w}\ \frac{1}{\nabla g(\theta^0)^{\!\top}J_w(\theta^0)^{-1}\nabla g(\theta^0)},
\end{equation}
where $J_w=\sum_aw_aJ_a$, the last maximum is over allocations $w$ with $J_w$ positive definite (a supremum if no maximizer has this property), and, by the argument of \cref{thm:price0}(ii), every level-$\alpha$ certificate that asserts the claim with probability one at $\theta$ needs at least $2V^\ast g(\theta)^{-2}\log(1/\alpha)\{1+o(1)\}$ trials in expectation. The price depends on the mechanism only through the selection gap $g(\theta)$; the margin $F(\theta)-c$ does not enter.
\end{theorem}

\begin{proof}
Write $\nabla g=\nabla g(\theta^0)$, $J_a=J_a(\theta^0)$, $h=g(\theta)$ and $\delta=\theta-\theta'$. By (a) and Taylor's theorem there is a function $\omega(r)\to0$ as $r\to0$ such that, for $\theta,\theta'$ within distance $r$ of $\theta^0$ and every action,
$\bigl|\KL(P^a_\theta\|P^a_{\theta'})-\tfrac12\delta^{\!\top}J_a\delta\bigr|\le\omega(r)\|\delta\|^2$ and $\bigl|g(\theta)-g(\theta')-\nabla g^{\!\top}\delta\bigr|\le\omega(r)\|\delta\|$.
The two expressions for $1/V^\ast$ agree by the minimax theorem, because $\delta^{\!\top}J_w\delta$ is convex in $\delta$ and linear in $w$, the simplex is compact, and for positive definite $J_w$ the inner minimum over $\nabla g^{\!\top}\delta=1$ equals $1/(\nabla g^{\!\top}J_w^{-1}\nabla g)$. Let $\delta^\ast$ attain the first minimum; it exists because $\max_a\delta^{\!\top}J_a\delta\ge\delta^{\!\top}(\sum_aJ_a)\delta/|A|$ is coercive.

\emph{Upper bound.} Fix $\eta>0$ and let $\theta'=\theta-(1+\eta)h\,\delta^\ast$. For $\theta$ close to $\theta^0$, $g(\theta')\le h-(1+\eta)h+\omega\,(1+\eta)h\|\delta^\ast\|\le0$, so $\theta'\in\cl{\Hnull}$, and for every allocation $w$
\[
D_w(\theta)\le\sum_aw_a\KL(P^a_\theta\|P^a_{\theta'})\le(1+\eta)^2h^2\Bigl\{\tfrac12\max_a\delta^{\ast\top}J_a\delta^\ast+\omega\|\delta^\ast\|^2\Bigr\}.
\]
Taking the maximum over $w$, then $\theta\to\theta^0$ and $\eta\to0$, gives $\limsup D^\ast(\theta)/h^2\le1/(2V^\ast)$.

\emph{Lower bound.} Fix $\epsilon\in(0,1)$ and a radius $r>0$ small enough that $2\omega(r)\le\epsilon\lambda_0/2$, where $\lambda_0>0$ is the smallest eigenvalue of $|A|^{-1}\sum_aJ_a$. Let $w^\ast$ attain $\max_w\min_{\nabla g^{\top}\delta=1}\delta^{\!\top}J_w\delta$ and let $w^\epsilon=(1-\epsilon)w^\ast+\epsilon u$ with $u$ uniform over actions, so that $J_{w^\epsilon}\succeq\epsilon\lambda_0I$. By (c), Pinsker's inequality and continuity of $P^a_\theta$ in total variation, there is $\kappa>0$ (depending on $r$) such that every $\theta'\in\cl{\Hnull}$ farther than $r$ from $\theta^0$ has $\sum_aw^\epsilon_a\KL(P^a_\theta\|P^a_{\theta'})\ge\epsilon\kappa$ for all $\theta$ near $\theta^0$; this exceeds $h^2/V^\ast$ once $h$ is small. For $\theta'\in\cl{\Hnull}$ within distance $r$ of $\theta^0$, where $g(\theta')\le0$,
\[
\sum_aw^\epsilon_a\KL(P^a_\theta\|P^a_{\theta'})\ \ge\ \tfrac12\delta^{\!\top}J_{w^\epsilon}\delta-\omega(r)\|\delta\|^2\ \ge\ \Bigl(1-\frac{2\omega(r)}{\epsilon\lambda_0}\Bigr)\tfrac12\delta^{\!\top}J_{w^\epsilon}\delta .
\]
If the left-hand side is at most $h^2/V^\ast$, then $\|\delta\|\le C_\epsilon h$ with $C_\epsilon=2(\epsilon\lambda_0V^\ast)^{-1/2}$, and $g(\theta')\le0$ gives $\nabla g^{\!\top}\delta\ge h-\omega(r)\|\delta\|\ge h\{1-\omega(r)C_\epsilon\}$. Minimizing the quadratic form under this linear constraint, and using $\min_{\nabla g^{\top}\delta=1}\delta^{\!\top}J_{w^\epsilon}\delta\ge(1-\epsilon)\min_{\nabla g^{\top}\delta=1}\delta^{\!\top}J_{w^\ast}\delta=(1-\epsilon)/V^\ast$, every falsifying mechanism is at divergence at least
\[
\Bigl(1-\frac{2\omega(r)}{\epsilon\lambda_0}\Bigr)\bigl\{1-\omega(r)C_\epsilon\bigr\}_+^2\,\frac{(1-\epsilon)\,h^2}{2V^\ast}
\]
from $\theta$ under $w^\epsilon$, and $D^\ast(\theta)\ge D_{w^\epsilon}(\theta)$ is at least this value. Letting $\theta\to\theta^0$, then $r\to0$ and then $\epsilon\to0$ gives $\liminf D^\ast(\theta)/h^2\ge1/(2V^\ast)$.
\end{proof}

\paragraph{Instances.} (i)~\emph{Integrated information.} In the two-phase region of \cref{thm:twophase}, $g=7\lambda_3-3\lambda_1-3\lambda_4$ and a transition records all four units, a single action with diagonal Fisher information $J_i=(\ln2)^2p_i/(1-p_i)$; then $V^\ast=\sum_i(\nabla g)_i^2/J_i$, which is \cref{thm:price} (with $g=7h$ on the path $\lambda^0+he_3$). If one unit is recorded per transition there are four actions, and $V^\ast=\min_w\sum_i(\nabla g)_i^2/(w_iJ_i)=(\sum_i|(\nabla g)_i|/\sqrt{J_i})^2$. (ii)~\emph{Circuits.} For the claim $\max_S\mu_S>\tau$ with one candidate queried per trial, at a table in which a single candidate $S$ sits on the threshold and all others are below it, $g=\mu_S-\tau$, the action that queries $S$ has $J=1/\{\tau(1-\tau)\}$ and $V^\ast=\tau(1-\tau)$: $D^\ast=\Delta^2/\{2\tau(1-\tau)\}$, the leading term of \cref{prop:witnessrate}. (iii)~\emph{Any select-then-report quantity.} If a quantity reports $r_j$ for the candidate $j$ that minimizes a smooth score $\sigma_j$, and near $\theta^0$ exactly two candidates $a,b$ compete with $r_a<c<r_b$ and $\nabla(\sigma_a-\sigma_b)(\theta^0)\ne0$, the closed null of the claim is $\{\sigma_b-\sigma_a\ge0\}$ and $g=\sigma_a-\sigma_b$: the reported values $r_a,r_b$ set the size of the cliff, and only the scores set the price.

\paragraph{Numerical check.} On 24 random instances with two to four Bernoulli cells, one to four actions that each observe a subset of the cells, and linear or quadratic edges, direct optimization of $D^\ast$ (an outer search over allocations and an inner minimization over the falsifying mechanisms) agrees with $g^2/2V^\ast$ to within $2.1\%$, $1.1\%$, $0.5\%$ and $0.3\%$ at selection gaps of $0.02$, $0.01$, $0.005$ and $0.0025$, consistent with an error of order $g$.

\begin{corollary}[Heavy-tailed budgets at any smooth edge]\label{cor:tail}
Let the assumptions of \cref{thm:edgeprice} hold at every point of a relatively open piece of the edge $\{g=0\}$ that contains the compact set $\Gamma$, let mechanisms be drawn from a density $f$ that is continuous and positive near $\Gamma$, and let $T_\Gamma$ be the set of mechanisms with $g>0$ whose nearest point of the edge lies in $\Gamma$. Then the cost coefficient $1/D^\ast$ of the true claims satisfies
\[
\Pr\{\theta\in T_\Gamma,\ 1/D^\ast\ge u\}=\{1+o(1)\}\,u^{-1/2}\int_{\Gamma}\frac{f(\theta)\sqrt{2V^\ast(\theta)}}{\|\nabla g(\theta)\|}\,dS(\theta)\qquad(u\to\infty),
\]
so the expected certification budget over the population is infinite at every level $\alpha$.
\end{corollary}
\begin{proof}
The bounds in the proof of \cref{thm:edgeprice} are uniform over $\Gamma$: the Taylor remainders and the eigenvalue $\lambda_0$ by continuity on a compact set, the separation constant because the falsifying mechanisms farther than $r$ from a point of $\Gamma$ are farther than $r/2$ from every edge point within $r/2$ of it, and $V^\ast$ is continuous on $\Gamma$. Hence, uniformly over $\Gamma$, $1/D^\ast\ge u$ exactly when $0<g\le\sqrt{2V^\ast/u}\{1+o(1)\}$. By the coarea formula the probability of the layer $\{0<g\le s\}$ near $\Gamma$ is $s\int_\Gamma f/\|\nabla g\|\,dS\,\{1+o(1)\}$ as $s\to0$; substituting $s=\sqrt{2V^\ast/u}$ pointwise on $\Gamma$ gives the claim, and $\int^\infty u^{-1/2}du$ diverges.
\end{proof}

\subsection{The price of evidence at a cliff of \texorpdfstring{$\vphi$}{phi}, and heavy tails}\label{app:price}\label{app:tail}
\begin{theorem}[The price of evidence at a cliff]\label{thm:price}
Consider the four-parameter family of \cref{thm:twophase} (known wiring, state-independent reliabilities), a design in which every trial is one transition with all four units recorded, and the claim $\vphi>1/2$. On the path $\lambda_h=\lambda^0+h\,e_3$, $0<h<1/200$, every mechanism has a unique minimizing cut and $\vphi(\lambda_h)-1/2\to0.389$ bits as $h\to0$, yet
\begin{equation}\label{eq:price}
D^\ast(\lambda_h)=0.5939\,h^2+o(h^2),\qquad\text{so that}\qquad \lim_{\alpha\to0}\frac{T^\ast_\alpha(\lambda_h)}{\log(1/\alpha)}=\frac{1.684}{h^{2}}\,\{1+o(1)\},
\end{equation}
where $T^\ast_\alpha$ is the smallest expected number of transitions of any level-$\alpha$ certificate that asserts the claim with probability one; $D^\ast$ is in nats per transition and $h$ in bits of log-reliability. If only one unit can be recorded per transition, $D^\ast(\lambda_h)=h^2/3.975+o(h^2)$. This value is approached by recording units 1, 3 and 4 in proportions $0.22:0.58:0.20$ together with a vanishing share of transitions on unit~2; without that share, $D_w=0$.
\end{theorem}

\begin{theorem}[Heavy-tailed budgets]\label{thm:tail}
In the model with all 64 transition probabilities free and all 16 preparations available, let $N$ be the two-phase neighborhood of \cref{thm:twophase} and let $N'\subset N$ be a smaller neighborhood at positive distance from the boundary of $N$. For a high-phase mechanism in $N'$ let $s$ be its distance from the separating hypersurface along the coordinate $t$ of \cref{app:twophase}. There are constants $0<A\le B$ with $A s^2\le D^\ast\le B s^2$ on $N'$. Consequently, for mechanisms drawn from any density that is bounded above and below on $N'$, the cost coefficient $1/D^\ast$ of the true claims satisfies
\begin{equation}\label{eq:tail}
\Pr\{1/D^\ast\ge u\}=\Theta\bigl(u^{-1/2}\bigr)\qquad(u\to\infty),
\end{equation}
and for every fixed $\alpha$ the expected number of trials of any level-$\alpha$ certificate that certifies the high phase, averaged over the density, is infinite. With a fixed budget of $n$ trials and $\alpha<1-\beta$, no certificate reaches power $1-\beta$ at a high-phase mechanism of $N'$ with $s<\{\kl(1-\beta,\alpha)/(Bn)\}^{1/2}$, although each such mechanism satisfies the claim with a margin of at least $\epsilon$.
\end{theorem}

\paragraph{Proof of \cref{thm:price}.} A transition that records all four units is a single action whose Fisher information about $\lambda$ is diagonal, $J_i=(\ln2)^2p_i/(1-p_i)$. By \cref{thm:twophase,thm:null}, the closed null of the claim $\vphi>1/2$ within $U$ is the half-space $\{g\le0\}$ with $g=7\lambda_3-3\lambda_1-3\lambda_4$. The closed null is a closed subset of the cube $[0,1]^4$, and every mechanism other than $\lambda^0$ has a different law, so the null outside any neighborhood of $\lambda^0$ is separated from it. The assumptions of \cref{thm:edgeprice} therefore hold at $\lambda^0$, and since $g(\lambda_h)=7h$ it gives $D^\ast(\lambda_h)=49h^2/(2V)+o(h^2)$ with $V=\nabla g^{\!\top}J^{-1}\nabla g=9/J_1+49/J_3+9/J_4$. At $\lambda^0$, $p=(0.8011,0.8888,0.7579,0.8236)$ and $J=(1.935,3.842,1.504,2.243)$, so $49/(2V)=0.59394$ and $2V/49=1.68367$. \Cref{thm:price0} converts $D^\ast$ into the stated limit; $\vphi(\lambda_h)\to q_0(\lambda^0_1+\lambda^0_3+\lambda^0_4)=0.8888$ bits, and uniqueness of the minimizing cut for $0<h<1/200$ is \cref{thm:twophase}(ii). (Farther along the path the claim fails: at $h=4/15$ the cut of unit~1 alone ties with the ten-connection cut and then replaces it.)

If one unit is recorded per transition there are four actions, and the action that records unit $i$ has Fisher information $J_ie_ie_i^{\!\top}$. \Cref{thm:edgeprice} gives $V^\ast=\min_w\sum_i(\nabla g)_i^2/(w_iJ_i)=(\sum_i|(\nabla g)_i|/\sqrt{J_i})^2$ by the Cauchy--Schwarz inequality, with minimizer $w_i\propto|(\nabla g)_i|/\sqrt{J_i}=(0.219,0,0.578,0.203)$ and cost constant $2V^\ast/49=3.9748$. This allocation never records unit~2, and with $w_2=0$ exactly $D_w=0$: lowering $\lambda_2$ below $3(\lambda_3+\lambda_4)/7\approx0.57$ leaves the records of units 1, 3 and 4 unchanged, makes the cut of unit~2 alone the minimizer and gives $\vphi=q\lambda_2<1/2$. The value $h^2/3.975$ is therefore a supremum, approached by allocations that give unit~2 a vanishing share; this is the role of $w^\epsilon$ in the proof of \cref{thm:edgeprice}.

\paragraph{Proof of \cref{thm:tail}.} Let all 16 preparations be available and all four units recorded, so that each of the 64 transition probabilities is observed directly as a Bernoulli cell. In the neighborhood $N$ of \cref{app:twophase}, the closed null of the claim $\vphi>1/2$ is the region $\{t\le t_0(z)\}$. \emph{Upper bound.} For a high-phase mechanism $\theta=(z,t_0(z)+s)$ and the boundary point $\theta'=(z,t_0(z))$, reliabilities are bounded away from $1/2$ and $1$, so every action has $\KL(P^a_\theta\|P^a_{\theta'})\le Bs^2$ and $D^\ast\le Bs^2$. \emph{Lower bound.} Under the uniform allocation over preparations, Pinsker's inequality gives a divergence of at least $\tfrac18\sum_j(\theta_j-\theta'_j)^2$ from $\theta$ to any $\theta'$ in the cube $[0,1]^{64}$. A falsifying mechanism inside $N$ is at Euclidean distance at least a constant times $s$, because the graph of $t_0$ is Lipschitz; one outside $N$ is at distance at least $r=\mathrm{dist}(N',\partial N)>0$. Hence $D^\ast\ge c\min\{s,r\}^2\ge As^2$, since $s$ is bounded on $N'$. \emph{Tail.} The event $\{1/D^\ast\ge u\}$ is sandwiched between the layers $\{s\le(Bu)^{-1/2}\}$ and $\{s\le(Au)^{-1/2}\}$, whose probabilities are proportional to their thickness under a density bounded above and below (Fubini over the graph), which is \cref{eq:tail}. By \cref{thm:price0}(ii), a level-$\alpha$ certificate that certifies $\theta$ has $\E_\theta\tau\ge\log(1/\alpha)/(Bs^2)$, and $\int_0s^{-2}ds=\infty$; if it fails to certify a set of positive measure its expected time is infinite there. \emph{Fixed budget.} For any design with $n$ trials, the laws of the record under $\theta$ and $\theta'$ differ by at most $nBs^2$ in divergence. A certificate asserts the claim with probability at most $\alpha$ under $\theta'$, which lies in the closed null (\cref{thm:price0}(i)); if it asserts it with probability at least $1-\beta>\alpha$ under $\theta$, data processing gives $nBs^2\ge\kl(1-\beta,\alpha)$.

\subsection{The robust region and certification by containment}\label{app:region}
\begin{theorem}[A robust region of high integration]\label{thm:region}
In the four-unit parity network with input masks $(6,12,15,7)$, let each of the 64 state-dependent probabilities $p_i(x)$ that unit $i$ outputs the parity of its inputs correctly vary independently in $[1699/2048,\,857/1024]\approx[0.830,0.837]$. For every such transition law and every current state, $\vphi>1.004261$ bits.
\end{theorem}

The second parity network has input masks $(6,12,15,7)$: unit $i$ reads the units whose bits are set in its mask, so unit~1 reads units 2 and 3, unit~2 reads units 3 and 4, unit~3 reads all four units and unit~4 reads units 1--3. Its map from past states to correct next states is invertible. Each unit outputs the parity of its inputs correctly with a probability $p_i(x)$ that may depend on the past state $x$; these 64 probabilities lie in $[l,u]$ with $l=1699/2048$, $u=857/1024$ and $w=u-l$. Summing Bayes factors over the bijective output words bounds the correct output's probability in $[l^4,u^4]$ and its inverse cause's posterior in $[c_-,c_+]$ with $c_-=l^4/\{l^4+(1+w)^4-u^4\}$ and $c_+=u^4/\{u^4+(1-w)^4-l^4\}$. For each cut, let $r$ count units whose parity loses a relevant input and $v$ units whose severed inputs are all irrelevant; with $A=\log_2\{2l/(1+w)\}$, $B=\log_2\{2u/(1-w)\}$ and $U=\log_2(u/l)$, normalized losses lie between $(rA-vU)/e$ and $(rB+vU)/e$ after factoring out the specified probability. Checking all 150 cut signatures leaves exactly two possible minimizers, with incoming masks $(8,13,11,7)$ and $(14,1,11,7)$, both with $(r,v,e)=(3,1,10)$---the two essential triples of \cref{thm:null}; every other cut's lower bound exceeds the larger candidate's upper bound by more than $0.0125$ in these factored units. The scalar then has four uniform lower bounds---selected raw loss $P_-(3A-U)$, cause specification $c_-\log_2(16c_-)$, effect specification $l^4\log_2\{16l^4/(1+w)^4\}$ and differentiation $-\log_2P_+$, with $P_\pm$ the extreme specified probabilities---equal to $1.0042614$, $1.3581242$, $1.3639901$ and $1.0099379$ bits, all above $1.004261431503$ as exact rational inequalities. Competing output words have probability at most $u^3(1-l)$, which keeps the specified states unique. All logarithms are enclosed by the series $\ln t=2\sum_{k\ge0}z^{2k+1}/(2k+1)$, $z=(t-1)/(t+1)$, truncated after 512 terms with an explicit tail bound. Finally, an $m$-bit cause posterior with specified probability $v$ has cause cap at most $\min\{2^{-t}(m-t),t\}$ with $t=-\log_2v$, maximized at $B_m$ with $B_m(1+2^{B_m})=m$; since $B_3=1$, no candidate of at most three units can exceed one bit.

\begin{lemma}[Certification by containment]\label{lem:containment}
Let $\mathcal B$ be a fixed region with $\vphi>c$ on $\mathcal B$, and let $C_N$ be a confidence set with coverage at least $1-\alpha$. The rule ``certify $\vphi>c$ if $C_N\subseteq\mathcal B$'' issues a false certificate with probability at most $\alpha$.
\end{lemma}
\begin{proof}
A false certificate requires $C_N\subseteq\mathcal B$ while the true law has $\vphi\le c$ and so lies outside $\mathcal B$, hence outside $C_N$.
\end{proof}

\subsection{Witness certificates}\label{app:witness}
Each query names one of $K$ candidates; a query of candidate $S$ returns an independent Bernoulli outcome with mean $\mu_S$, and the choice of candidate may depend on all earlier outcomes. After $t$ queries, candidate $S$ has $s_S$ successes in $N_S$ queries. The null of the claim $\max_S\mu_S>\tau$ is $\{\mu:\mu_S\le\tau\ \forall S\}$. Let
\[
U_S=\frac{\int_0^1u^{s_S}(1-u)^{N_S-s_S}du}{\sup_{u\le\tau}u^{s_S}(1-u)^{N_S-s_S}},\qquad
\Lambda_{S,\ell}=\Bigl(\frac{q_\ell}{\tau}\Bigr)^{s_S}\Bigl(\frac{1-q_\ell}{1-\tau}\Bigr)^{N_S-s_S},\quad q_\ell=\tau+(1-\tau)2^{-\ell},
\]
with $U_S=\Lambda_{S,\ell}=1$ before the first query of $S$. The \emph{product certificate} is $E^{\Pi}=\prod_SU_S$, which is \cref{eq:directS} with independent uniform mixtures. The \emph{witness certificate} mixes over candidates instead of multiplying:
\begin{equation}\label{eq:witness}
E^{\mathrm W}_t=\frac1{2K}\sum_{S}\Bigl\{U_S(t)+\sum_{\ell\ge1}\frac{1}{\ell(\ell+1)}\,\Lambda_{S,\ell}(t)\Bigr\}.
\end{equation}
Its first term is robust to large gaps; the second replaces the estimation penalty of a continuous mixture by the price of locating the right scale.

\begin{theorem}[Witness certificates]\label{thm:witness}
For any adaptive choice of queries, $E^{\mathrm W}$ is a certificate of level $\alpha$ for the claim $\max_S\mu_S>\tau$ when stopped at $E^{\mathrm W}_t\ge1/\alpha$. If some candidate $S$ has $\mu_S=\tau+\Delta$ with $0<\Delta\le(1-\tau)/2$ and the statistic is evaluated after every query, the certificate is issued no later than the $T$-th query of $S$, where
\begin{equation}\label{eq:witnessbound}
\E[T]\ \le\ \frac{8V}{3\Delta^{2}}\Bigl\{\log\frac{2K}{\alpha}+\log\bigl[\ell(\ell+1)\bigr]+\log\frac{q_\ell}{\tau}\Bigr\},\qquad \ell=\Bigl\lceil\log_2\frac{1-\tau}{\Delta}\Bigr\rceil,
\end{equation}
with $V=\max_{u\in[\tau,\tau+\Delta]}u(1-u)$: of order $\Delta^{-2}\{\log(K/\alpha)+\log\log(1/\Delta)\}$, with the candidates below threshold entering only through $\log K$. With the grid truncated at $\ell\le L$, as in our implementation ($L=32$), the bound holds for $\Delta\ge(1-\tau)2^{-L}$.
\end{theorem}

\paragraph{Validity.} Fix a null $\mu$. Then $U_S\le M_S:=\int_0^1u^{s_S}(1-u)^{N_S-s_S}du/\{\mu_S^{s_S}(1-\mu_S)^{N_S-s_S}\}$, a mixture likelihood ratio that changes only when $S$ is queried and then has conditional mean one, so $M_S$ is a nonnegative martingale for the full history under any adaptive rule, and so are $\prod_SM_S$ and $K^{-1}\sum_SM_S$. When $S$ is queried, $\Lambda_{S,\ell}$ is multiplied by a factor with conditional mean $\mu_Sq_\ell/\tau+(1-\mu_S)(1-q_\ell)/(1-\tau)=1+(\mu_S-\tau)(q_\ell-\tau)/\{\tau(1-\tau)\}\le1$, so it is a nonnegative supermartingale, and so is any mixture with weights $\{\ell(\ell+1)\}^{-1}$, which sum to one. Hence $E^{\Pi}$ and $E^{\mathrm W}$ are dominated by nonnegative supermartingales started at one, and Ville's inequality gives level $\alpha$ for both. (The implementation truncates the grid at $\ell=32$ and places the remaining weight on the constant one, which preserves validity.)

\paragraph{What the product pays.} For every candidate $S$, $\log E^{\mathrm W}\ge\log U_S-\log(2K)$, whereas $\log E^{\Pi}=\log U_S+\sum_{S'\ne S}\log U_{S'}$. For a candidate with $\mu_{S'}<\tau$, the empirical mean is eventually below $\tau$ and Stirling's formula gives $\log U_{S'}=-\tfrac12\log N_{S'}+\tfrac12\log\{2\pi\mu_{S'}(1-\mu_{S'})\}+o(1)$. A sampler that keeps every candidate at $N_{S'}\approx\sqrt t/K$ queries---the exploration that protects against a wrong early choice---therefore costs the product $(K-1)\{\tfrac14\log t-\tfrac12\log K\}+O(K)$ nats in addition to the witness's own $\tfrac12\log t$: about 200 nats for $K=42$ at $t=4\times10^{11}$, where the claim itself requires $\log(1/\alpha)=3$.

\paragraph{Proof of \cref{eq:witnessbound}.} Let the witness have $\mu_S=p=\tau+\Delta$ and choose $\ell$ with $d_\ell:=(1-\tau)2^{-\ell}\in(\Delta/2,\Delta]$. Since $E^{\mathrm W}\ge\Lambda_{S,\ell}/\{2K\ell(\ell+1)\}$, the certificate is issued once $\log\Lambda_{S,\ell}\ge H:=\log\{2K\ell(\ell+1)/\alpha\}$. In the witness's own queries $\log\Lambda_{S,\ell}$ is a random walk with increments at most $a=\log(q_\ell/\tau)$ and mean $m=p\log(q_\ell/\tau)+(1-p)\log\{(1-q_\ell)/(1-\tau)\}=\int_\tau^{q_\ell}(p-u)/\{u(1-u)\}du>0$. If $T$ is its first passage above $H$, the stopped walk at $T\wedge n$ is at most $H+a$, so Wald's identity gives $m\,\E[T\wedge n]\le H+a$ and, by monotone convergence, $\E T\le(H+a)/m$. Finally $m\ge V^{-1}\int_\tau^{q_\ell}(p-u)du=V^{-1}(\Delta d_\ell-d_\ell^2/2)\ge3\Delta^2/(8V)$, because $d_\ell\in[\Delta/2,\Delta]$.

\paragraph{Other certificates of the same claim.}\label{app:baselines} We compare the witness certificate with four statistics computed from the same counts; each is valid for any adaptive choice of queries. (i)~The \emph{product certificate} $E^\Pi$ above. (ii)~A \emph{one-sided product}, $\prod_SG_S$ with $G_S=(1-\tau)^{-1}\int_\tau^1(u/\tau)^{s_S}\{(1-u)/(1-\tau)\}^{N_S-s_S}du$: each integrand is a nonnegative supermartingale under $\mu_S\le\tau$ by the computation above, so $G_S$ and the product are. This is the product form of multi-armed testing by betting \citep{sandoval2026multiarmed}, with a uniform prior on the bet; it does not pay for estimating means below $\tau$, but each candidate that turns out to be below $\tau$ still costs a factor of order $1/N_S$. (iii)~A \emph{union bound over Bernoulli confidence sequences}, which asserts the claim when $\max_SU_S\ge K/\alpha$: by Ville's inequality each $U_S$ exceeds $K/\alpha$ with probability at most $\alpha/K$. (iv)~A \emph{union bound over sub-Gaussian confidence sequences}, with $U_S$ replaced by the normal mixture $H_S=\int_0^\infty\exp\{l(s_S-\tau N_S)-N_Sl^2/8\}\,2\phi_\rho(l)\,dl$, where $\phi_\rho$ is the centered normal density with standard deviation $\rho$; by Hoeffding's lemma each integrand is a nonnegative supermartingale under $\mu_S\le\tau$. We use $\rho=4(1-\tau)$. The witness certificate dominates (iii) up to the factor two in \cref{eq:witness}; what it adds is the multiscale term, which replaces the $\tfrac12\log N_S$ penalty of the Beta mixture by $\log\{\ell(\ell+1)\}$.

\begin{proposition}[Information distance of an existential claim]\label{prop:witnessrate}
With one candidate queried per trial, the existential claim $\max_S\mu_S>\tau$ has $D^\ast(\mu)=\max_S\kl(\mu_S,\tau)\mathbf 1\{\mu_S>\tau\}$, which equals $\Delta^2/\{2\tau(1-\tau)\}\{1+O(\Delta)\}$ when the largest mean is $\tau+\Delta$. The universal claim $\max_S\mu_S<\tau$ has $D^\ast(\mu)=\{\sum_S1/\kl(\mu_S,\tau)\}^{-1}$: its price is the sum of the prices of the candidates.
\end{proposition}
\begin{proof}
For proportions $w$, the nearest null point truncates each coordinate at $\tau$, so $D_w=\sum_Sw_S\kl(\mu_S,\tau)\mathbf 1\{\mu_S>\tau\}$, a linear function of $w$ maximized at a vertex. For the universal claim the nearest null point raises a single coordinate to $\tau$, so $D_w=\min_Sw_S\kl(\mu_S,\tau)$, which is maximized by equalizing its terms, $w_S\propto1/\kl(\mu_S,\tau)$.
\end{proof}

\subsection{Search and certify: candidates that cannot be enumerated}\label{app:search}
In a large model the class $\mathcal S$ of candidates (all sets of at most $k$ attention heads, say) is too large to list, and a uniform weight $1/|\mathcal S|$ in \cref{eq:witness} would charge $\log|\mathcal S|$ to every claim. The witness certificate does not need the list. Let a \emph{search} be any procedure that, at times of its choosing, proposes a candidate $S_j\in\mathcal S$ and gives it a weight $\omega_j\ge0$, where the candidate, its weight and the time of the proposal may depend on everything observed so far, including earlier queries and any other data, provided that, given all of it, the outcome of each later query of a candidate $S$ is an independent Bernoulli$(\mu_S)$ draw (the search does not see the inputs of future queries). A weight is fixed when its candidate is proposed and is never revised, and $\sum_j\omega_j\le1$ always. Let $E_j(t)=\tfrac12\{U_{S_j}+\sum_{\ell}\Lambda_{S_j,\ell}/[\ell(\ell+1)]\}$ be the witness evidence of the $j$-th proposal computed from the queries made to it after it was proposed, with $E_j=1$ until then, and let
\begin{equation}\label{eq:search}
E^{\mathrm S}_t=1+\sum_{j\,:\,\text{proposed by }t}\omega_j\,\{E_j(t)-1\}.
\end{equation}

\begin{theorem}[Search and certify]\label{thm:search}
For any search and any adaptive choice of queries, $E^{\mathrm S}$ stopped at $E^{\mathrm S}_t\ge1/\alpha$ is a certificate of level $\alpha$ for the claim $\sup_{S\in\mathcal S}\mu_S>\tau$, whatever the size of $\mathcal S$. If the search proposes a candidate with $\mu_S=\tau+\Delta$, $0<\Delta\le(1-\tau)/2$, and gives it weight $\omega$, and the statistic is evaluated after every query, the certificate is issued no later than the $T$-th query made to that candidate after its proposal, where, conditionally on the proposal,
\begin{equation}\label{eq:searchbound}
\E[T]\ \le\ \frac{8V}{3\Delta^{2}}\Bigl\{\log\frac{2}{\omega\alpha}+\log\bigl[\ell(\ell+1)\bigr]+\log\frac{q_\ell}{\tau}\Bigr\},
\end{equation}
with $\ell$ and $V$ as in \cref{thm:witness}. With the weights $\omega_j=1/\{j(j+1)\}$, the multiplicity $\log K$ of \cref{eq:witnessbound} is replaced by $\log\{j(j+1)\}$, where $j$ is the rank at which the search proposes the witness: the price of searching a class is the rank of the answer in the search, not the size of the class.
\end{theorem}

\paragraph{Proof.} Fix a null $\mu$, with $\mu_S\le\tau$ for every $S\in\mathcal S$. By the validity argument of \cref{thm:witness}, the evidence of a fixed candidate, computed from the queries made to it after a given time, is dominated by a nonnegative supermartingale that equals one at that time; because the $j$-th candidate and its weight are determined when it is proposed, the same holds for $E_j$ after the proposal, with a dominating process $\bar E_j$, and $\omega_j\{\bar E_j(t)-1\}$ has nonpositive conditional increments. Hence $\bar E^{\mathrm S}_t=1+\sum_j\omega_j\{\bar E_j(t)-1\}$ is a supermartingale with $\bar E^{\mathrm S}_0=1$; it is nonnegative because $\bar E_j\ge0$ and $\sum_j\omega_j\le1$; and it dominates $E^{\mathrm S}$. Ville's inequality gives the level. For the cost, $E^{\mathrm S}_t\ge\omega E_j(t)\ge\omega\Lambda_{S,\ell}/\{2\ell(\ell+1)\}$ for the proposal in question, and the proof of \cref{eq:witnessbound} applies with $K$ replaced by $1/\omega$.

\paragraph{Remarks.} (i)~Nothing is required of the search: it may use gradients, attribution scores, a discovery sample or the outcomes of the audit itself, and it may be wrong. Evidence about a candidate starts when the candidate is proposed, which is what makes selection harmless. (ii)~The statistic used in the live audits of this paper is $\sum_j\omega_jE_j$, which is at most $E^{\mathrm S}$. (iii)~The comparison with a design that splits the data into a discovery sample and a held-out test of one selected candidate is that the certificate may hold several candidates at once, may add candidates while it runs, and stops by itself. (iv)~Assigning evidence budgets to hypotheses as they are proposed is the idea of alpha-investing and of online testing with e-values \citep{foster2008alpha,xu2024online}, which control error rates over a stream of separate hypotheses; here the stream serves one composite claim, and the object controlled is the probability of asserting it falsely.

\subsection{Certificates of universal claims}\label{app:universal}
\begin{proposition}[Universal claims need no correction for multiplicity]\label{prop:universal}
For a finite list $\mathcal S$, let $E^{-}_S$ be the witness evidence against $\mu_S\ge\tau$, obtained from \cref{eq:witness} with $K=1$ by exchanging successes with failures and $\tau$ with $1-\tau$. Consider the rule that stops querying a candidate once $E^{-}_S\ge1/\alpha$ and asserts $\max_{S\in\mathcal S}\mu_S<\tau$ when every candidate has been retired in this way. It is a certificate of level $\alpha$ for any adaptive choice of queries in which the candidates evaluated on an input are chosen before any outcome on that input is seen; this includes designs that evaluate many candidates on the same inputs. If every $\mu_S\le\tau-\Delta_S$ with $0<\Delta_S\le\tau/2$ and $E^-_S$ is evaluated after every query of $S$, candidate $S$ is retired after at most $T_S$ queries, with $\E[T_S]\le\{8V_S/(3\Delta_S^2)\}\{\log(2/\alpha)+\log[\ell_S(\ell_S+1)]+\log[(1-q^-_{\ell_S})/(1-\tau)]\}$, where $q^-_\ell=\tau(1-2^{-\ell})$, $\ell_S=\lceil\log_2(\tau/\Delta_S)\rceil$ and $V_S=\max_{u\in[\tau-\Delta_S,\tau]}u(1-u)$.
\end{proposition}
\begin{proof}
If the claim is false there is $S_0$ with $\mu_{S_0}\ge\tau$, and the rule can assert the claim only if $E^-_{S_0}$ reaches $1/\alpha$. Given the whole past, including the outcomes of all candidates on earlier inputs, the next recorded outcome of $S_0$ is a Bernoulli$(\mu_{S_0})$ draw, so $E^-_{S_0}$ is dominated by a nonnegative supermartingale started at one for the joint history, however the outcomes of the candidates on a common input depend on each other. Ville's inequality gives the level without a union bound. (The condition on the order of choices matters: a design that queries a second candidate on an input only when a first one failed there records too many failures.) The cost bound is \cref{eq:witnessbound} with $K=1$, applied to each candidate.
\end{proof}
The two halves of a statement about circuit size therefore have different prices. The upper bound ``at most $k$ heads suffice'' is existential and costs the price of its best witness, $\log(1/\alpha)/\kl(\mu_S,\tau)$. The lower bound ``no $k-1$ heads suffice'' is universal and costs the sum of the prices of all smaller sets, in agreement with \cref{prop:witnessrate}; with $\mu_S$ far below $\tau$ for most sets, each of them costs a few queries, and the total is of the order of the number of sets.

\subsection{What a fixed budget can certify}\label{app:budget}
The price of \cref{thm:price0} is an expected number of trials. An experimenter has a budget, and wants to know what it buys. The change of measure that gives the price also bounds the probability of success within any budget, and turns the heavy tail of \cref{cor:tail} into a law for the share of a population that an experiment can resolve.

\begin{proposition}[Budget and power]\label{prop:budget}
Under the hypotheses of \cref{thm:price0}, run a certificate of level $\alpha$ for the claim $F>c$ with any adaptive design, and let $\pi_B(\theta)=\Pr_\theta(\tau\le B)$ be the probability that it asserts the claim within $B$ trials. For every mechanism $\theta$ and every $B$ with $\pi_B(\theta)\ge\alpha$,
\begin{equation}\label{eq:budget}
\kl\bigl(\pi_B(\theta),\alpha\bigr)\ \le\ B\,D^\ast(\theta),\qquad\text{and hence}\qquad \pi_B(\theta)\ \le\ \frac{B\,D^\ast(\theta)+\log2}{\log(1/\alpha)} ,
\end{equation}
where $\kl(x,y)=x\log(x/y)+(1-x)\log\{(1-x)/(1-y)\}$. The second bound holds for every $B$, because $\alpha\log(1/\alpha)<\log2$. A certificate that succeeds within $B$ trials with probability at least $1-\beta>\alpha$ therefore exists only if $B\ge\kl(1-\beta,\alpha)/D^\ast(\theta)$.
\end{proposition}
\begin{proof}
Fix $\theta'\in\Hnull$ and consider the record of the first $\tau\wedge B$ trials, with $N_a$ the number of them that use action $a$. As in the proof of \cref{thm:price0}(ii), Wald's identity gives the divergence between the laws of this record under $\theta$ and $\theta'$ as $\sum_a\E_\theta[N_a]\KL(P^a_\theta\|P^a_{\theta'})$, and the data-processing inequality applied to the event $\{\tau\le B\}$, which the record determines, bounds it below by $\kl(\pi_B(\theta),\pi_B(\theta'))$. Because $\pi_B(\theta')\le\alpha\le\pi_B(\theta)$ and $y\mapsto\kl(x,y)$ decreases on $(0,x]$, this is at least $\kl(\pi_B(\theta),\alpha)$. The divergence is continuous in $\theta'$, with values in $[0,\infty]$, so the inequality extends to $\cl{\Hnull}$. With $w_a=\E_\theta[N_a]/\E_\theta[\tau\wedge B]$ it reads $\kl(\pi_B(\theta),\alpha)\le\E_\theta[\tau\wedge B]\sum_aw_a\KL(P^a_\theta\|P^a_{\theta'})\le B\sum_aw_a\KL(P^a_\theta\|P^a_{\theta'})$ for every $\theta'\in\cl{\Hnull}$, and the infimum over $\theta'$ is $B\,D_w(\theta)\le B\,D^\ast(\theta)$. The second bound follows from $\kl(x,\alpha)\ge x\log(1/\alpha)-\log2$.
\end{proof}

\begin{corollary}[What a budget resolves]\label{cor:resolution}
In the setting of \cref{cor:tail} and under the hypotheses of \cref{thm:price0}, fix $\pi_0\in(\alpha,1)$ and let every mechanism be studied with at most $B$ trials, by any certificates of level $\alpha$ and any designs. The mechanisms of $T_\Gamma$ whose true claim is asserted within the budget with probability at least $\pi_0$ omit a set of probability at least
\begin{equation}\label{eq:resolution}
\{1+o(1)\}\Bigl\{\frac{\kl(\pi_0,\alpha)}{B}\Bigr\}^{1/2}\int_{\Gamma}\frac{f(\theta)\sqrt{2V^\ast(\theta)}}{\|\nabla g(\theta)\|}\,dS(\theta)\qquad(B\to\infty).
\end{equation}
The part of a population that an experiment leaves unresolved is thus bounded below by a floor that shrinks as the inverse square root of the budget: four times the budget at most halves this floor. For a threshold claim about a success probability the rate is attained up to an iterated logarithm. Let a success probability $\mu$ have a density $f$ that is continuous and positive at a threshold $\tau\in(0,1)$ (from here on $\tau$ denotes the threshold of \cref{app:witness}, not a stopping time), and let each $\mu$ be tested for $B$ queries with the certificate of \cref{thm:witness} for the claim $\mu>\tau$ and its mirror image (\cref{prop:universal}) for $\mu<\tau$, both with $K=1$. The probability that $\mu$ falls where the true claim is asserted within the budget with probability below $\pi_0$ is at least $\{1+o(1)\}\,2f(\tau)\{2\tau(1-\tau)\kl(\pi_0,\alpha)/B\}^{1/2}$ and at most $O\bigl(\{[\log(1/\alpha)+\log\log B]/B\}^{1/2}\bigr)$.
\end{corollary}
\begin{proof}
By \cref{prop:budget}, a mechanism with $1/D^\ast>B/\kl(\pi_0,\alpha)$ is asserted within the budget with probability below $\pi_0$, and \cref{cor:tail} with $u=B/\kl(\pi_0,\alpha)$ gives the probability of such mechanisms in $T_\Gamma$. For a success probability $\mu=\tau\pm\Delta$, $D^\ast=\kl(\mu,\tau)=\Delta^2/\{2\tau(1-\tau)\}\{1+O(\Delta)\}$ (\cref{prop:witnessrate}), so the same argument applies on each side of $\tau$ and the two layers $\{0<\pm(\mu-\tau)<\Delta_B\}$, $\Delta_B=\{2\tau(1-\tau)\kl(\pi_0,\alpha)/B\}^{1/2}\{1+o(1)\}$, have total probability $2f(\tau)\Delta_B\{1+o(1)\}$. For the upper bound, \cref{eq:witnessbound} with $K=1$ and Markov's inequality show that the certificate is issued within $B$ queries with probability at least $\pi_0$ once $B(1-\pi_0)\ge\{8V/(3\Delta^2)\}\{\log(2/\alpha)+\log[\ell(\ell+1)]+\log(q_\ell/\tau)\}$ with $\ell=\lceil\log_2\{(1-\tau)/\Delta\}\rceil$ and $V\le1/4$ as in \cref{thm:witness}. For $\Delta\le(1-\tau)/2$ and $B\ge16$ this holds whenever $\Delta\ge C_0\{[\log(1/\alpha)+\log\log B]/B\}^{1/2}$, with a constant $C_0$ that depends on $\tau$ and $\pi_0$ only, because $\log[\ell(\ell+1)]\le2\log\log B+O(1)$ for the gaps in question. For $\Delta>(1-\tau)/2$ the proof of \cref{eq:witnessbound} with $\ell=1$ gives a mean increment of at least $(1-\tau)^2/2$ and so $\E[T]\le2(1-\tau)^{-2}\{\log(4/\alpha)+\log[(1+\tau)/(2\tau)]\}$, which is covered after enlarging $C_0$. The mirror image holds below $\tau$, and the density is bounded near $\tau$.
\end{proof}

\paragraph{Unequal costs.} Trials need not cost the same: a recording is cheaper than an intervention, and a cached forward pass cheaper than a generated answer. Let action $a$ cost $\kappa_a>0$ and let the cost of an experiment be the sum of the costs of its trials. The proof of \cref{thm:price0}(ii) gives $\sum_a\E_\theta[N_a]\KL(P^a_\theta\|P^a_{\theta'})\ge\log(1/\alpha)$ for every $\theta'\in\cl{\Hnull}$. Writing $\mathcal K=\sum_a\kappa_a\E_\theta[N_a]$ for the expected cost and $w_a=\E_\theta[N_a]/\sum_b\E_\theta[N_b]$, the left-hand side equals $\mathcal K\sum_aw_a\KL(P^a_\theta\|P^a_{\theta'})/\sum_aw_a\kappa_a$, so every certificate of level $\alpha$ that asserts the claim with probability one under $\theta$ has expected cost
\begin{equation}\label{eq:costs}
\mathcal K\ \ge\ \frac{\log(1/\alpha)}{D^\ast_\kappa(\theta)},\qquad D^\ast_\kappa(\theta)=\max_w\ \inf_{\theta'\in\cl{\Hnull}}\ \frac{\sum_aw_a\KL(P^a_\theta\|P^a_{\theta'})}{\sum_aw_a\kappa_a} ,
\end{equation}
the information distance per unit of cost. The bound is attained: drawing actions independently from a maximizer $w$, \cref{thm:price0}(iii) and Wald's identity give an expected cost of $\{1+o(1)\}(\sum_aw_a\kappa_a)\log(1/\alpha)/D_w(\theta)=\{1+o(1)\}\log(1/\alpha)/D^\ast_\kappa(\theta)$ as $\alpha\to0$, and the tracking argument of \cref{thm:adaptive} extends to an unknown mechanism once trials are replaced by accumulated cost in its last step. With equal costs \cref{eq:costs} is \cref{thm:price0}(ii); with unequal costs the best design shifts trials towards actions that are informative per unit of cost: an uninformative action drops out, and an expensive one is used only as far as no cheaper action separates the same falsifying mechanisms.

\subsection{What is classical and what is new}\label{app:novelty}
The theory combines tools of sequential analysis with results that are specific to claims about mechanism. \Cref{tab:novelty} separates the two.

\begin{table}[h]\centering\small
\caption{Results of this Supplementary Note, the earlier results they build on, and what each adds.}\label{tab:novelty}
\renewcommand{\arraystretch}{1.2}
\begin{tabularx}{\linewidth}{@{}>{\raggedright\arraybackslash}p{0.24\linewidth}>{\raggedright\arraybackslash}p{0.31\linewidth}>{\raggedright\arraybackslash}X@{}}
\toprule
Result & Builds on & New here\\
\midrule
Certifiability and price (\cref{thm:price0}) & change-of-measure bounds for sequential tests and best-arm identification \citep{chernoff1959sequential,garivier2016optimal,kaufmann2016complexity}; universal inference \citep{wasserman2020universal} & claims about any functional of a mechanism, with falsifying sets that need be neither convex nor smooth; zero distance as the criterion for what an experiment with given access cannot certify\\
Adaptive attainment (\cref{thm:adaptive}) & tracking designs \citep{garivier2016optimal,degenne2019pure} & arbitrary falsifying sets; actions that observe several unknowns at once\\
Budget and power (\cref{prop:budget,cor:resolution}) & the same change of measure \citep{kaufmann2016complexity} & the inverse-square-root law for the share of a population that a budget leaves unresolved\\
Same function, different $\vphi$ (\cref{prop:samefunction,prop:unfolding}) & the unfolding argument \citep{doerig2019unfolding} & an exact pair with one input--output law, and the recording that separates it\\
Hidden units (\cref{thm:floor,thm:product}) & partial identification & exact floors on interval width; the product symmetry and the intervention that breaks it\\
Cliffs of $\vphi$ (\cref{prop:edges,thm:twophase,thm:null}) & the definition of $\vphi$ \citep{albantakis2023iit4} & quantized jumps; an open two-phase set free of ties and symmetry; the exact falsifying set\\
Price at an edge and heavy tails (\cref{thm:edgeprice,cor:tail}) & local expansions of information & a price that depends on the selection gap alone, for any quantity that selects and reports; the tail exponent $-1/2$\\
One-sided limit (\cref{thm:envelope}) & impossibility results for discontinuous functionals \citep{donoho1988one,hirano2012impossibility} & the lower semicontinuous envelope as the certifiable quantity\\
Witness certificates (\cref{thm:witness}) & mixture martingales and confidence sequences \citep{howard2021time}; their use for a given circuit \citep{asiaee2026certified} & evidence averaged over candidates with a multiscale mixture, at a price of $\log K$ and $\log\log(1/\Delta)$\\
Search and certify (\cref{thm:search}) & alpha-investing and online testing with e-values \citep{foster2008alpha,xu2024online} & one composite claim over a class too large to list, at the price of the rank of its witness\\
Universal claims (\cref{prop:universal}) & intersection--union tests \citep{berger1982multiparameter} & validity at any time, with candidates evaluated on shared inputs and no correction for their number\\
\bottomrule
\end{tabularx}
\end{table}

\section{Experimental protocols}\label{app:protocols}

\subsection{Recording designs, budgets and intervals}\label{app:designs}
Three designs record the first parity network after a known intervention (\cref{tab:designs}). An intervention and each update cost one unit, each recorded bit costs $c\in\{0.25,1,4\}$, a calibration trial costs $1+c$, and installing an extra channel costs 512. A total budget $B\in\{8192,32768,131072\}$ is split between calibration (a fraction $f\in\{0.05,0.25,0.5\}$ of what remains after installation) and recording, so that $M=\lfloor f(B-s)/(1+c)\rfloor$ calibration and $N=\lfloor(B-s-M(1+c))/(1+t+kc)\rfloor$ recording trials are run, where $s=512$ for the four-unit design and $0$ otherwise, $t\in\{1,2\}$ is the readout delay and $k$ the number of recorded units. Four sources (equal reliabilities $421/512$; one lowered by $15/4096$; heterogeneous $(0.78,0.80,0.83,0.85)$; weak $(0.56,0.62,0.68,0.74)$), two sensor errors ($0.02$, $0.10$) and all allocations give 648 cells with 256 datasets each (165{,}888 datasets). Unless stated otherwise, results use $\eta=0.10$, $c=1$, $f=0.25$.

\begin{table}[h]\centering\small
\caption{Recording designs in the first parity network.}\label{tab:designs}
\begin{tabular}{@{}llcc@{}}\toprule
Design & Recorded & Cost per trial & Installation\\\midrule
3 units, $t{+}1$ & units 1--3, one update after the intervention & $2+3c$ & 0\\
4 units, $t{+}1$ & all four units, one update after & $2+4c$ & 512\\
3 units, $t{+}2$ & units 1--3, two updates after & $3+3c$ & 0\\\bottomrule
\end{tabular}
\end{table}

Exact $99\%$ Clopper--Pearson intervals \citep{clopper1934use} for the recorded frequencies and the calibration rate are combined by a union bound into a simultaneous $95\%$ confidence set and transported through \cref{eq:closed} with conservative analytic enclosures. For one-step data the observed error frequency of unit $i$ is $\eta+(1-2\eta)(1-p_i)$, which is inverted to bound $p_i$; an unrecorded unit retains $p_i\in[1/2,1]$. Empty compatible sets count as coverage failures and receive the full scale as width. For directional contrasts, independent $97.5\%$ intervals for conditions A and B give the $95\%$ interval $[L_A-U_B,\,U_A-L_B]$ for $\Delta\vphi$; the conditions share recorded reliabilities $(0.78,0.80,0.83)$ and $\eta=0.10$ and differ only in the hidden unit ($0.85$ versus $0.55$), with 256 paired datasets per design and per-condition budget (32{,}768 or 131{,}072).

\subsection{Two-step inference: Moment, Box and Joint}\label{app:joint}
Three recorded endpoints after two steps have sufficient statistics in five groups of words, with probabilities $g_0=(1+s)/8+v/2$, $g_7=(1+s)/8-v/2$ and three pair terms, where $s=u_{12}+u_{13}+u_{23}$ and $v=(1-2\eta)\prod_ia_i$. \emph{Moment} inverts one single-unit and three pair moments to a box of reliabilities. \emph{Box} uses all five sufficient counts and projects the probability region onto a box by exact linear-fractional optimization. \emph{Joint} uses the same constraints but bounds $\vphi$ over their joint compatible region, enumerating the 15 strata of possible weakest-unit ties and bisecting surviving boxes to depth eight. On all 55{,}296 saved two-step datasets, mean widths over 216 cells were $1.11$ (Moment), $1.14$ (Box) and $0.91$ bits (Joint), with minimum cell coverage $247/256$ for Joint; most of the gain came at weak reliabilities ($59\%$ narrower there, $4$--$9\%$ elsewhere). A depth-sensitivity analysis on a fixed subset shows the remaining outer-to-inner gap shrinking between depths 6 and 12 from $0.39$--$0.43$ to $0.09$--$0.10$ bits at the equal, near-equal and heterogeneous sources, and from $0.18$ to below $0.001$ bits at the weak source.

\subsection{Calibration experiment}\label{app:calibration}
The estimator knows only the wiring and that $p_i\in[1/2,1]$. Sensor inversions with probability $0.02$ or $0.10$ are common to all recorded channels of a trial; in a block condition one inversion persists for 32 trials. Each of 96 cells has 256 datasets, with $N\in\{512,4096,32768\}$ recording and $M\in\{128,1024\}$ calibration trials. ``Ignore'' sets $\eta=0$; ``plug-in'' replaces $\eta$ by its point estimate; ``calibration-aware'' bounds $\eta$ jointly with the recording frequencies; ``known'' uses the true $\eta$ and is an oracle, relative to which calibration-aware intervals are $14$ times wider at 128 calibration trials. With block-correlated errors, block-aware bounds keep nominal coverage.

\subsection{Logistic networks and model checks}\label{app:logistic}
Eight four-unit stochastic logistic networks were drawn with off-diagonal weights uniform on $[0.45,1.45]$, self-weights uniform on $[-0.30,-0.05]$ and biases equal to minus half the incoming weight sum plus a uniform offset on $[-0.25,0.25]$; probabilities are rounded to multiples of $2^{-16}$. Weights and biases are known except for the bias $d$ of the hidden unit. Balanced interventions visit all 16 states, and $\vphi$ is evaluated at four states fixed in advance. Two continuous model families are compared: \emph{Bias}, with $d\in[-1.5,1.5]$ and $\eta\in[0.01,0.20]$, and \emph{Bias+history}, which adds a hidden self-history term $h\in[-0.25,0.25]$. A model is rejected when no parameter value is compatible with all simultaneous constraints, which has probability at most $0.05$ for a correct model. Scenarios include in-model truths, a hidden self-history shift and shifts of a recorded unit's bias by $0.05$, $0.10$ and $0.20$; 576 cells have 128 datasets each (73{,}728 datasets). At the largest budget and finest grid, mean widths were $0.21$, $0.034$ and $0.035$ bits for the three designs. Equal-trial controls reproduce the detection ordering ($90\%$, $85\%$ and $12\%$ at shift $0.20$).

\subsection{Edge prevalence, product symmetry, containment power and surrogate fits}\label{app:new}
\paragraph{Edge prevalence (\cref{fig:geometry}b; \cref{fig:edgestats}).} Two families of 200 random four-unit logistic networks: \emph{excitatory}, drawn as in \cref{app:logistic} without rounding; and \emph{mixed-sign}, with weights $\mathcal N(0,2.5^2)$ and biases equal to minus half the input sum plus $\mathcal N(0,0.25^2)$. For every state with $\vphi>0$ (631 states of 200 excitatory and 357 states of 117 mixed-sign networks) we drew two random unit directions in the 20-dimensional parameter space and evaluated $\vphi$ and $\vphiraw$ on 2{,}001 points of $[-1,1]$; every step larger than $10^{-3}$ bits and ten times the median step was bisected 45 times and kept if the jump persisted. Edge counts and specified states on each side were recorded to test \cref{prop:edges}. Five-unit networks (30 parameters, 1{,}061 cuts): 60 per family drawn the same way, one random direction per state with $\vphi>0$, 1{,}001 points on $[-1,1]$, same bisection rule (\cref{tab:edges}). The slices in \cref{fig:geometry}b vary one weight ($w_{43}$) and one bias of a mixed-sign network on a $420\times420$ grid.

\paragraph{Product symmetry and $\vphi$ (\cref{fig:access}h--j).} Parity networks of five units in which every unit reads all the others, with recorded reliabilities $(0.78,0.80,0.83)$ and hidden fidelities $(a_4,a_5)$. Recorded trajectory laws were computed exactly by forward recursion over all recorded histories. $\vphi$ was computed at the all-zero state on a $240\times240$ grid offset to avoid exact ties, along equal-product curves (601 points each), and separately at the two exact ties on the curve $a_4a_5=0.32$.

\paragraph{One-sided certificates at an edge.}\label{app:onesided} At the parity source with all four reliabilities equal to $421/512$, and at the source with one reliability lowered by $10^{-4}$, we drew $N$ complete transitions (256 datasets per cell), formed simultaneous Clopper--Pearson intervals $[l_i,h_i]$ for the four reliabilities at level $1-0.05/4$ each, and reported $L=\min\{q_l\log_2(2m_l),\,q_l\log_2(16q_l),\,-\log_2 q_h\}$ with $q_l=\prod_il_i$, $q_h=\prod_ih_i$ and $m_l=\min_il_i$. Since $\vphiraw$ increases in $q$ and $m$ through its first two terms and decreases through the third, $L\le\vphiraw\le\vphi$ whenever the intervals cover, so $\vphi>L$ is a valid certificate on both sides of the edge (\cref{tab:onesided}).

\paragraph{Containment power (\cref{fig:logic}).} For each confidence construction (Hoeffding, Chernoff--KL and Clopper--Pearson, each at familywise $\alpha=0.05$ by a union bound over cells), the acceptance range of counts in each cell was found by bisection, and the certification probability is the exact product of binomial probabilities over cells: 64 cells with $N/16$ observations each when all transition probabilities are free, four cells with $N$ observations each under the four-parameter model. The exact Hoeffding power agrees with the Monte Carlo rates of \cref{tab:cert}.

\paragraph{Learned surrogates (\cref{fig:fitting}).} For 100 random networks per family whose largest state value exceeds $0.02$ bits, we drew $N\in\{2^{10},2^{12},2^{14},2^{16},2^{18}\}$ complete transitions under balanced interventions, 64 times each, fitted four logistic regressions (one per unit) by Newton's method, and computed the plug-in $\hat\vphi$ and $\hat\vphi_{\rm raw}$ at the target state, together with the predictive divergence $\mathbb E_x\,\mathrm{KL}(K_\theta(\cdot\mid x)\,\|\,K_{\hat\theta}(\cdot\mid x))$. The trajectory in \cref{fig:fitting}b runs full-batch gradient ascent on the log-likelihood from zero initialization for a mixed-sign target selected for having an edge within distance $0.03$.

\subsection{Direct certificates, information distances and cost populations}\label{app:directexp}
\paragraph{Information distance.} For the four-parameter parity family with masks $(6,12,15,7)$ and full recording, $D^\ast(\lambda)$ is the smallest of the divergences from $\lambda$ to the seven pieces of \cref{eq:null}. Each raw-loss branch is searched over $t=\lambda(S)$ on a grid of 40--48 values followed by bounded refinement, with the convex inner problem solved by sequential quadratic programming; the cap piece is one convex problem. The same routine with the four single-unit forms and no selection cones gives the distance for $\vphiraw$.

\paragraph{Sequential runs (\cref{fig:logic}f,g; \cref{fig:geometry}e).} Each run draws transitions from a fixed source and evaluates the direct certificate on a geometric grid of sample sizes (ratio $1.08$, starting at $\log(1/\alpha)/(4D^\ast)$), stopping at the first success. Evaluating an anytime-valid certificate on a subset of times preserves its level. To save computation, a checkpoint is passed to the rigorous search only if the likelihood of the best feasible null point found by a local solver---a lower bound on the null supremum---leaves room for a certificate; in every run the rigorous search succeeded at the first checkpoint that passed this screen. Region experiment: $c=1.004261431503$; sources at the center of the proved region ($p_i=0.8333$), halfway to its boundary ($p=(0.8314,0.8351,0.8314,0.8351)$), and outside it, $\lambda=(0.70,0.77,0.736,0.738)$ and $\lambda=(0.70,0.69,0.78,0.77)$; 400 runs each (\cref{tab:direct}). Cliff experiment: $c=1/2$, sources $\lambda^0+h\,e_3$ for seven values of $h$; 300 runs each (\cref{tab:edgecost}).

\paragraph{Cost population (\cref{fig:geometry}f,g).} 6{,}000 mechanisms with reliabilities drawn independently and uniformly from $[0.76,0.86]$. The claim $\vphi>1/2$ is true for $87.4\%$ of them (smallest true value $0.628$ bits, largest false value $0.323$ bits); for the comparison, $\vphiraw>0.2714$ bits, its population median. For every true claim we computed $D^\ast$ as above.

\subsection{Trained recurrent networks}\label{app:gru}
\paragraph{Networks and task.} A GRU cell with six hidden units reads a binary sequence of length six as one-hot inputs; a linear readout of the hidden state gives two logits, and outputs are sampled from the softmax at temperature one. Networks are trained on the running parity of all prefixes (cross-entropy, Adam, learning rate $0.01$, batch size 128, 2{,}000 steps) and saved every 100 steps: 30 networks in the first study (seeds 0--9 for protocol development, 10--29 for confirmation) and ten more (seeds 100--109) for replication. All 40 reach $100\%$ accuracy on the 64 sequences.

\paragraph{Interchange interventions.} A context is a triple (base prefix, donor prefix, suffix) of three bits each; all 512 are equally likely. The network reads the base prefix and the donor prefix separately; the hidden coordinates in $S$ take the donor's values and the others the base's; the network then reads the suffix. Success is the event that the sampled output equals parity(donor prefix) $\oplus$ parity(suffix). $\mu_S$ is the exact average success probability over contexts, computed for all 64 subsets at every checkpoint (replacing all units reproduces the donor run and replacing none the base run). Candidates are the 42 subsets of at most three units.

\paragraph{Edges.} For each pair of consecutive checkpoints with accuracy at least $97\%$ at both ends, parameters were interpolated linearly on a grid of 17 points, and for each final network four random directions were scanned; a crossing of $\max_{|S|\le3}\mu_S$ through $\tau$ is \emph{isolated} if a single candidate crosses with nonzero slope while all others stay below. The audited configurations lie on the interpolation between checkpoints 100 and 200 of network 10 (crossing at coordinate $0.2938$, slope $0.0757$, runner-up $0.0135$ below $\tau$, $\Mt$ from 4 to 3), at gaps $\Delta=0.04\times2^{-j}$, $j=0,\dots,12$, located by root finding and verified in 50- and 75-digit arithmetic; and on the first isolated rising crossing of networks 102, 103 and 109, at $\Delta=10^{-3}$ and $2.5\times10^{-4}$. A single candidate exceeds $\tau$ in every configuration except the two largest gaps of network 10, where three ($\mu_S=0.940$, $0.932$, $0.902$) and two ($0.920$, $0.901$) do; $D^\ast$ is set by the largest.

\paragraph{Audits.} A query of candidate $S$ draws a context uniformly and a Bernoulli outcome with the tabulated success probability, so it is a Bernoulli$(\mu_S)$ draw; audits were simulated exactly from the tables, with batches of queries of a fixed candidate drawn from the corresponding binomial law. Costs are numbers of queries. The sampling rule sees counts only. It begins with one query of every candidate. At each later step, a candidate with fewer than $\lfloor\sqrt t/K\rfloor$ queries is brought up to that number; otherwise the empirically best candidate is queried if its mean exceeds $\tau$, and else the candidate with the largest index $\hat\mu_S+\sqrt{2\log(t+1)/N_S}$. A step issues a batch of $5\%$ of the queried candidate's current count (at least one query). All statistics of \cref{app:witness} are evaluated after every batch of the same trajectory; evaluating an anytime-valid certificate on a subset of times preserves its level, and stopping times are resolved to about $5\%$. \emph{True claims:} 1{,}000 audits at each of the 13 gaps on the network of \cref{fig:circuits}c and at each of the two gaps of the three replication networks (19{,}000 audits); every audit ended in a certificate under every statistic (\cref{tab:baselines}). \emph{False claims:} six tables in which no candidate exceeds $\tau$---the four configurations on the low side of the crossing, with the best candidate $0.0025$, $0.005$, $0.01$ and $0.02$ below $\tau$; a table with all 42 candidates exactly at $\tau$; and the table of the configuration with gap $0.00125$ with its witness lowered to $\tau$---with 2{,}000 audits of $10^{7}$ queries each (50{,}000 for the table with all candidates at $\tau$, the least favorable), evaluated at levels $0.05$, $0.01$ and $0.001$ (\cref{tab:nulls}). The plug-in readout queries every candidate $n$ times and asserts the claim if any empirical mean exceeds $\tau$; its error probability is computed exactly from the binomial laws.

\subsection{Language models}\label{app:lm}
\paragraph{Models.} GPT-2 small (12 layers of 12 attention heads, width 768, 124 million parameters) as released by OpenAI \citep{radford2019language}, and the five GPT-2-small models of the Mistral project of the Stanford Center for Research on Foundation Models \citep{mistral2021}, trained on OpenWebText for 400{,}000 steps with one training recipe and the random seeds 21, 49, 81, 343 and 777. Weights were taken from the Hugging Face hub (\texttt{openai-community/gpt2}; \texttt{stanford-crfm/alias-gpt2-small-x21}, \texttt{battlestar-gpt2-small-x49}, \texttt{caprica-gpt2-small-x81}, \texttt{darkmatter-gpt2-small-x343}, \texttt{expanse-gpt2-small-x777}) and run in PyTorch \citep{paszke2019pytorch}. The Mistral models divide the attention logits of layer $l$ by $l+1$ in addition to the usual scaling; our forward pass follows each model's configuration. On 256 task prompts its last-position logits agree with the reference implementation \citep{wolf2020transformers} to within $5\times10^{-13}$ in double precision for all six models; the tables below were computed in single precision, which differs from our double-precision result by at most $2.4\times10^{-4}$ in any logit.

\paragraph{Task and prompts.} The task is indirect-object identification \citep{wang2023interpretability}. A prompt names two people and then repeats one of them as the subject of a clause that needs an indirect object, as in `When Mary and John went to the store, John gave a drink to'; the correct continuation is the name that is not repeated. We wrote twelve templates of this form and filled them with two distinct names drawn from 149 first names, a place drawn from 16 and an object drawn from 16, all of which are single tokens of the GPT-2 vocabulary. A \emph{prompt pair} consists of a base prompt and a donor prompt that are identical except for the repeated name, so that the two prompts dictate different answers; which of the two names is repeated in the base prompt is chosen at random. A model answers a prompt correctly if the logit of the correct name exceeds the logit of the other name. Task accuracy is reported over the 4{,}096 prompts of the population below.

\paragraph{Interchange interventions.} The units are attention heads at the last token position. For a set $S$ of heads, the base prompt is run with the output of every head in $S$ at the last position (the head's value-weighted sum, before the output projection of its attention layer) replaced by the output of the same head at the last position of the donor run. The intervention \emph{succeeds} if the logit of the donor's answer then exceeds the logit of the base's answer. Because attention is causal, an intervention at the last position leaves all earlier positions unchanged; we therefore recompute the last position only, with the keys and values of earlier positions cached from the base run. This is exact, and it agrees with a full forward pass with replacement to $10^{-5}$ in the read-out for every model.

\paragraph{Candidate pools and exact tables.} For each model, on a discovery set of 1{,}024 prompt pairs, we measured for each of the 144 heads the mean shift of the logit difference toward the donor's answer when that head alone is exchanged, as a fraction of the difference between the donor and base runs. The 14 heads with the largest shift form the model's \emph{pool}. For GPT-2 small the pool contains the S-inhibition heads (7.3, 7.9, 8.6, 8.10) and name-mover heads (9.9, 10.0 and the backup name movers 9.7, 10.1, 10.2, 10.6, 10.10) of the published circuit \citep{wang2023interpretability}, together with heads 6.1, 9.3 and 9.5; its smallest sufficient set consists of three of the four S-inhibition heads. On a \emph{population} of 2{,}048 independently drawn prompt pairs (2{,}047 distinct, one of which also occurs in the discovery set) we computed the outcome of the intervention for every one of the $2^{14}=16{,}384$ subsets of the pool on every pair, 33.6 million interventions per model. $\mu_S$ is the fraction of pairs on which $S$ succeeds and $\Mt$ the size of the smallest $S$ with $\mu_S>\tau$. Within the population, all $\mu_S$ are exact.

\paragraph{Exactness over all heads.} To check that the pools do not hide smaller circuits, we also computed $\mu_S$ on the population for every single head and for every one of the 10{,}296 pairs of the 144 heads of each model. In GPT-2 small no single head exceeds $0.073$ and no pair exceeds $0.466$, the value of the best pair of the pool; its circuit size at $\tau=0.9$, three, is therefore exact over all sets of heads. In each of the five seed models no single head exceeds $0.48$ and exactly one pair exceeds $0.9$, the pair found in the pool, so their circuit size of two is exact over all sets of heads and the sufficient pair is unique. The circuit sizes of the seed models at $\tau=0.95$ (two where the pair exceeds $0.95$, three otherwise) are exact in the same sense; that of GPT-2 small at $\tau=0.95$ (four) is relative to its pool.

\paragraph{Audits.} A query names a set $S$ and returns the outcome of the intervention on a pair drawn uniformly from the population, a Bernoulli$(\mu_S)$ draw. For the claim `some set of at most $k$ heads has $\mu_S>\tau$' the candidates are the $K=\sum_{j\le k}\binom{14}{j}$ subsets of at most $k$ pool heads. The sampling rule and the statistics are those of \cref{app:gru}. We audited (i)~the claim $\Mt\le k^\ast$ at $\tau=0.9$, with $k^\ast$ the model's true circuit size (1{,}000 audits per model); (ii)~for each $k\in\{k^\ast-1,k^\ast,k^\ast+1\}$ with $2\le k\le6$, true claims with the threshold (if above $0.5$) placed $0.04$, $0.02$, $0.01$, $0.005$ and $0.0025$ below the best set of at most $k$ heads (200 audits each; \cref{tab:lmlaw}); and (iii)~false claims with the threshold placed $0.02$ and $0.005$ above the best set of at most $k^\ast$ heads (300 audits of $10^{6}$ queries each, evaluated at levels $0.05$, $0.01$ and $0.001$).

\paragraph{Plug-in readout.} The plug-in readout evaluates all candidates of a claim on the same $n$ prompt pairs, drawn with replacement from the population, and asserts the claim if the largest empirical accuracy exceeds $\tau$. For false claims with the threshold placed $0.0025$ to $0.04$ above the best set of at most $k^\ast$ heads we report the fraction of 2{,}000 repetitions in which it asserts the claim (thresholds of $0.999$ or more are omitted, which leaves four models at the largest gap).

\paragraph{Training checkpoints.} For the model with seed 21 we downloaded the checkpoints saved after 25, 50, 100, 150, 200, 250, 300 and 350 thousand training steps (repository tags \texttt{checkpoint-}$N$; the released model is the checkpoint at 400 thousand steps). At each checkpoint we computed task accuracy on the population and the interchange accuracy of all 1{,}471 sets of at most four heads of the final model's pool. Task accuracy is $80.7\%$, $85.8\%$, $92.5\%$ and $97.3\%$ at the first four checkpoints and between $98.5\%$ and $99.2\%$ from 200 thousand steps on; over that last stretch the best pair of heads has $\mu_S=0.958$, $0.937$, $0.944$, $0.940$ and $0.963$, so that $M_{0.95}$ is 2, 3, 3, 3 and 2 (\cref{fig:lmdetail}d).

\subsection{Prospective audits of instruction-tuned language models}\label{app:live}
The audits of \cref{app:lm} draw their queries from tables computed in advance, which is what allows thousands of repetitions and exact error rates. The audits of this section are run once and live: every query is an independently drawn prompt pair and a real intervention on the model, the claims and their predicted costs are written down before the first query, and the result is checked afterwards on an independent validation sample.

\begin{sloppypar}\paragraph{Models.} We audited seven instruction-tuned models of three families: \texttt{Qwen/Qwen2.5-0.5B-Instruct} (24 layers of 14 attention heads, 0.49 billion parameters), \texttt{Qwen/Qwen2.5-1.5B-Instruct} (28 layers of 12 attention heads, 1.54 billion parameters), \texttt{Qwen/Qwen2.5-3B-Instruct} (36 layers of 16 attention heads, 3.09 billion parameters) and \texttt{Qwen/Qwen2.5-7B-Instruct} (28 layers of 28 attention heads, 7.62 billion parameters) \citep{yang2024qwen25}; \texttt{HuggingFaceTB/SmolLM2-360M-Instruct} (32 layers of 15 attention heads, 0.36 billion parameters) and \texttt{HuggingFaceTB/SmolLM2-1.7B-Instruct} (24 layers of 32 attention heads, 1.71 billion parameters) \citep{allal2025smollm2}; \texttt{ibm-granite/granite-3.1-2b-instruct} (40 layers of 32 attention heads, 2.53 billion parameters) \citep{granite2024granite}. The largest model was loaded with four-bit weights (NF4 quantization of its linear layers, computed in half precision); the audited system is the model as loaded. The 16 audited combinations of model and task are the three tasks for the Qwen2.5 models of up to 3 billion parameters and for SmolLM2-1.7B, indirect-object identification and factual recall for SmolLM2-360M, and indirect-object identification for Granite and for the largest model. Weights were taken from the Hugging Face hub and run in half precision through the transformers library \citep{wolf2020transformers}, with the exceptions stated; the audited system is the model at that precision. Prompts are given to the models as plain text.\end{sloppypar}

\paragraph{Tasks.} Each task draws a base prompt and a donor prompt that dictate different one-token answers, have the same number of tokens and end in the same token; every filler is a single token of the model's own vocabulary. (i)~\emph{Indirect-object identification}, with the templates, names, places and objects of \cref{app:lm}. (ii)~\emph{Factual recall}: the capital of a country, in six templates such as `The capital of France is'; the donor names a different country with the same number of tokens (67 countries), and the answers are the first tokens of the two capitals. (iii)~\emph{Multiple choice}, after the CopyColors task of the Mechanistic Interpretability Benchmark \citep{mueller2025mib}: a sentence states the colour of an object, a question asks for it, and four options labelled A to D follow; the donor lists the same options with the correct one moved to another letter, and the answers are the two letters. A model solves a prompt if the logit of its answer exceeds the logit of the alternative.

\paragraph{Interventions.} As in \cref{app:lm}, a query replaces the outputs of a set $S$ of attention heads at the final token of the base prompt by their outputs at the final token of the donor prompt, and succeeds if the model then prefers the donor's answer. Head outputs are read and written at the input of the output projection of each attention layer. Searches over many sets on the same pairs reuse the activations of earlier tokens through the key--value cache, so that each intervention is one single-token forward pass; live queries use two ordinary forward passes per pair. The engine was checked against identities that an interchange at the final token must satisfy. In double precision, for GPT-2 small and Qwen2.5-0.5B on the three tasks, the read-out with no head replaced equals that of an ordinary forward pass on the base prompt, the read-out with every head of every layer replaced equals that of an ordinary forward pass on the donor prompt, batched and single queries agree, and the two ways of running a query agree, all to within $9.9\times10^{-14}$. For GPT-2 small the read-outs agree to within $1.2\times10^{-5}$ with the independent implementation of \cref{app:lm}, which runs in single precision. In half precision, as used in the audits, the outcome of the intervention agreed with single precision in $100.00\%$ of 6{,}000 queries for Qwen2.5-0.5B, with random sets of two to nine heads (which succeed in about $1\%$ of queries). At that precision the two ways of running a query, cached single-token passes (searches, pilot, lower bounds, validation) and ordinary forward passes (live queries), gave the same outcome in $99.98\%$ of 6{,}000 queries for the same model.

\paragraph{Protocol.} For each model and task, with independent random streams for each stage:
\begin{enumerate}[leftmargin=1.6em,itemsep=1pt,topsep=2pt,label=(\arabic*)]
\item \emph{Behaviour.} Accuracy on 2{,}000 pairs. Audits are reported for the combinations of model and task with accuracy above $0.9$, which leaves out one: the smallest model does not solve the multiple-choice task.
\item \emph{Discovery.} On 256 pairs, the effect of exchanging each head alone, for every head of the model; the 32 heads with the largest effect form a pool, and a beam search of width 6 adds one pool head at a time, keeping the sets with the highest success rate on the discovery pairs, up to 16 heads or until the best set of two consecutive sizes succeeds on at least $99.5\%$ of the pairs. Nothing is assumed about this search and nothing it finds is taken on trust.
\item \emph{Pilot.} The three best sets of each size are evaluated on 4{,}000 further pairs.
\item \emph{Registration.} For each threshold $\tau\in\{0.5,0.6,0.7,0.8,0.9,0.95,0.98\}$, let $k_\tau$ be the smallest size whose best pilot rate exceeds $\tau+0.002$. The claims `some set of at most $k$ heads has $\mu_S>\tau$' are registered for $k=k_\tau$ and $k=k_\tau+1$. Each claim lists up to six candidate sets (three when $k=1$), the sets of at most $k$ heads with the highest pilot rates, with weights $\omega_j=1/\{j(j+1)\}$ in that order, and a predicted cost: the median and the 10th and 90th percentiles of the stopping time of the certificate in 400 simulated audits whose success probabilities are drawn from the Beta posteriors of the pilot counts. The claims, candidates, weights and predictions are written to a file whose hash is stored with the results, before any query of the next stage.
\item \emph{Certification.} Claims with a predicted median within the budget of the model are audited. A query draws a prompt pair independently from the task's generator and applies the intervention for one candidate. The budgets were 200{,}000 queries per claim for models below a billion parameters, 100{,}000 for Qwen2.5-1.5B and SmolLM2-1.7B, 50{,}000 for Qwen2.5-3B and Granite and 30{,}000 for the largest model. The statistic is $\sum_j\omega_jE_j$ of \cref{thm:search}, the sampling rule is that of \cref{app:gru} with batches of at least four queries, and the audit stops when the statistic reaches $1/\alpha$, $\alpha=0.05$.
\item \emph{Lower bounds.} Where $k_\tau\ge2$, the universal claim `no single head has $\mu_S\ge\tau$' is audited over all heads of the model with the certificate of \cref{prop:universal}: each round evaluates every undecided set on the same freshly drawn pairs (two pairs while more than 64 sets are undecided, and $\min\{128,2000/\text{undecided}\}$ afterwards), a set is retired when its evidence reaches $1/\alpha$, and a claim with a set still undecided after 4{,}000 pairs is not certified. A set that never succeeds is retired after two pairs at $\tau\ge0.9$ and after eight at $\tau=0.5$, so for most sets the certificate rests on a handful of prompts; this is what a universal claim costs when almost every candidate is far from the threshold, and the sets close to it receive hundreds of queries. Where $k_\tau\ge3$ at $\tau=0.9$ and the model has at most 200{,}000 pairs of heads, the claim `no pair of heads has $\mu_S\ge\tau$' is audited in the same way over all pairs.
\item \emph{Validation.} The set with the largest weighted evidence in each certified claim is evaluated on 20{,}000 independently drawn pairs (10{,}000 for models with more than a billion parameters). This gives the success rate used to check the claim and to compute its bound $\log(1/\alpha)/\kl(\mu_S,\tau)$. For the largest model, which was run with four-bit weights, the pool had 24 heads, the beam width was 4, the pilot used 2{,}000 pairs and the validation 5{,}000.
\end{enumerate}
Error levels are per claim: a certified upper bound $\Mt\le k$ and a certified lower bound $\Mt\ge k'$ each hold with error probability at most $\alpha$, and an exact circuit size, which rests on two or three such certificates, with error probability at most their sum. Prompt pairs are drawn independently at every stage and are not excluded across stages; the generator of factual recall has $2.6\times10^{4}$ distinct pairs and the other two more than $10^{6}$.

\paragraph{The cost of enumeration.} For comparison we computed what a certificate with uniform weights over the whole class would need. The class of a claim with $k$ heads in a model with $n$ heads has $N=\sum_{i\le k}\binom ni$ sets, and such a certificate is issued when the evidence of one set reaches $N/\alpha$. Its cost is the median stopping time of that set's evidence at the validated success rate, over 200 simulated runs in which every query goes to the witness; it charges nothing for finding the witness.

\paragraph{Results.} \Cref{tab:live} lists the audits and \cref{fig:livedetail} shows the sets found by the search. Accuracy, the fraction of prompts on which the logit of the answer exceeds that of the alternative, was between $98.3\%$ and $100.0\%$ in the 16 combinations of model and task (the answer was the single most likely next token on $54\%$ to $100\%$ of prompts, and on $7\%$ for SmolLM2-1.7B on multiple choice). 224 claims were registered and 223 were audited (the predicted cost of the remaining one exceeded the budget of its model). 220 claims were certified, after 32 to 45{,}329 live queries (335{,}245 in all), and for every one of them the success rate of the certified set on the validation sample exceeded the threshold; for 215 the lower end of its exact $99\%$ interval did so as well, and for the other 5 the validation rate exceeded the threshold by at most $1.1$ percentage points. The 3 audited claims that were not certified are all at $\tau=0.98$ in the smallest model and used their full budget of 200{,}000 queries each: over 165{,}978 to 198{,}920 queries their most-queried sets succeeded at rates of $0.9804$ to $0.9810$, too close to the threshold for that budget, which is the regime in which the price of evidence grows as the inverse square of the gap. The number of queries lay within the registered interval (10th to 90th percentile) for 196 of the 220 certified claims ($89\%$), within a factor of two of the registered median for $85\%$ and within a factor of three for $95\%$; the correlation between the logarithms of predicted and realized cost was $0.92$. Relative to $\log(1/\alpha)/\kl(\mu_S,\tau)$ at the validation success rate, the audits cost a median of $3.0$ times the bound (interquartile range $1.8$ to $4.0$); the bound constrains the expected cost, and single audits fall on both sides of it. The classes of the certified claims contain up to $1.1\times10^{26}$ sets. A certificate with uniform weights over the class would have needed a median of $3.4$ times the queries paid (interquartile range $2.3$ to $5.2$; more than ten times for 16 claims). All 118 universal claims were certified, with 2{,}286{,}490 queries in all. At $\tau=0.9$, ruling out every single head of a model took 672 to 2{,}560 queries, and ruling out every pair of heads took $1.1\times10^{5}$ to $3.3\times10^{5}$ queries for 56{,}280 to 165{,}600 pairs. The certified upper and lower bounds on $M_{0.9}$ coincide in 6 of the 16 combinations. Error levels are per claim: every certificate has level $0.05$, and a circuit size certified exactly rests on two or three certificates. The searches that preceded the audits used $6.5\times10^{6}$ interventions on the discovery samples and $1.5\times10^{6}$ on the pilot samples, on cached activations. One further combination, SmolLM2 360M on multiple choice, is left out of all counts because the model does not solve the task (accuracy $49.4\%$); 2 claims were registered for it, 1 audited and 1 certified (after 74{,}856 queries).

\subsection{Audit of a safety mechanism: refusal}\label{app:refusal}
\paragraph{Claim.} Instruction-tuned models refuse harmful requests. Arditi et al.\ \citep{arditi2024refusal} reported that in many models this behaviour is mediated by one direction of the residual stream. We audit two forms of that statement as existential claims. \emph{Ablation}: some direction $r$ in a candidate set is such that, when the component along $r$ is removed from the residual stream at every layer and position, the model answers a harmful request without refusing with probability above $\tau$. \emph{Addition}: some candidate is such that adding it at its layer makes the model refuse a harmless request with probability above $\tau$.

\paragraph{Data and candidates.} Instructions are the harmful and harmless splits released with ref.~\citenum{arditi2024refusal}, which collect public benchmarks. The audit populations are the 572 harmful test instructions and 1{,}000 harmless test instructions drawn once. Prompts use each model's chat template with its default system prompt. A candidate is the difference between the mean residual-stream activations on 256 harmful and 256 harmless training instructions at one layer (every layer but the last) and one of the last five token positions of the prompt, which follow the instruction. A response is scored as a refusal if its first 12 generated tokens (greedy decoding) contain one of 23 refusal phrases such as `I cannot' or `I'm sorry'; only this binary outcome is recorded.

\paragraph{Protocol.} The protocol is that of \cref{app:live}: every candidate is tried on 24 discovery instructions; the twelve best are evaluated on 100 instructions drawn from the audit population; the six best of these are registered with weights $1/\{j(j+1)\}$ and predicted costs for each $\tau\in\{0.5,0.75,0.9\}$ that the best pilot rate exceeds by more than $0.005$ (where the best pilot rate falls short of the lowest such threshold by more than $0.05$, the universal claim that no candidate succeeds with probability $\tau$ or more is audited over all candidates instead); each live query draws an instruction from the audit population with replacement and applies the intervention for one candidate, in batches of at least eight and within a budget of 20{,}000 queries; and the leading candidate of each certified claim is then evaluated on the whole audit population, which gives its success rate in that population exactly.

\paragraph{Results.} \Cref{tab:refusal} lists the audits. Qwen2.5 0.5B refused $64.7\%$ of the harmful and $7.5\%$ of the harmless requests; the ablation claim at $\tau=0.9$ was certified after 200 queries, and with the certified direction (layer 13) removed the model no longer refused $98.6\%$ of the harmful requests; the addition claim at $\tau=0.9$ was certified after 886 queries (layer 13; $94.5\%$ of harmless requests refused). Qwen2.5 1.5B refused $90.6\%$ of the harmful and $6.9\%$ of the harmless requests; the ablation claim at $\tau=0.9$ was certified after 120 queries, and with the certified direction (layer 15) removed the model no longer refused $99.8\%$ of the harmful requests; the addition claim at $\tau=0.9$ was certified after 120 queries (layer 16; $97.9\%$ of harmless requests refused). Granite 3.1 2B refused $80.8\%$ of the harmful and $0.9\%$ of the harmless requests; the ablation claim at $\tau=0.9$ was certified after 192 queries, and with the certified direction (layer 14) removed the model no longer refused $99.5\%$ of the harmful requests; the addition claim at $\tau=0.9$ was certified after 328 queries (layer 14; $98.0\%$ of harmless requests refused). Qwen2.5 3B refused $84.3\%$ of the harmful and $1.8\%$ of the harmless requests; the ablation claim at $\tau=0.9$ was certified after 120 queries, and with the certified direction (layer 19) removed the model no longer refused $100.0\%$ of the harmful requests; the addition claim at $\tau=0.9$ was certified after 120 queries (layer 21; $99.5\%$ of harmless requests refused). Qwen2.5 7B refused $71.3\%$ of the harmful and $0.8\%$ of the harmless requests; the ablation claim at $\tau=0.9$ was certified after 200 queries, and with the certified direction (layer 14) removed the model no longer refused $99.8\%$ of the harmful requests; the addition claim at $\tau=0.9$ was certified after 120 queries (layer 16; $99.8\%$ of harmless requests refused). The ablation and addition claims are certified separately, each for its own leading candidate; in every model the two leading candidates lie within 2 layers of each other. All 30 claims registered for these 5 models were certified (30 of 30), and each is true of its audit population, as the population rates show. Each audit was preceded by 2{,}760 to 4{,}680 generations for discovery and 1{,}200 for the pilot. The two SmolLM2 models refuse only $6\%$ and $48\%$ of the harmful requests without any intervention, so ablation says little about them; they are listed in the table for completeness, and for both the universal claim that no candidate direction makes the model refuse half of the harmless requests was certified over all candidates (1{,}404 and 1{,}495 queries).

\begin{table}[h!]\centering\small
\caption{\textbf{Audit of refusal.} Fraction of harmful and of harmless requests that each model refuses; layer of the certified direction; with that direction removed, fraction of the 572 harmful requests of the audit population that are no longer refused (at the largest certified threshold); number of live queries to the certificate of the claim that some direction succeeds with probability above $\tau$ (n.c., not certified within the budget; --, not registered); and the same for the addition claim, which is certified separately for its own candidate direction, on 1{,}000 harmless requests, where success is a refusal. A response counts as a refusal when its first 12 generated tokens contain a refusal phrase.}\label{tab:refusal}
\setlength{\tabcolsep}{4pt}
\resizebox{\linewidth}{!}{%
\begin{tabular}{@{}lrrcrrrrrrrr@{}}
\toprule
 & \multicolumn{2}{c}{refusals (\%)} & & \multicolumn{4}{c}{ablation, harmful requests} & \multicolumn{4}{c}{addition, harmless requests} \\
\cmidrule(lr){2-3}\cmidrule(lr){5-8}\cmidrule(lr){9-12}
model & harmful & harmless & layer & not refused (\%) & $\tau=0.5$ & $0.75$ & $0.9$ & refused (\%) & $\tau=0.5$ & $0.75$ & $0.9$ \\
\midrule
SmolLM2 360M & 6.1 & 3.3 & 1/32 & 100.0 & 56 & 72 & 120 & -- & -- & -- & -- \\
Qwen2.5 0.5B & 64.7 & 7.5 & 13/24 & 98.6 & 56 & 72 & 200 & 94.5 & 56 & 96 & 886 \\
Qwen2.5 1.5B & 90.6 & 6.9 & 15/28 & 99.8 & 56 & 72 & 120 & 97.9 & 56 & 72 & 120 \\
SmolLM2 1.7B & 47.9 & 2.1 & 1/24 & 100.0 & 56 & 72 & 120 & -- & -- & -- & -- \\
Granite 3.1 2B & 80.8 & 0.9 & 14/40 & 99.5 & 56 & 72 & 192 & 98.0 & 56 & 96 & 328 \\
Qwen2.5 3B & 84.3 & 1.8 & 19/36 & 100.0 & 56 & 72 & 120 & 99.5 & 56 & 72 & 120 \\
Qwen2.5 7B & 71.3 & 0.8 & 14/28 & 99.8 & 56 & 72 & 200 & 99.8 & 56 & 72 & 120 \\
\bottomrule
\end{tabular}}
\end{table}

\subsection{Ten training runs of a language model}\label{app:runs}
\paragraph{Models and checkpoints.} Pythia-160m \citep{biderman2023pythia} is a GPT-NeoX transformer with 12 layers, 12 attention heads per layer, a width of 768 and 162 million parameters, trained on the Pile for 143{,}000 steps of 1{,}024 sequences of 2{,}048 tokens. The original run and nine further runs with other random seeds \citep{vanderwal2025polypythias} are public (\texttt{EleutherAI/pythia-160m} and \texttt{EleutherAI/pythia-160m-seed1} to \texttt{-seed9}), each with 154 checkpoints. We used 56 checkpoints: step 143{,}000 of every run; steps 1{,}000, 2{,}000, 3{,}000, 4{,}000, 5{,}000, 6{,}000, 8{,}000, 10{,}000, 13{,}000, 16{,}000, 20{,}000, 25{,}000, 30{,}000, 36{,}000, 43{,}000, 50{,}000, 57{,}000, 64{,}000, 71{,}000, 79{,}000, 86{,}000, 93{,}000, 100{,}000, 107{,}000, 114{,}000, 121{,}000, 128{,}000, 135{,}000 of the original run; and steps 57{,}000 and 100{,}000 of each further run. The weights were used in half precision, as released.

\paragraph{Tasks.} Every task has a population of 1{,}000 prompt pairs, drawn once and used for every checkpoint and run. \emph{Indirect-object identification}: the templates and vocabulary of \cref{app:lm}; base and donor differ only in the repeated name. \emph{Subject--verb agreement}: six templates such as `The \textit{teacher} near the \textit{doctors}', with a singular or plural subject and an intervening noun of either number; the answers are ` is' and ` are'. \emph{Pronoun gender}: six templates such as `I talked to \textit{Mary} this morning because', with one-token female and male first names; the answers are ` she' and ` he'. \emph{Induction}: a sequence of random word tokens in which the pair $A\,B$ occurs twice and which ends in $A$; the donor has $C$ in place of $B$. Accuracy is the fraction of the 2{,}000 prompts on which the logit of the prompt's own answer exceeds that of its partner's answer.

\paragraph{Tables.} For each of the 224 combinations of checkpoint and task we computed the read-out after exchanging each of the 144 heads alone on each of the 1{,}000 pairs; the mean of the normalized effect $(r_{\mathrm{patched}}-r_{\mathrm{base}})/(r_{\mathrm{donor}}-r_{\mathrm{base}})$, clipped to $[0,1]$, ranked the heads; a beam search of width 4 over the 24 highest-ranked heads grew sets up to size 12 on the first 256 pairs, ranking sets by success rate and then by mean read-out, and stopped early once two consecutive sizes reached $99.5\%$; the three best sets of every size were then evaluated on all 1{,}000 pairs. In 213 combinations the accuracy was at least $90\%$; claims were formed for these only. A checkpoint and task take about $4\times10^{5}$ single-token forward passes and 6 to 26 seconds on a laptop graphics processor (search with the option \texttt{--kmax 12}). For four runs the success rate of all 10{,}296 pairs of heads on indirect-object identification was computed as well.

\paragraph{Claims and audits.} For each such combination and each of six thresholds, the true claim is `some set of at most $k$ heads has $\mu_S>\tau$' with $k$ the smallest size at which an evaluated set clears $\tau$; its candidates are the up to six evaluated sets of at most $k$ heads with the largest success rates on the 256 search pairs, in that order, with weights $1/\{j(j+1)\}$, as in the live audits. No sufficient set of up to 12 heads was found in 79 of the 1{,}278 combinations of checkpoint, task and threshold. This gave 1{,}199 true claims. Each was audited 100 times with the certificate of \cref{thm:search}, a query being a pair drawn uniformly from the population; all but seven were certified in every audit within $10^{7}$ queries (the remaining claims in at least $96\%$ of audits; a candidate just below the threshold can hold the sampling rule for long). The 855 false claims replace $k$ by $k-1$; none of their candidates clears $\tau$, and where $k-1=1$ no head of the model does. Each was audited 200 times for 20{,}000 queries. The plug-in readout draws $n$ pairs with replacement, evaluates every candidate on them and asserts the claim if any empirical rate exceeds $\tau$ (1{,}000 repetitions). The estimate of the search is the largest success rate of the candidates on the 256 search pairs. Between successive analyzed checkpoints of a run (1{,}000 to 8{,}000 steps apart in the original run, 43{,}000 in the others) whose accuracies were both at least $90\%$ and differed by less than two points (161 cases), the size of the smallest sufficient set found changed in 66 at $\tau=0.9$ and in 106 at one of the thresholds $0.8$, $0.9$ and $0.95$; the smallest sufficient set at $\tau=0.9$ changed in 83, and the head with the largest mean effect in 21. \Cref{tab:runs} lists the ten runs at the end of training.

\begin{table}[h]\centering\small
\caption{Ten training runs of Pythia-160m at the end of training (step 143{,}000). For each task: accuracy (\%), size $M$ of the smallest set of heads found with success rate above $0.9$, and the largest success rate of a single head (exact over all 144 heads). Every rate is exact for the task's population of 1{,}000 prompt pairs.}\label{tab:runs}
\setlength{\tabcolsep}{4pt}
\begin{tabular}{@{}lcccccccccccc@{}}\toprule
 & \multicolumn{3}{c}{indirect object} & \multicolumn{3}{c}{agreement} & \multicolumn{3}{c}{pronoun gender} & \multicolumn{3}{c}{induction}\\\cmidrule(lr){2-4}\cmidrule(lr){5-7}\cmidrule(lr){8-10}\cmidrule(l){11-13}
Run & acc. & $M$ & single & acc. & $M$ & single & acc. & $M$ & single & acc. & $M$ & single\\\midrule
original & 97.0 & 5 & 0.289 & 94.5 & 4 & 0.734 & 97.6 & 4 & 0.230 & 99.2 & $>$12 & 0.038\\
seed 1 & 98.1 & 2 & 0.865 & 98.0 & 3 & 0.603 & 100.0 & 4 & 0.142 & 99.4 & $>$12 & 0.171\\
seed 2 & 99.1 & 1 & 0.994 & 99.2 & 2 & 0.837 & 100.0 & 3 & 0.101 & 99.7 & $>$12 & 0.012\\
seed 3 & 90.6 & 3 & 0.754 & 97.9 & 3 & 0.318 & 100.0 & 3 & 0.117 & 99.7 & 12 & 0.047\\
seed 4 & 98.6 & 3 & 0.277 & 97.7 & 4 & 0.204 & 98.4 & 4 & 0.168 & 99.4 & $>$12 & 0.051\\
seed 5 & 98.4 & 3 & 0.236 & 98.9 & 2 & 0.451 & 99.7 & 2 & 0.536 & 99.8 & 12 & 0.012\\
seed 6 & 98.9 & 2 & 0.614 & 96.4 & 2 & 0.401 & 99.1 & 4 & 0.192 & 99.8 & $>$12 & 0.011\\
seed 7 & 98.5 & 2 & 0.673 & 96.9 & 4 & 0.152 & 99.6 & 3 & 0.337 & 99.8 & 12 & 0.097\\
seed 8 & 97.9 & 1 & 0.978 & 98.8 & 3 & 0.421 & 100.0 & 3 & 0.393 & 99.6 & 12 & 0.020\\
seed 9 & 96.0 & 2 & 0.889 & 99.2 & 2 & 0.751 & 100.0 & 3 & 0.116 & 98.9 & $>$12 & 0.039\\
\bottomrule
\end{tabular}
\end{table}

\section{Additional results}\label{app:results}

\begin{table}[h]\centering\small
\caption{Directional contrasts between two conditions that differ only in the hidden unit (recorded reliabilities $(0.78,0.80,0.83)$; $\Delta\vphi=0.2433$ bits; 256 paired datasets per row; budgets are per condition). Every interval covered the true difference and none supported the wrong sign.}\label{tab:contrast}
\begin{tabular}{@{}llccc@{}}\toprule
Budget & Design & Sign resolved & Unresolved & Mean width of $\Delta\vphi$ interval (bits)\\\midrule
32{,}768 & 3 units, $t{+}1$ & 0 & 256 & 2.41\\
32{,}768 & 4 units, $t{+}1$ & 256 & 0 & 0.98\\
32{,}768 & 3 units, $t{+}2$ & 0 & 256 & 1.62\\
131{,}072 & 3 units, $t{+}1$ & 0 & 256 & 2.21\\
131{,}072 & 4 units, $t{+}1$ & 256 & 0 & 0.43\\
131{,}072 & 3 units, $t{+}2$ & 256 & 0 & 0.97\\\bottomrule
\end{tabular}
\end{table}

\begin{table}[h]\centering\small
\caption{Two-step inference on identical data (55{,}296 datasets, 216 equally weighted cells). Left: overall. Right: Joint by source, and its change relative to Box.}\label{tab:joint}
\setlength{\tabcolsep}{4pt}
\setlength{\tabcolsep}{4pt}
\setlength{\tabcolsep}{4pt}
\setlength{\tabcolsep}{4pt}
\setlength{\tabcolsep}{4pt}
\begin{tabular}{@{}lcc@{}}\toprule
Procedure & Mean width (bits) & Min.\ cell coverage\\\midrule
Moment & 1.112 & 248/256\\
Box & 1.138 & 256/256\\
Joint & 0.914 & 247/256\\
4 units, $t{+}1$ (own data) & 0.693 & 255/256\\\bottomrule
\end{tabular}\hfill
\begin{tabular}{@{}lcc@{}}\toprule
Source & Joint (bits) & Joint $-$ Box\\\midrule
equal & 1.063 & $-0.044$\\
near-equal & 1.066 & $-0.041$\\
heterogeneous & 1.029 & $-0.103$\\
weak & 0.498 & $-0.710$\\\bottomrule
\end{tabular}
\end{table}

\begin{table}[h]\centering\small
\caption{Eight random logistic networks, hidden bias $0.25$, budget 131{,}072 (Bias family). Widths and coverage at the finest grid; rejection rates for unmodeled shifts of a recorded unit's bias at the coarse grid (equal-cost designs).}\label{tab:logistic}
\begin{tabular}{@{}lccccc@{}}\toprule
Design & Mean width (bits) & Coverage & Reject, shift 0.05 & Reject, 0.10 & Reject, 0.20\\\midrule
3 units, $t{+}1$ & 0.214 & 0.976 & 0.067 & 0.228 & 0.952\\
4 units, $t{+}1$ & 0.034 & 0.986 & 0.033 & 0.153 & 0.851\\
3 units, $t{+}2$ & 0.035 & 0.993 & 0.001 & 0.011 & 0.120\\\bottomrule
\end{tabular}
\end{table}

\begin{table}[h]\centering\small
\caption{Certification of $\vphi>1.004$ bits by containment with 64 free probabilities. Left: Monte Carlo results with the Hoeffding rule (256 datasets per cell); every one of the 3{,}072 confidence sets contained its true law. Right: transitions needed for $95\%$ certification probability under three exact confidence constructions.}\label{tab:cert}
\begin{tabular}{@{}lccc@{}}\toprule
Source & $N=2^{23}$ & $2^{25}$ & $2^{27}$\\\midrule
center & 2/256 & 256/256 & 256/256\\
halfway & 0/256 & 30/256 & 256/256\\
boundary & 0/256 & 0/256 & 0/256\\
one cell outside & 0/256 & 0/256 & 0/256\\\bottomrule
\end{tabular}\hfill
\begin{tabular}{@{}lcc@{}}\toprule
Confidence set & center & halfway\\\midrule
Hoeffding & 12.5\,M & 47.6\,M\\
Chernoff--KL & 8.9\,M & 33.6\,M\\
Clopper--Pearson & 7.5\,M & 28.2\,M\\\bottomrule
\end{tabular}
\end{table}

\begin{table}[h]\centering\small
\caption{Direct certification of $\vphi>1.004261$ bits at level $\alpha=0.05$ in the four-parameter parity family (400 sequential runs per source), against certification by containment in the proved region under the same model (exact power). $D^\ast$: information distance to the closed null (nats per transition); bound: $\log(1/\alpha)/D^\ast$.}\label{tab:direct}
\setlength{\tabcolsep}{4pt}
\begin{tabular}{@{}lcccrrcrr@{}}\toprule
 & & & & \multicolumn{3}{c}{Direct certificate} & \multicolumn{2}{c}{Containment}\\\cmidrule(lr){5-7}\cmidrule(l){8-9}
Source & $\vphi$ & $D^\ast$ & bound & median & $95\%$ & certified & $50\%$ & $95\%$\\\midrule
center of the region & 1.053 & $7.27\times10^{-4}$ & 4{,}123 & 38{,}615 & 52{,}538 & 400/400 & 158{,}316 & 258{,}131\\
halfway to its boundary & 1.053 & $7.27\times10^{-4}$ & 4{,}121 & 38{,}615 & 56{,}742 & 400/400 & 507{,}134 & 928{,}512\\
outside (two reliabilities) & 1.046 & $4.12\times10^{-4}$ & 7{,}276 & 62{,}895 & 92{,}786 & 400/400 & never & never\\
outside (all four) & 1.060 & $5.10\times10^{-4}$ & 5{,}871 & 50{,}758 & 80{,}550 & 400/400 & never & never\\
\bottomrule
\end{tabular}
\end{table}

\begin{table}[h]\centering\small
\caption{The price of evidence at a cliff: direct certification of $\vphi>1/2$ at level $\alpha=0.05$ on the path $\lambda^0+h\,e_3$ (300 sequential runs per distance). Every run ended in a certificate. The margin $\vphi-1/2$ (bits) is nearly constant; the information distance $D^\ast$ scales as $h^{2}$ and the cost as $h^{-2}$, at 8 to 14 times the bound. Bound: $\log(1/\alpha)/D^\ast$.}\label{tab:edgecost}
\setlength{\tabcolsep}{3.2pt}
\begin{tabular}{@{}lcccccccc@{}}\toprule
 & & & & & \multicolumn{3}{c}{Transitions to certification} & \\\cmidrule(lr){6-8}
$h$ & margin & $D^\ast$ & $D^\ast/h^2$ & bound & median & 25th pct. & 75th pct. & median/bound\\\midrule
0.04 & 0.432 & $1.00\times10^{-3}$ & 0.6273 & $2.98\times10^{3}$ & $2.40\times10^{4}$ & $1.76\times10^{4}$ & $3.02\times10^{4}$ & 8.0\\
0.02 & 0.410 & $2.44\times10^{-4}$ & 0.6102 & $1.23\times10^{4}$ & $1.14\times10^{5}$ & $9.09\times10^{4}$ & $1.34\times10^{5}$ & 9.3\\
0.01 & 0.400 & $6.02\times10^{-5}$ & 0.6019 & $4.98\times10^{4}$ & $5.00\times10^{5}$ & $4.29\times10^{5}$ & $5.84\times10^{5}$ & 10.1\\
0.005 & 0.394 & $1.49\times10^{-5}$ & 0.5979 & $2.00\times10^{5}$ & $2.35\times10^{6}$ & $1.87\times10^{6}$ & $2.74\times10^{6}$ & 11.7\\
0.0025 & 0.391 & $3.72\times10^{-6}$ & 0.5959 & $8.04\times10^{5}$ & $9.43\times10^{6}$ & $8.09\times10^{6}$ & $1.10\times10^{7}$ & 11.7\\
0.00125 & 0.390 & $9.30\times10^{-7}$ & 0.5949 & $3.22\times10^{6}$ & $4.08\times10^{7}$ & $3.50\times10^{7}$ & $5.14\times10^{7}$ & 12.7\\
0.000625 & 0.390 & $2.32\times10^{-7}$ & 0.5944 & $1.29\times10^{7}$ & $1.76\times10^{8}$ & $1.60\times10^{8}$ & $2.06\times10^{8}$ & 13.7\\
\bottomrule
\end{tabular}
\end{table}

\begin{table}[h]\centering\small
\caption{One-sided certificates at the edge of the parity network (256 datasets per row). The certified lower bound is valid in every dataset on both sides of the edge, while the plug-in estimate with four free reliabilities fails at the tie.}\label{tab:onesided}
\begin{tabular}{@{}lcccccc@{}}\toprule
Source & true $\vphi$ & $N$ & mean bound & smallest bound & bound valid & plug-in MAE\\\midrule
four equal reliabilities & 1.129 & $2^{12}$ & 0.288 & 0.265 & 256/256 & 0.798\\
 & & $2^{16}$ & 0.318 & 0.312 & 256/256 & 0.801\\
 & & $2^{20}$ & 0.326 & 0.324 & 256/256 & 0.802\\
one lowered by $10^{-4}$ & 0.328 & $2^{12}$ & 0.287 & 0.266 & 256/256 & 0.015\\
 & & $2^{16}$ & 0.318 & 0.313 & 256/256 & 0.004\\
 & & $2^{20}$ & 0.325 & 0.324 & 256/256 & 0.002\\\bottomrule
\end{tabular}
\end{table}

\begin{table}[h]\centering\small
\caption{Learned surrogate models: median predictive divergence of the fitted model (nearly identical in the two families) and probability that the plug-in value errs by more than $20\%$ (100 random networks per family, 64 fits per network and sample size).}\label{tab:sysid}
\begin{tabular}{@{}rccccc@{}}\toprule
 & & \multicolumn{2}{c}{Excitatory} & \multicolumn{2}{c}{Mixed-sign}\\\cmidrule(lr){3-4}\cmidrule(l){5-6}
$N$ & median KL (nats) & $\vphi$ & $\vphiraw$ & $\vphi$ & $\vphiraw$\\\midrule
$2^{10}$ & $9.5\times10^{-3}$ & 0.164 & 0.098 & 0.541 & 0.489\\
$2^{12}$ & $2.4\times10^{-3}$ & 0.045 & 0.003 & 0.303 & 0.257\\
$2^{14}$ & $5.9\times10^{-4}$ & 0.020 & 0 & 0.114 & 0.088\\
$2^{16}$ & $1.5\times10^{-4}$ & 0.013 & 0 & 0.032 & 0.019\\
$2^{18}$ & $3.7\times10^{-5}$ & 0.008 & 0 & 0.010 & 0.002\\\bottomrule
\end{tabular}
\end{table}

\begin{table}[h]\centering\small
\caption{Edges in random four- and five-unit logistic networks. Each scan follows one random direction through a state with $\vphi>0$ over a parameter distance of one on either side (two directions per state for four units, one for five); every discontinuity is confirmed by bisection. Networks: those with at least one state with $\vphi>0$, of the number drawn. Jumps are relative to the larger side.}\label{tab:edges}
\setlength{\tabcolsep}{4pt}
\begin{tabular}{@{}llcrrccc@{}}\toprule
Units & Family & Networks & Scans & Edges & Edge within distance 1 & Median jump & Jumps $\ge25\%$\\\midrule
4 & excitatory & 200/200 & 1{,}262 & 424 & 26\% & 25\% & 100\%\\
4 & mixed-sign & 117/200 & 714 & 425 & 44\% & 50\% & 99\%\\
5 & excitatory & 60/60 & 143 & 73 & 33\% & 33\% & 99\%\\
5 & mixed-sign & 37/60 & 116 & 76 & 48\% & 45\% & 97\%\\\bottomrule
\end{tabular}
\end{table}

\begin{table}[h]\centering\small
\caption{False claims in trained recurrent networks. Number of audits ($10^{7}$ interventions each) in which each certificate asserted the false claim $\Mt\le3$ at levels $\alpha=0.05/0.01/0.001$, and the exact probability (\%) that a plug-in readout with $n$ queries per candidate asserts it. A valid certificate asserts a false claim in at most a fraction $\alpha$ of audits; the table with all candidates exactly at $\tau$ is the least favorable.}\label{tab:nulls}
\setlength{\tabcolsep}{3.5pt}
\resizebox{\linewidth}{!}{%
\begin{tabular}{@{}lrcccccccc@{}}\toprule
 & & \multicolumn{5}{c}{Certificates (false assertions at $0.05/0.01/0.001$)} & \multicolumn{3}{c}{Plug-in (\%)}\\\cmidrule(lr){3-7}\cmidrule(l){8-10}
Configuration & audits & product & betting product & union, Hoeffding & union, Bernoulli & witness & $n=32$ & $100$ & $1{,}000$\\\midrule
best candidate $0.0025$ below $\tau$ & 2{,}000 & 0/0/0 & 0/0/0 & 0/0/0 & 1/0/0 & 1/0/0 & 93 & 62 & 42\\
best candidate $0.005$ below $\tau$ & 2{,}000 & 0/0/0 & 0/0/0 & 0/0/0 & 0/0/0 & 0/0/0 & 93 & 59 & 32\\
best candidate $0.01$ below $\tau$ & 2{,}000 & 0/0/0 & 0/0/0 & 0/0/0 & 0/0/0 & 0/0/0 & 91 & 52 & 16\\
best candidate $0.02$ below $\tau$ & 2{,}000 & 0/0/0 & 0/0/0 & 0/0/0 & 0/0/0 & 0/0/0 & 87 & 39 & 2\\
all 42 candidates at $\tau$ & 50{,}000 & 0/0/0 & 2318/468/41 & 0/0/0 & 154/37/5 & 706/149/15 & 100 & 100 & 100\\
one candidate at $\tau$, others below & 2{,}000 & 0/0/0 & 0/0/0 & 0/0/0 & 0/0/0 & 0/0/0 & 95 & 66 & 53\\
\bottomrule
\end{tabular}}
\end{table}

\begin{table}[p]\centering\scriptsize
\caption{Language models: claims `some set of at most $k$ heads has $\mu_S>\tau$' with the threshold placed at a fixed gap (percentage points) below the best set of at most $k$ heads. $K$: number of candidates. Bound: $\log(1/\alpha)/D^\ast$. Witness: median number of interventions at $\alpha=0.05$ (200 audits per row; every audit ended in a certificate under every statistic). Last three columns: median cost of the product certificate and of the two union bounds relative to the witness certificate.}\label{tab:lmlaw}
\setlength{\tabcolsep}{4pt}
\begin{tabular}{@{}lccccrrcccc@{}}\toprule
Model & $k$ & $K$ & $\tau$ & gap & bound & witness & witness/bound & product & union, Hoeffding & union, Bernoulli\\\midrule
GPT-2 small & 3 & 470 & 0.8653 & 4 & 397 & 4{,}458 & 11.2 & 22.9 & 1.5 & 1.06\\
GPT-2 small & 3 & 470 & 0.8853 & 2 & 1{,}441 & 11{,}080 & 7.7 & 38.9 & 1.8 & 1.03\\
GPT-2 small & 3 & 470 & 0.8953 & 1 & 5{,}456 & 31{,}983 & 5.9 & 57.7 & 2.3 & 1.12\\
GPT-2 small & 3 & 470 & 0.9003 & 0.5 & 21{,}194 & 98{,}648 & 4.7 & 76.9 & 2.8 & 1.16\\
GPT-2 small & 3 & 470 & 0.9028 & 0.25 & 83{,}496 & 399{,}209 & 4.8 & 78.1 & 2.9 & 1.16\\
GPT-2 small & 4 & 1{,}471 & 0.9419 & 4 & 152 & 2{,}943 & 19.4 & 18.5 & 2.1 & 1.02\\
GPT-2 small & 4 & 1{,}471 & 0.9619 & 2 & 444 & 4{,}837 & 10.9 & 33.6 & 4.1 & 1.14\\
GPT-2 small & 4 & 1{,}471 & 0.9719 & 1 & 1{,}432 & 22{,}390 & 15.6 & 51.1 & 3.6 & 1.08\\
GPT-2 small & 4 & 1{,}471 & 0.9769 & 0.5 & 5{,}003 & 60{,}702 & 12.1 & 81.0 & 4.8 & 1.08\\
GPT-2 small & 4 & 1{,}471 & 0.9794 & 0.25 & 18{,}527 & 133{,}990 & 7.2 & 143.1 & 8.5 & 1.16\\
\addlinespace[2pt]
seed 21 & 2 & 106 & 0.9229 & 4 & 218 & 882 & 4.0 & 8.2 & 3.7 & 1.28\\
seed 21 & 2 & 106 & 0.9429 & 2 & 711 & 3{,}212 & 4.5 & 14.3 & 4.0 & 1.15\\
seed 21 & 2 & 106 & 0.9529 & 1 & 2{,}501 & 9{,}154 & 3.7 & 19.4 & 5.7 & 1.27\\
seed 21 & 2 & 106 & 0.9579 & 0.5 & 9{,}293 & 33{,}014 & 3.6 & 23.2 & 6.6 & 1.31\\
seed 21 & 2 & 106 & 0.9604 & 0.25 & 35{,}723 & 127{,}719 & 3.6 & 25.9 & 7.0 & 1.33\\
seed 21 & 3 & 470 & 0.9571 & 4 & 91 & 1{,}418 & 15.6 & 8.1 & 3.1 & 0.99\\
seed 21 & 3 & 470 & 0.9771 & 2 & 211 & 2{,}108 & 10.0 & 12.4 & 6.9 & 1.01\\
seed 21 & 3 & 470 & 0.9871 & 1 & 525 & 3{,}552 & 6.8 & 18.2 & 15.7 & 1.10\\
seed 21 & 3 & 470 & 0.9921 & 0.5 & 1{,}430 & 8{,}072 & 5.6 & 22.0 & 27.6 & 1.31\\
seed 21 & 3 & 470 & 0.9946 & 0.25 & 4{,}307 & 24{,}404 & 5.7 & 59.4 & 36.9 & 1.29\\
\addlinespace[2pt]
seed 49 & 2 & 106 & 0.8667 & 4 & 393 & 1{,}924 & 4.9 & 11.9 & 1.8 & 1.11\\
seed 49 & 2 & 106 & 0.8867 & 2 & 1{,}424 & 6{,}388 & 4.5 & 16.4 & 2.2 & 1.11\\
seed 49 & 2 & 106 & 0.8967 & 1 & 5{,}386 & 19{,}770 & 3.7 & 21.6 & 2.8 & 1.20\\
seed 49 & 2 & 106 & 0.9017 & 0.5 & 20{,}911 & 79{,}449 & 3.8 & 22.1 & 2.8 & 1.17\\
seed 49 & 2 & 106 & 0.9042 & 0.25 & 82{,}360 & 319{,}518 & 3.9 & 24.9 & 3.1 & 1.22\\
seed 49 & 3 & 470 & 0.9522 & 4 & 112 & 1{,}306 & 11.6 & 10.8 & 3.3 & 0.97\\
seed 49 & 3 & 470 & 0.9722 & 2 & 291 & 2{,}084 & 7.2 & 17.3 & 7.1 & 1.08\\
seed 49 & 3 & 470 & 0.9822 & 1 & 829 & 4{,}160 & 5.0 & 26.3 & 14.4 & 1.29\\
seed 49 & 3 & 470 & 0.9872 & 0.5 & 2{,}610 & 16{,}400 & 6.3 & 30.1 & 15.2 & 1.29\\
seed 49 & 3 & 470 & 0.9897 & 0.25 & 8{,}965 & 55{,}938 & 6.2 & 59.4 & 18.0 & 1.32\\
\addlinespace[2pt]
seed 81 & 2 & 106 & 0.9351 & 4 & 176 & 758 & 4.3 & 7.1 & 4.3 & 1.26\\
seed 81 & 2 & 106 & 0.9551 & 2 & 542 & 2{,}032 & 3.8 & 14.6 & 6.2 & 1.31\\
seed 81 & 2 & 106 & 0.9651 & 1 & 1{,}821 & 6{,}932 & 3.8 & 19.2 & 7.3 & 1.26\\
seed 81 & 2 & 106 & 0.9701 & 0.5 & 6{,}564 & 24{,}178 & 3.7 & 22.5 & 8.5 & 1.22\\
seed 81 & 2 & 106 & 0.9726 & 0.25 & 24{,}781 & 100{,}787 & 4.1 & 22.2 & 8.8 & 1.27\\
seed 81 & 3 & 470 & 0.9590 & 4 & 81 & 908 & 11.3 & 11.7 & 5.1 & 0.98\\
seed 81 & 3 & 470 & 0.9790 & 2 & 174 & 2{,}005 & 11.5 & 11.0 & 7.3 & 0.70\\
seed 81 & 3 & 470 & 0.9890 & 1 & 390 & 3{,}370 & 8.6 & 14.4 & 16.6 & 1.01\\
seed 81 & 3 & 470 & 0.9940 & 0.5 & 924 & 6{,}092 & 6.6 & 18.9 & 35.5 & 1.06\\
seed 81 & 3 & 470 & 0.9965 & 0.25 & 2{,}372 & 11{,}700 & 4.9 & 24.7 & 75.7 & 1.27\\
\addlinespace[2pt]
seed 343 & 2 & 106 & 0.8653 & 4 & 397 & 2{,}005 & 5.0 & 11.6 & 1.8 & 1.08\\
seed 343 & 2 & 106 & 0.8853 & 2 & 1{,}441 & 6{,}508 & 4.5 & 16.0 & 2.3 & 1.08\\
seed 343 & 2 & 106 & 0.8953 & 1 & 5{,}456 & 22{,}657 & 4.2 & 19.8 & 2.6 & 1.13\\
seed 343 & 2 & 106 & 0.9003 & 0.5 & 21{,}194 & 81{,}286 & 3.8 & 23.8 & 2.9 & 1.21\\
seed 343 & 2 & 106 & 0.9028 & 0.25 & 83{,}496 & 320{,}584 & 3.8 & 26.0 & 3.1 & 1.21\\
seed 343 & 3 & 470 & 0.9502 & 4 & 120 & 1{,}378 & 11.5 & 10.8 & 3.1 & 1.02\\
seed 343 & 3 & 470 & 0.9702 & 2 & 321 & 2{,}250 & 7.0 & 17.7 & 6.6 & 1.13\\
seed 343 & 3 & 470 & 0.9802 & 1 & 946 & 6{,}900 & 7.3 & 28.0 & 8.6 & 1.13\\
seed 343 & 3 & 470 & 0.9852 & 0.5 & 3{,}072 & 19{,}954 & 6.5 & 50.6 & 11.9 & 1.18\\
seed 343 & 3 & 470 & 0.9877 & 0.25 & 10{,}805 & 57{,}271 & 5.3 & 67.0 & 17.0 & 1.24\\
\addlinespace[2pt]
seed 777 & 2 & 106 & 0.9156 & 4 & 243 & 1{,}162 & 4.8 & 11.3 & 2.9 & 1.16\\
seed 777 & 2 & 106 & 0.9356 & 2 & 810 & 3{,}520 & 4.3 & 15.7 & 3.7 & 1.15\\
seed 777 & 2 & 106 & 0.9456 & 1 & 2{,}899 & 10{,}894 & 3.8 & 20.8 & 4.9 & 1.21\\
seed 777 & 2 & 106 & 0.9506 & 0.5 & 10{,}895 & 41{,}485 & 3.8 & 22.4 & 5.2 & 1.26\\
seed 777 & 2 & 106 & 0.9531 & 0.25 & 42{,}149 & 147{,}773 & 3.5 & 27.2 & 6.1 & 1.27\\
seed 777 & 3 & 470 & 0.9566 & 4 & 93 & 1{,}304 & 14.0 & 9.0 & 3.3 & 1.00\\
seed 777 & 3 & 470 & 0.9766 & 2 & 220 & 1{,}966 & 8.9 & 13.8 & 7.5 & 1.01\\
seed 777 & 3 & 470 & 0.9866 & 1 & 557 & 3{,}366 & 6.0 & 20.1 & 16.6 & 1.14\\
seed 777 & 3 & 470 & 0.9916 & 0.5 & 1{,}551 & 7{,}462 & 4.8 & 25.0 & 30.4 & 1.38\\
seed 777 & 3 & 470 & 0.9941 & 0.25 & 4{,}779 & 26{,}060 & 5.5 & 64.3 & 34.8 & 1.30\\
\bottomrule
\end{tabular}
\end{table}

\begin{table}[h!]\centering\small
\caption{\textbf{Five certificates of the claim $\Mt\le3$ in trained recurrent networks, on identical queries.} Median number of interventions at level $\alpha=0.05$ (1{,}000 audits per row; every audit ended in a certificate under every statistic). Product: the direct certificate for the whole table of 42 candidates. Betting product: product of one-sided mixtures. Union: union bound over candidates of sub-Gaussian (Hoeffding) or Bernoulli confidence sequences. Bound: $\log(1/\alpha)/D^\ast$ with $D^\ast=\kl(\tau+\Delta,\tau)$. Last column: median cost of the product certificate relative to the witness certificate. Stopping times are resolved to about $5\%$.}\label{tab:baselines}
\setlength{\tabcolsep}{3.5pt}
\resizebox{\linewidth}{!}{%
\begin{tabular}{@{}lcccccccc@{}}\toprule
 & & & \multicolumn{2}{c}{Estimate every candidate} & \multicolumn{3}{c}{Existential form} & \\\cmidrule(lr){4-5}\cmidrule(lr){6-8}
Network & gap $\Delta$ & bound & product & betting product & union, Hoeffding & union, Bernoulli & witness & saving\\\midrule
network 10 & $4.00\times10^{-2}$ & $2.93\times10^{2}$ & $5.84\times10^{3}$ & $4.95\times10^{3}$ & $3.57\times10^{3}$ & $1.56\times10^{3}$ & $1.28\times10^{3}$ & 4.5\\
network 10 & $2.00\times10^{-2}$ & $1.26\times10^{3}$ & $3.36\times10^{4}$ & $4.24\times10^{4}$ & $1.37\times10^{4}$ & $6.62\times10^{3}$ & $5.81\times10^{3}$ & 5.8\\
network 10 & $1.00\times10^{-2}$ & $5.23\times10^{3}$ & $1.65\times10^{5}$ & $2.30\times10^{5}$ & $5.51\times10^{4}$ & $2.63\times10^{4}$ & $2.25\times10^{4}$ & 7.4\\
network 10 & $5.00\times10^{-3}$ & $2.12\times10^{4}$ & $7.69\times10^{5}$ & $1.10\times10^{6}$ & $2.27\times10^{5}$ & $9.97\times10^{4}$ & $8.37\times10^{4}$ & 9.2\\
network 10 & $2.50\times10^{-3}$ & $8.56\times10^{4}$ & $3.33\times10^{6}$ & $5.13\times10^{6}$ & $9.35\times10^{5}$ & $3.97\times10^{5}$ & $3.28\times10^{5}$ & 10.2\\
network 10 & $1.25\times10^{-3}$ & $3.44\times10^{5}$ & $1.43\times10^{7}$ & $2.22\times10^{7}$ & $3.89\times10^{6}$ & $1.62\times10^{6}$ & $1.32\times10^{6}$ & 10.8\\
network 10 & $6.25\times10^{-4}$ & $1.38\times10^{6}$ & $6.18\times10^{7}$ & $9.58\times10^{7}$ & $1.66\times10^{7}$ & $6.92\times10^{6}$ & $5.21\times10^{6}$ & 11.8\\
network 10 & $3.12\times10^{-4}$ & $5.52\times10^{6}$ & $2.67\times10^{8}$ & $4.34\times10^{8}$ & $7.15\times10^{7}$ & $2.97\times10^{7}$ & $2.23\times10^{7}$ & 12.0\\
network 10 & $1.56\times10^{-4}$ & $2.21\times10^{7}$ & $1.15\times10^{9}$ & $1.88\times10^{9}$ & $2.94\times10^{8}$ & $1.22\times10^{8}$ & $8.70\times10^{7}$ & 13.3\\
network 10 & $7.81\times10^{-5}$ & $8.83\times10^{7}$ & $4.98\times10^{9}$ & $8.52\times10^{9}$ & $1.21\times10^{9}$ & $5.03\times10^{8}$ & $3.57\times10^{8}$ & 13.9\\
network 10 & $3.91\times10^{-5}$ & $3.53\times10^{8}$ & $2.15\times10^{10}$ & $3.68\times10^{10}$ & $5.23\times10^{9}$ & $2.07\times10^{9}$ & $1.47\times10^{9}$ & 14.6\\
network 10 & $1.95\times10^{-5}$ & $1.41\times10^{9}$ & $9.30\times10^{10}$ & $1.59\times10^{11}$ & $2.15\times10^{10}$ & $8.52\times10^{9}$ & $5.77\times10^{9}$ & 16.1\\
network 10 & $9.77\times10^{-6}$ & $5.65\times10^{9}$ & $4.02\times10^{11}$ & $6.88\times10^{11}$ & $9.08\times10^{10}$ & $3.68\times10^{10}$ & $2.49\times10^{10}$ & 16.1\\
\midrule
network 102 & $1.00\times10^{-3}$ & $5.38\times10^{5}$ & $2.34\times10^{7}$ & $3.61\times10^{7}$ & $6.26\times10^{6}$ & $2.65\times10^{6}$ & $2.14\times10^{6}$ & 10.9\\
network 102 & $2.50\times10^{-4}$ & $8.62\times10^{6}$ & $4.34\times10^{8}$ & $7.07\times10^{8}$ & $1.11\times10^{8}$ & $4.41\times10^{7}$ & $3.61\times10^{7}$ & 12.0\\
network 103 & $1.00\times10^{-3}$ & $5.38\times10^{5}$ & $2.33\times10^{7}$ & $3.61\times10^{7}$ & $6.26\times10^{6}$ & $2.65\times10^{6}$ & $2.19\times10^{6}$ & 10.6\\
network 103 & $2.50\times10^{-4}$ & $8.62\times10^{6}$ & $4.34\times10^{8}$ & $6.74\times10^{8}$ & $1.11\times10^{8}$ & $4.61\times10^{7}$ & $3.61\times10^{7}$ & 12.0\\
network 109 & $1.00\times10^{-3}$ & $5.38\times10^{5}$ & $2.33\times10^{7}$ & $3.61\times10^{7}$ & $6.25\times10^{6}$ & $2.61\times10^{6}$ & $2.14\times10^{6}$ & 10.9\\
network 109 & $2.50\times10^{-4}$ & $8.62\times10^{6}$ & $4.14\times10^{8}$ & $6.74\times10^{8}$ & $1.11\times10^{8}$ & $4.61\times10^{7}$ & $3.61\times10^{7}$ & 11.5\\
\bottomrule
\end{tabular}}
\end{table}

\begin{table}[h!]\centering\small
\caption{\textbf{Six language models on the indirect-object task.} Task accuracy (\%) on 4{,}096 prompts; largest interchange accuracy $\mu_S$ over sets of at most $k$ of the model's 14 pool heads (2{,}048 prompt pairs); circuit size $\Mt$ at $\tau=0.9$, which is exact over all 144 heads (\cref{app:lm}); number $K$ of candidates of the true claim $\Mt\le k$; median number of interventions for the witness certificate of that claim ($\alpha=0.05$, 1{,}000 audits), and the cost of the product certificate relative to it; and the smallest sufficient set (layer.head).}\label{tab:lm}
\setlength{\tabcolsep}{3pt}
\resizebox{\linewidth}{!}{%
\begin{tabular}{@{}lccccccccrrrl@{}}\toprule
 & & \multicolumn{6}{c}{best $\mu_S$ with $|S|\le k$} & & & & & \\\cmidrule(lr){3-8}
Model & acc. & 1 & 2 & 3 & 4 & 5 & 6 & $\Mt$ & $K$ & witness & product/witness & smallest sufficient set\\\midrule
GPT-2 small (OpenAI) & 99.6 & 0.072 & 0.466 & 0.905 & 0.982 & 0.997 & 0.998 & 3 & 470 & 93{,}916 & 73 & 8.10, 8.6, 7.9\\
seed 21 & 99.2 & 0.423 & 0.963 & 0.997 & 0.998 & 1.000 & 1.000 & 2 & 106 & 528 & 6 & 10.10, 8.4\\
seed 49 & 98.9 & 0.375 & 0.907 & 0.992 & 0.999 & 1.000 & 1.000 & 2 & 106 & 43{,}756 & 22 & 10.7, 8.0\\
seed 81 & 99.7 & 0.477 & 0.975 & 0.999 & 1.000 & 1.000 & 1.000 & 2 & 106 & 391 & 5 & 10.5, 8.7\\
seed 343 & 99.5 & 0.314 & 0.905 & 0.990 & 0.998 & 1.000 & 1.000 & 2 & 106 & 72{,}728 & 24 & 10.0, 8.6\\
seed 777 & 99.8 & 0.290 & 0.956 & 0.997 & 1.000 & 1.000 & 1.000 & 2 & 106 & 656 & 7 & 9.0, 8.7\\
\bottomrule
\end{tabular}}
\end{table}

\end{document}